\documentclass[11pt]{book}
\usepackage[english]{babel}
\usepackage[utf8]{inputenc}

\usepackage[a4paper,width=150mm,top=25mm,bottom=25mm,bindingoffset=6mm]{geometry}
\usepackage{fancyhdr}
\usepackage[ruled,vlined,linesnumbered,resetcount,algochapter]{algorithm2e}
\RestyleAlgo{boxruled}
\LinesNumbered

\usepackage{amsthm,amsmath,amsfonts}
\newtheorem{thm}{Theorem}[chapter]
\newtheorem{defn}{Definition}[chapter]
\newtheorem{lemma}{Lemma}[chapter]
\DeclareMathOperator*{\E}{\mathbb{E}}
\DeclareMathOperator*{\Prob}{\mathbb{P}}
\DeclareMathOperator*{\eps}{\varepsilon}
\allowdisplaybreaks

\usepackage{float}
\usepackage{multicol}
\usepackage[pdftex]{graphicx}
\usepackage{subcaption}
\graphicspath{{./Figures/}}
\usepackage{color}
\usepackage{epstopdf}

\usepackage[most]{tcolorbox}
\tcbset{
    center title,
    left=0pt,
    right=0pt,
    top=0pt,
    bottom=0pt,
    colback=blue!10!white,
    colframe=blue!50!black,
    width=\dimexpr\textwidth\relax,
    enlarge left by=0mm,
    boxsep=5pt,
}
    
\usepackage{hyperref}
\hypersetup{
    colorlinks=true,
    linkcolor=blue,
    filecolor=magenta,      
    urlcolor=blue,
}

\usepackage{tikz-cd}

\usepackage[autostyle]{csquotes}
\usepackage{epigraph}

\usepackage{biblatex}
\AtBeginDocument{}
\begin{document}

\begin{titlepage}
\newcommand{\HRule}{\rule{\linewidth}{0.5mm}}
\center % Center everything on the page

%----------------------------------------------------------------------------------------
%	HEADING SECTIONS
%----------------------------------------------------------------------------------------

\textsc{\LARGE TECHNICAL UNIVERSITY OF CRETE}\\[0.4cm] % Name of your university/college
\textsc{\normalsize SCHOOL OF ELECTRICAL \& COMPUTER ENGINEERING }\\[0.5cm] % Major heading such as course name
%\textsc{\large Department of Computing}\\[0.5cm] % Minor heading such as course title
\includegraphics[scale=0.75]{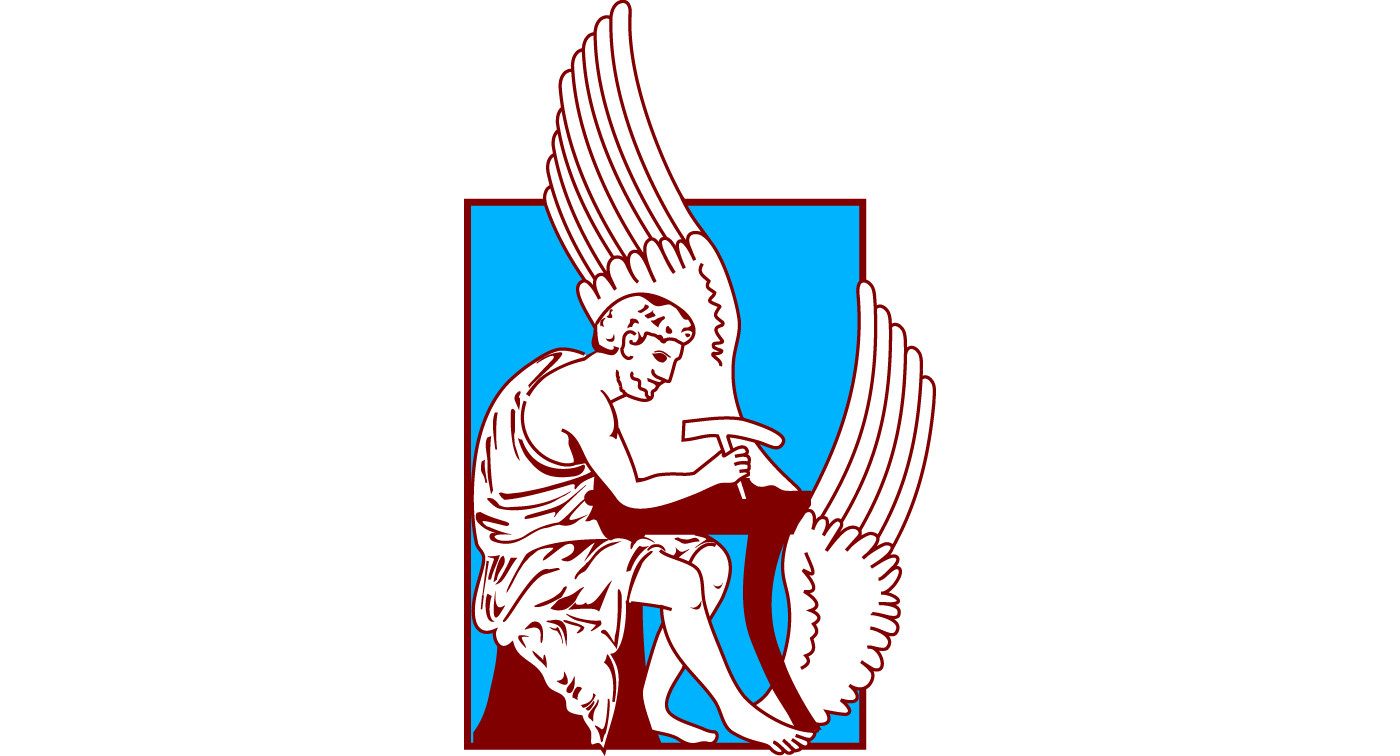}\\[1.5cm]
\textsc{\normalsize DIPLOMA THESIS }\\[0.75cm]
%----------------------------------------------------------------------------------------
%	TITLE SECTION
%----------------------------------------------------------------------------------------
\makeatletter
\HRule \\[0.4cm]
{ \Huge \bfseries Data Analytics with \\[0.3cm] Differential Privacy}\\[0.4cm]
%{ \huge \bfseries Distributed \& Streaming Data Analytics \\[0.3cm] with Differential Privacy}\\[0.4cm] % Title of your document
\HRule \\[1.5cm]
 
%----------------------------------------------------------------------------------------
%	AUTHOR SECTION
%----------------------------------------------------------------------------------------

\begin{minipage}{0.4\textwidth}
\begin{flushleft} \large
\emph{Author:}\\
Vassilis V. Digalakis, Jr.
\linebreak \linebreak \linebreak
\end{flushleft}
\end{minipage}
~
\begin{minipage}{0.4\textwidth}
\begin{flushright} \large
\emph{Thesis Committee:} \\
Prof. Minos Garofalakis (\emph{Advisor})\\
Prof. George Karystinos \\
Prof. Aggelos Bletsas
\end{flushright}
\end{minipage}\\[2cm]
\makeatother

\vfill 

%----------------------------------------------------------------------------------------
%	DATE SECTION
%----------------------------------------------------------------------------------------
{\small A thesis submitted in partial fulfillment of the requirements 
for the degree of \\ Diploma in Electrical and Computer Engineering }\\[0.2cm]
{\large July 14, 2018}
%{\large \today}\\[2cm] % Date, change the \today to a set date if you want to be precise

\end{titlepage}

\shipout\null
\setcounter{page}{1}
\pagenumbering{roman}

\chapter*{}
\thispagestyle{plain}
\begin{center}
    \Large
    \textbf{Data Analytics with Differential Privacy}
    
    \vspace{1cm}
    
    \large
    \textbf{by Vassilis V. Digalakis, Jr.}\\[0.3cm]
    \textbf{Thesis Advisor: Prof. Minos Garofalakis}
    
    \vspace{2cm}
    \Large
    \textbf{Abstract}
    \vspace{1cm}
\end{center}

Differential privacy is the state-of-the-art definition for privacy, that
\enquote{addresses the paradox of learning nothing about an individual while learning useful information about a population} (Dwork and Roth).
In other words, differential privacy guarantees that any analysis performed on a sensitive dataset leaks no information about the individuals whose data are contained therein.
In this thesis, we develop differentially private algorithms to analyze distributed and streaming data.

In the distributed model, we consider the particular problem of 
learning -in a distributed fashion- a global model of the data,
that can subsequently be used for arbitrary analyses.
We build upon PrivBayes, a differentially private method that approximates
the high-dimensional distribution of a centralized dataset
as a product of low-order distributions,
utilizing a Bayesian Network model.
Specifically, we examine three novel approaches to learning a global Bayesian Network from distributed data,
while offering the differential privacy guarantee to all local datasets.
Our work includes a detailed theoretical analysis of the distributed, differentially private entropy estimator which we use in one of our algorithms,
as well as a detailed experimental evaluation, using both synthetic and real-world data.

In the streaming model, we focus on the problem of estimating the density of a stream of users (or, more generally, elements),
which expresses the fraction of all users that actually appear in the stream.
We offer one of the strongest privacy guarantees for the streaming model, namely user-level pan-privacy, which ensures that the privacy of any user is protected,
even against an adversary that observes, on rare occasions, the internal state of the algorithm.
We provide a detailed analysis of an existing, sampling-based algorithm for the problem, and 
propose two novel modifications that significantly improve it, both theoretically and experimentally, by optimally using all the allocated \enquote{privacy budget}.

\chapter*{}
\thispagestyle{plain}
\begin{center}
    \Large    
    \textbf{Acknowledgments}
    \vspace{1cm}
\end{center}

After five years of truly hard yet exciting work, and before leaving Greece to pursue my PhD degree in the US,
I cannot but express my gratitude towards the people who were by my side and made everything I achieved possible.

First and foremost, I would like to thank my advisor, \emph{Prof. Minos Garofalakis}, 
for exposing me to research and pointing me towards exciting topics.
I am also grateful towards the other two members of the committee.
I had numerous intriguing discussions with \emph{Prof. George Karystinos}, both about technical issues and about Olympiacos.
I have to admit that the ones I enjoyed the most were -surprisingly!- the former. 
\emph{Prof. Aggelos Bletsas} always encouraged and supported me,
and taught me my first two MIT courses.
Hopefully the ones that will follow at MIT will be equally interesting.

I owe a lot to many, among my other Professors at TUC, especially 
\emph{Prof. Stavros Christodoulakis}, for his wise advice when I had to make critical decisions,
\emph{Prof. Michail Lagoudakis}, for his kindness and support, and
\emph{Prof. Daphne Manoussaki}, for offering me the opportunity to serve as a teaching assistant.

Nothing would have been possible without \emph{my family and my close friends}.
I want to thank \emph{my mother}, for her constant and unconditional support, and
\emph{Korina}, whose achievements inspired me as well, and motivated me to work even harder.
And of course, no words are enough to thank \emph{Areti}; 
she has been by my side both in good and in bad times, for many many years now.
Last but not least, I acknowledge the person to whom I owe the most, \emph{my father},
who managed -through his continuous guidance- to convert one of the world's most impatient persons, to a more patient, persistent and determined one.

\tableofcontents

\chapter{Introduction}
\setcounter{page}{1}
\pagenumbering{arabic}
\epigraph{ \enquote{Historically, privacy was almost implicit, because it was hard to find and gather information. But in the digital world, whether it's digital cameras or satellites or just what you click on, we need to have more explicit rules - not just for governments but for private companies.}}{\textit{Bill Gates}}

The importance of privacy in the era of Big Data is well-understood.
The recent Facebook-Cambridge Analytica scandal is only the last in a series of major privacy breaches,
that point out the need to reconsider our perspective on the significance of protecting the data that we generate.
At the same time, however, it would be a huge mistake to miss the opportunities
that the massive availability of data offers.

Until recently, the need for privacy was much less established,
as our data mostly lived within our personal computers and were protected by our passwords.
The need to securely store our data within our computers,
combined with the needs to perform secure electronic transactions and communicate securely (to mention but a few), 
were addressed by cryptography and formal approaches were developed early on in the history of computer science.

In contrast, nowadays most of our data live in the Cloud, or in the servers of Google, Facebook, Amazon etc.,
and only a tiny fraction of them still lives within our personal computers.
Furthermore, taking the explosion of analytics into account,
we may even not be able to prevent our data from being included in analyses,
both for advertisement and for research purposes.
How will our privacy be protected?

Cryptography may still be the answer in some cases,
but even if we manage to develop methods to analyze encrypted data sufficiently well,
there are still some unsurmountable problems.
Once an individual's data enter a database, they may even stay there forever;
nobody can guarantee that the cryptographic techniques used will still protect the individual's privacy in, say, 25 years from now.
Moreover, the effect of an individual's data on the result of an analysis, which will essentially be published, cannot be protected by cryptography.

That being said, we have to distinguish between security and privacy.
The fact that our data are securely stored today
does not mean that our privacy is protected; neither today nor in the future.
Classical cryptography has provided decent solutions in guaranteeing security.
In the case of privacy, however, a totally different calculus applies.
It is clear that a formal definition for privacy is also essential,
together with a set of tools (e.g. algorithms, software),
that will allow us guarantee that the privacy (according to this definition) of individuals who contribute their data to datasets and analyses is protected.

\section{The Promise of Differential Privacy}
\epigraph{ \enquote{You will not be affected, adversely or otherwise, by allowing your data to be used in any study or analysis, no matter what other studies, 
datasets or information sources are available.}}{\textit{The promise of differential privacy.\\Cynthia Dwork \& Aaron Roth}}

Early approaches in privacy-preserving data analysis suffered from a series of attacks (e.g. linkage attacks, differencing attacks, reconstruction attacks),
and commonly used privacy definitions, like k-anonymity \cite{sweeney2002k}, were shown to be flawed.
On the contrary, \emph{differential privacy} meets the requirements in being a formal privacy definition.
Introduced in 2006 by Cynthia Dwork, Frank McSherry, Kobi Nissim, and Adam Smith \cite{dwork2006calibrating}, 
differential privacy is a strong and mathematically rigorous guarantee,
that describes a promise, made by a data holder, 
to an individual that contributes its data to a dataset, 
that the individual's privacy will be protected.

As Dwork and Roth \cite{dwork2014algorithmic} argue, differential privacy addresses the paradox of learning nothing about an individual while learning useful information about a population.
In addition to being a strong privacy guarantee, differential privacy as a definition has mathematical properties
that facilitate the design of algorithms which satisfy it.
As a result, numerous methods that realize the differential privacy guarantee have been developed in the past few years,
and now we are definitely one step closer to privacy-preserving data analysis.

\section{Distributed and Streaming Data}
The paradigm of our data living in a static database that is protected by a curator (e.g. database administrator) belongs to the past to a large extent.
Nowadays, our data are everywhere and in various forms:
\begin{itemize}
\item[-] They are often distributed among several databases.
This is common, for example, in medical applications,
where different hospitals possess different parts of a joint database 
of the clinical records of individuals.
\item[-] Or, they are dynamically created and arrive continuously in a stream.
Being able to monitor such a stream and extract statistics is important for many disciplines, like -for instance- epidemiology.
\end{itemize}
We mention these two particular applications due to their obvious connections with privacy; medical data are by definition sensitive.

\section{Thesis Organization \& Contributions}
In this section we jointly outline the organization of this thesis and its key contributions. \\

In \textbf{Chapter 2}, we formalize the notion of differential privacy.
We give the basic definitions and theorems, and
present the most common mechanisms that are used as building blocks to provide the differential privacy guarantee.
In addition, we describe in detail the \emph{traditional differential privacy model}, and
emphasize some points that often cause confusion in the literature. \\

In \textbf{Chapter 3}, we develop differentially private algorithms to analyze \emph{distributed data}, and
we consider a model where a large dataset is horizontally distributed among mutually distrustful parties.
We address the particular problem of distributed learning of Bayesian
Networks with differential privacy.
The key contributions of the chapter are the following:
\begin{itemize}
\item[-] We formally describe differential privacy in the distributed model,
we examine several alternative approaches to solving our problem, and 
we provide a detailed survey of the related work that addresses ours, or similar models.
\item[-] We present PrivBayes \cite{zhang2014privbayes},\cite{zhang2017privbayes}, the state-of-the-art method in learning Bayesian Networks with differential privacy,
we identify the challenges that arise when moving PrivBayes from a centralized to a distributed environment, and
we examine three novel approaches to learning a global Bayesian Network from distributed data, while offering the differential privacy guarantee to all local datasets.
\item[-] We provide a detailed theoretical analysis of the distributed, differentially private entropy estimator which we use in one of our algorithms.
\item[-] We experimentally evaluate our algorithms using both synthetic and real-world data.
\end{itemize}

In \textbf{Chapter 4}, we develop differentially private algorithms to
analyze \emph{streaming data}, and
we consider the cashier-register streaming model.
We address the particular problem of estimating the density of a stream
of users, and
we offer one of the strongest privacy guarantees for the streaming model,
namely user-level pan-privacy,
which ensures that the privacy of any user is protected,
even against an adversary that observes, on rare occasions,
the internal state of the algorithm.
The key contributions of the chapter are the following:
\begin{itemize}
\item[-] We formally describe differential privacy in the streaming model,
we analyze in-depth the existing definitions and approaches, and
we provide an extensive survey of the related work.
\item[-] We provide for the first time a detailed analysis of the sampling-based, pan-private density estimator, proposed by Dwork et al. \cite{dwork2010pan},
we identify its main limitation, in that it does not use all the allocated privacy budget.
\item[-] We examine two different novel approaches to modifying the original estimator,
based on optimally tuning the Bernoulli distributions it uses, as well as on using continuous distributions (Laplace, Gaussian), and
we analyze the theoretic guarantees that our modified estimators offer.
\item[-] We experimentally compare our algorithms.
\end{itemize}

In \textbf{Chapter 5}, we conclude this thesis, and
we provide directions for future work and possible extensions to our solutions.

\chapter{Differential Privacy}
In this chapter, we formalize the notion of differential privacy.
In addition, we present some key theorems that follow from the definitions which we introduce,
and describe the most common mechanisms that are used as building blocks to provide the differential privacy guarantee.

We consider a \emph{model of computation}, where a database, or (more generally) a dataset $D$ contains the sensitive data of individuals, called the data owners;
each record/tuple/data point (the terms are used interchangeably) in $D$ corresponds to a single individual.
We assume the existence of a trusted entity by the data owners, called the data holder or curator.
The data holder has direct access to the sensitive dataset and analyzes it,
ensuring that any output produced by the analysis does not violate the owners' privacy.
A key thing to notice is that the data holder and the data analyst is a common (and trusted) entity.
A line of work considers the data analyst to be a separate, untrusted entity that observes the result of the analysis, which is performed by the (trusted) data holder.
We do not follow this approach, and we refer to the untrusted entity against which we want to protect the data as the adversary.
We call this the \emph{traditional differential privacy model}.

We also need to distinguish between the \emph{interactive} and the \emph{non-interactive} case.
\begin{itemize}
\item[-] In the interactive case, the data holder is adaptively asked queries on the sensitive dataset and still has to protect the owners' privacy.
In other words, the adversary observes the output of every analysis performed by the data holder, and may pose a new query based on this output.
\item[-] In the non-interactive case, the data holder analyzes the sensitive dataset only once.
The output of the analysis is published, so the adversary has access to it, but the original dataset is not accessed again and may even be destroyed.
\end{itemize}
In our work, we focus on the non-interactive case, so that the analysis that is to be performed on the sensitive data is decided and known in advance.

\section{Basic Definitions \& Theorems}
In defining differential privacy, it is useful to think of the dataset $D$ as a finite collection of records from a universe $\mathcal{U}$.
The histogram representation of $D$ is a function $\text{hist}(D): \mathcal{U} \rightarrow \mathbb{N}^{|\mathcal{U}|}$,
where the $i$-th entry $\text{hist}(D)_i$ (for some $i \in \{1,...,|\mathcal{U}|\}$) represents the number of tuples in $D$ of type $\mathcal{U}[i]$ (assuming that an arbitrary ordering is defined over $\mathcal{U}$).

We proceed by introducing the notion of adjacency between datasets, which (informally) refers to a pair of datasets that differ on a single record.
Depending on the interpretation of the word \enquote{differ}, two different definitions have been used in the literature.
\begin{defn}[Adjacency] \label{defn_adj}
Datasets $D$ and $D'$ are adjacent, denoted $adj(D,D')$, if ...
\begin{itemize}
\item[\textbf{A:}] ... $D$ can be obtained from $D'$ by either adding or removing a single record, so: 
$$||\text{hist}(D)-\text{hist}(D')||_1 = \sum_{i=1}^{|\mathcal{U}|} |\text{hist}(D)_i-\text{hist}(D')_i| \leq 1$$
\item[\textbf{B:}] ... $D$ can be obtained from $D'$ by changing the value of a single record, so: 
$$||\text{hist}(D)-\text{hist}(D')||_1 = \sum_{i=1}^{|\mathcal{U}|} |\text{hist}(D)_i-\text{hist}(D')_i|  \leq 2 $$
\end{itemize}
\end{defn}
As we will shortly see, the notion of adjacency plays a key role in the definition of differential privacy,
and depending on the approach we follow to defining adjacency, we end up with a (slightly) different flavor of differential privacy (Kifer and Machanavajjhala \cite{kifer2011no}).
If Definition \ref{defn_adj}(\textbf{A}) is used, we speak of \emph{unbounded} differential privacy, whereas if Definition \ref{defn_adj}(\textbf{B}) is adopted, we speak of \emph{bounded} differential privacy.
Bounded differential privacy derives its name from the fact that the adjacent datasets involved have essentially the same size;
in unbounded differential privacy there is no such restriction.
In the literature, both approaches are assumed to be fine and one is free to choose whichever is more convenient.

We are now ready to formally define differential privacy, which intuitively guarantees that a randomized algorithm accessing a sensitive dataset produces similar outputs on similar (adjacent) inputs.
As a result, the impact of any single record (individual) on the algorithm's output is negligible, and hence no information is leaked about the individuals whose data are in the dataset.
\begin{defn}[Differential Privacy] \label{defn_dp}
A randomized algorithm $\textbf{Alg}: \mathcal{U} \rightarrow \mathcal{O}$ is $\eps$-differentially private
if for all $O \subseteq \mathcal{O}$, and for all pairs of adjacent datasets $D,D'$,
$$ \Prob[\textbf{Alg}(D) \in O ] \ \leq \ e^{\eps} \Prob[\textbf{Alg}(D') \in O ]$$
where the probability space is over the coin flips of $\textbf{Alg}$.
\end{defn}

The parameter $\eps$, called the \emph{privacy budget}, quantifies the privacy risk.
In general, smaller values of $\eps$ imply more privacy, as the distributions of outputs of the algorithm for adjacent inputs tend to \enquote{come closer}.
How we should pick $\eps$ in practice is an open question;
Hsu et al. \cite{hsu2014differential} address this question and propose a method for choosing $\eps$.

An additional remark is that $\eps$-differential privacy is a statistical in nature, \emph{worst-case guarantee}, in that it ensures that for every run of algorithm $\textbf{Alg}$ (on  every possible input), the output observed is almost equally likely to be observed on every adjacent input.
Therefore, several \emph{relaxations} of the original Definition \ref{defn_dp} have been considered; for example, $(\eps,\delta)$-differential privacy guarantees that for every pair of adjacent datasets, it is extremely unlikely that the observed output will be more likely to be observed with the one input of the pair than with the other.
In other words, $(\eps,\delta)$-differential privacy allows a privacy leakage with some small probability $\delta$.

We next present some key theorems; their proofs can be found in \cite{dwork2014algorithmic}. Theorem \ref{thm_post_proc} is known as the Post-Processing Theorem, and illustrates that differential privacy is immune to post-processing;
an adversary cannot compute a function of the output of a differentially private algorithm and make it less private.
\begin{thm}[Post-Processing] \label{thm_post_proc}
Let $\textbf{Alg}:\mathcal{U} \rightarrow \mathcal{O}$ be a randomized algorithm that satisfies $\eps$-differential privacy.
Let $f:\mathcal{O} \rightarrow \mathcal{O'}$ be an arbitrary randomized mapping.
Then $f(\textbf{Alg}): \mathcal{U} \rightarrow \mathcal{O'}$ also satisfies $\eps$-differential privacy.
\end{thm}

Theorem \ref{thm_composition}, known as the Sequential Composition Theorem, shows that if we compose multiple differentially private mechanisms, the privacy budget adds up.
For simplicity, we only present the theorem for the case of two mechanisms, but we note that it holds in general.
\begin{thm}[Sequential Composition] \label{thm_composition}
Let $\textbf{Alg}_1: \mathcal{U} \rightarrow \mathcal{O}_1$ be a randomized algorithm that satisfies $\eps_1$-differential privacy, and
let $\textbf{Alg}_2: \mathcal{U} \rightarrow \mathcal{O}_2$ be a randomized algorithm that satisfies $\eps_2$-differential privacy.
Then their combination $\textbf{Alg}_{1,2} = (\textbf{Alg}_1,\textbf{Alg}_2):\ \mathcal{U} \rightarrow \mathcal{O}_1 \times \mathcal{O}_2$ satisfies $(\eps_1+\eps_2)$-differential privacy.
\end{thm}

Differential privacy also composes when a sequence of differentially private mechanisms is applied in parallel on non-intersecting subsets of the entire dataset. Theorem \ref{thm_parallel_composition} is known as the Parallel Composition Theorem.
\begin{thm}[Parallel Composition] \label{thm_parallel_composition}
Let $\textbf{Alg}_1: \mathcal{U} \rightarrow \mathcal{O}_1$ be a randomized algorithm that satisfies $\eps_1$-differential privacy, and
let $\textbf{Alg}_2: \mathcal{U} \rightarrow \mathcal{O}_2$ be a randomized algorithm that satisfies $\eps_2$-differential privacy.
Let $D_1,D_2$ be two partitions of the dataset $D$, such that $D_1 \cup D_2 = D$ and $D_1 \cap D_2 = \emptyset$.
Then their combination $\textbf{Alg}_{1,2}(D) = (\textbf{Alg}_1(D_1),\textbf{Alg}_2(D_2)):\ \mathcal{U} \rightarrow \mathcal{O}_1 \times \mathcal{O}_2$ satisfies $(\max\{\eps_1,\eps_2\})$-differential privacy.
\end{thm}
An important thing to notice is that parallel composition does not hold for bounded differential privacy;
Definition \ref{defn_dp} is violated when removing a record from $D_1$ and adding a record to $D_2$ (or vice-versa).
Nevertheless, if our goal is to separately offer the differential privacy guarantee to each subset of $D$,
so that the adjacency relation is considered separately for $D_1$ and $D_2$,
then parallel composition does hold.
This is the case in the approach we follow to applying differential privacy in the distributed model, in Chapter 3.

Finally, the next Theorem \ref{thm_group}, protects the privacy of groups of size $k$.
It also addresses the case that multiple $(k)$ records in the dataset refer to the same individual.
\begin{thm}[Group Privacy] \label{thm_group}
Any $\eps$-differentially private algorithm $\textbf{Alg}: \mathcal{U} \rightarrow \mathcal{O}$ is $(k\eps)$-differentially private for groups of size $k$. That is, for all datasets $D,D'$ such that $||\text{hist}(D)-\text{hist}(D')||_1 \leq k$ and for all $O \subseteq \mathcal{O}$,
$$ \Prob[\textbf{Alg}(D) \in O ] \ \leq \ e^{k\eps} \Prob[\textbf{Alg}(D') \in O ]$$
where the probability space is over the coin flips of $\textbf{Alg}$.
\end{thm}

Now that we have developed a better understanding of differential privacy, we make an additional remark on definition \ref{defn_adj}.
Any algorithm that satisfies $\eps$-unbounded differential privacy also satisfies $2\eps$-bounded differential privacy, since changing the value of one record is equivalent to first removing the old version of the record and then adding the new one.

\section{Achieving Differential Privacy}
Differential privacy is a definition, and not an algorithm.
In practice, we are interested in developing algorithms that satisfy Definition \ref{defn_dp} and hence offer the differential privacy guarantee to their input datasets.
\begin{defn} \label{defn_priv_mech}
A randomized algorithm that satisfies Definition \ref{defn_dp} is called a privacy mechanism.
\end{defn}
As the reader may have noticed, a privacy mechanism is \emph{essentially} a \emph{randomized algorithm}, i.e.
an algorithm that employs a degree of randomness as part of its logic and produces an output that is a random variable (or vector).
We do not formally introduce the notion of randomized algorithms here and we refer the interested reader to the book by Mitzenmacher and Upfal \cite{mitzenmacher2005probability}.

In this section, we present three primitive differentially private mechanisms, which we use throughout our work.
More sophisticated mechanisms have been developed (e.g. sparse vector technique, multiplicative weights mechanism, subsample and aggregate framework),
that achieve much better results by reconsidering the computational goal of specific tasks.
We refer the interested reader to the monograph by Dwork and Roth \cite{dwork2014algorithmic} for a detailed presentation of such mechanisms.

\subsection{Randomized Response}
Randomized response is a research method proposed by Warner \cite{warner1965randomized} that allows respondents to a survey on a sensitive issue to protect their privacy against the interviewer, while still providing credible answers.
We next present an simple version of randomized response, based on an example Dwork and Roth \cite{dwork2014algorithmic}.
We slightly modify the original example, in that the data holder perturbs the sensitive dataset $D$ before publishing or analyzing it.
We assume that $D$ consists of a binary record $b$ (a bit) per individual,
which indicates whether the individual does or does not have a particular property.

\begin{algorithm}%[H]

\caption{Simple Randomized Response\label{Alg_rand_resp}}

\DontPrintSemicolon

\KwIn{Dataset $D$}

\KwOut{Differentially private dataset $\tilde{D}$}

Initialize $\tilde{D}=\emptyset$\;

\For{each $b \in D$}{
	Flip a fair coin\;
	\uIf{ \textbf{tails} }{
		$\tilde{b} = b$\;
	}
	\Else{
		Flip a second fair coin\;
		\uIf{ \textbf{heads} }{
			$\tilde{b} = 1$\;
		}
		\Else{
			$\tilde{b} = 0$\;
		}
	}
	Add $\tilde{b}$ to $\tilde{D}$ \;
}

Return $\tilde{D}$ \;

\end{algorithm}

Theorem \ref{thm_rand_resp} examines the privacy guarantees of \ref{Alg_rand_resp}.
We remark that, throughout this work, by $\log$ we refer to the natural logarithm.
\begin{thm}[Randomized response] \label{thm_rand_resp}
The version of randomized response described in Algorithm \ref{Alg_rand_resp} satisfies $\log 3$-differential privacy.
\end{thm}

The power of randomized response is that it provides plausible deniability, and it directly perturbs the sensitive dataset (privacy by process).
As a result, even if an individual's record indicates that it has the property in question, the individual may still credibly argue that it does not.

\subsection{Laplace Mechanism}
The Laplace mechanism \cite{dwork2006calibrating} provides a way to transform a numeric function $f: \mathcal{U} \rightarrow \mathbb{R}^N$ (that inputs a dataset $D$ and outputs a vector $f(D) \in \mathbb{R}^N$) into a differentially private mechanism.
We first introduce the notion of the sensitivity of $f$, which intuitively captures the effect of a single record on the output of $f$.
\begin{defn}[$\ell_1$-sensitivity] \label{defn_sensitivity}
The $\ell_1$-sensitivity of a function $f: \mathcal{U} \rightarrow \mathbb{R}^N$ is:
$$\Delta f = \max_{adj(D,D')} ||f(D)-f(D')||_1$$
\end{defn}

We next introduce the Laplace distribution, which is a symmetric, double-sided version of the exponential distribution.
Throughout our work, we slightly abuse notation and use $\text{Laplace}(\mu,b)$ (instead of $X \sim \text{Laplace}(\mu,b)$) to refer to a random variable that follows the Laplace Distribution.
Similarly, we use $\text{Bernoulli}(p)$ to refer to a Bernoulli random variable with parameter $p$, and so forth.
\begin{defn}[Laplace distribution] \label{defn_laplace}
A random variable $X \sim \text{Laplace}(\mu,b)$ has probability density function:
$$f_X(x|\mu,b) = \frac{1}{2b} e^{-\frac{|x-\mu|}{b}} \ , \ x \in \mathbb{R}$$
where $\mu$ is a location parameter, such that $\E[X] = \mu$, and $b>0$ is a scale parameter, such that $\text{var}(X) = 2b^2$.
\end{defn}

\begin{thm}[Laplace mechanism] \label{defn_laplace_mech}
Given any function $f: \mathcal{U} \rightarrow \mathbb{R}^N$, the Laplace mechanism, that on input $D \in \mathcal{U}$ outputs:
$$\tilde{f}(D) = f(D) + [X_1 \ ... \ X_N]^T $$
where $X_i$ are i.i.d. $\text{Laplace}(0,\frac{\Delta f}{\eps})$ random variables, satisfies $\eps$-differential privacy.
\end{thm}

\subsection{Exponential Mechanism}
The main limitation of the Laplace mechanism is that it can only handle numeric functions of the dataset.
In contrast, the Exponential mechanism \cite{mcsherry2007mechanism} provides a way to transform arbitrary (e.g. categorical) functions of the dataset into differentially private mechanisms.
We now consider an arbitrary function $f: \mathcal{U} \rightarrow \mathcal{O}$ that maps an input dataset $D$ to an arbitrary object $O \in \mathcal{O}$.
The Exponential mechanism is based upon a scoring/quality/utility function (the terms are used interchangeably) $q: \mathcal{U} \times \mathcal{O} \rightarrow \mathbb{R}$, that measures the quality $q(D,O)$ of the output $O$ when the input is $D$.

\begin{thm}[Exponential mechanism] \label{defn_expo_mech}
Given any function $f: \mathcal{U} \rightarrow \mathcal{O}$, the Exponential mechanism, that on input $D \in \mathcal{U}$ outputs an element $O \in \mathcal{O}$ with probability
$\propto  e^{\frac{\eps q(D,O)}{2 \Delta q}}$,
where $q: \mathcal{U} \times \mathcal{O} \rightarrow \mathbb{R}$
and $\Delta q = \max_{O \in \mathcal{O}} \max_{adj(D,D')} |q(D,O)-q(D',O)|$,
satisfies $\eps$-differential privacy.
\end{thm}

The Exponential mechanism defines a distribution $\textbf{p}_{EM}$ over the set of possible outputs, and then samples from $\textbf{p}_{EM}$.
Intuitively high-quality outputs are favored, as they are more likely to be sampled.
$\textbf{p}_{EM}$ can be arbitrarily complex, so the Exponential mechanism can be a double-edged sword:
\begin{itemize}
\item[-] On the one hand, the Exponential mechanism is general; for example, the Laplace mechanism can be viewed as an instance of the Laplace mechanism.
\item[-] On the other hand, in many occasions it may not even be possible to sample efficiently from $\textbf{p}_{EM}$.
\end{itemize}

\section{Data Analytics with Differential Privacy}
Differential privacy has been established as the state-of-the-art model in privacy-preserving data analytics.
Therefore, there now exists a vast literature in the field of data analytics with differential privacy; a survey of the related work is beyond the scope of this thesis.
We refer the interested reader to the survey by Chaudhuri and Sarwate \cite{sarwate2013signal} (and the many references therein), who provide an overview of the work that connects differential privacy with many areas that relate to data analytics,
namely statistics and robust statistics, signal processing and machine learning (classification, regression, dimensionality reduction, filtering).
Nevertheless, in Chapters 3 and 4 we do extensively review efforts that relate to the models that we examine, namely the distributed and the streaming model.

\chapter{Distributed Bayesian Network Learning with Differential Privacy}
\begin{figure}%[H]%[ht!]
    \centering
    \includegraphics[width=\columnwidth]{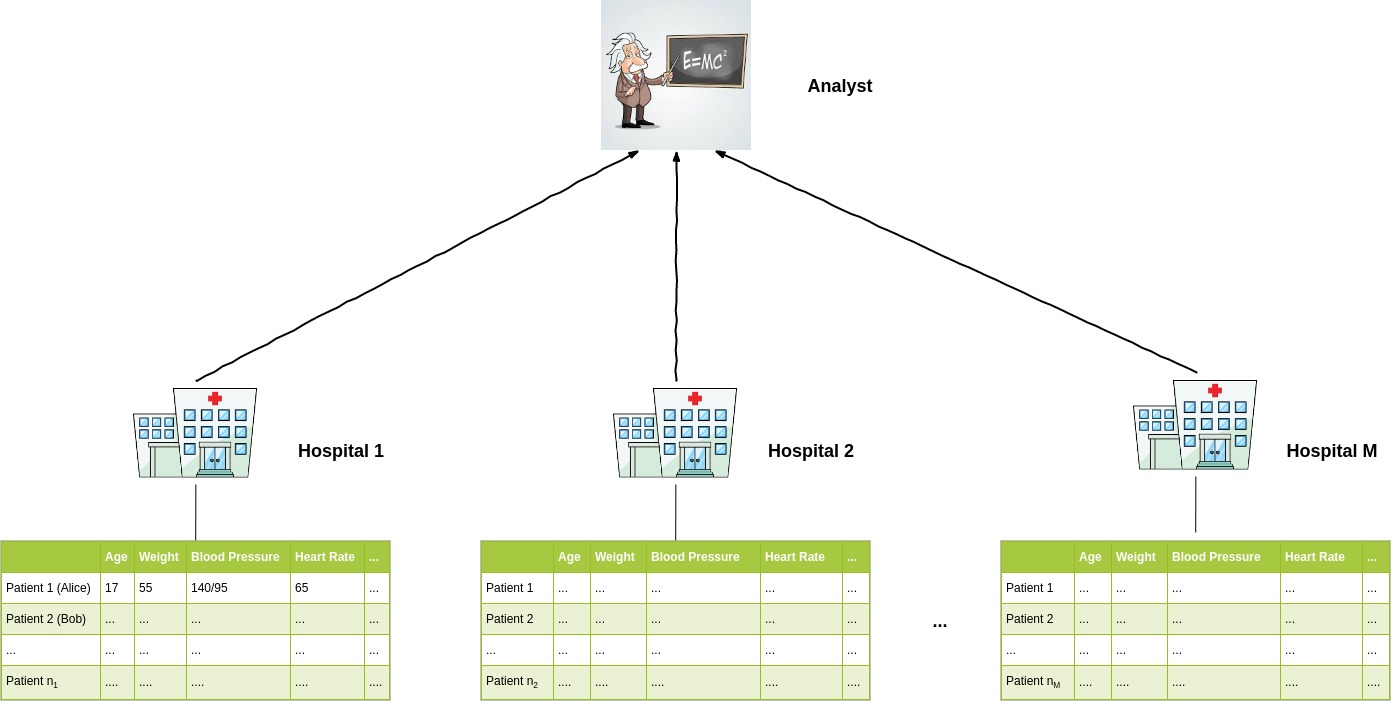}
    \caption{Motivating application}
	\label{FIG_motivating_application}
\end{figure}

We consider a model where
a large dataset is \emph{horizontally distributed among mutually distrustful parties} (data holders) that are not able or willing to share their part,
forming a distributed database.
Our goal is to perform privacy-preserving data mining,
and in particular, to make inferences about the population,
without compromising the privacy of the individuals whose data are used.
The model we described applies, for instance, in biomedical data analysis,
and constitutes a major limitation in biomedical research.
Hospitals and other trustworthy entities maintain the clinical records of individuals, but are unable to share and accurately analyze them, due to the risk of privacy breaches (Figure \ref{FIG_motivating_application}).

We aim to achieve something more general compared to (most) previous approaches in \emph{privacy-preserving data mining over distributed data}.
In this direction, we examine three approaches in learning -in a distributed and privacy-preserving fashion-
a model that approximates the (high-dimensional) data distribution as a product of low-order marginal and conditional distributions.
This is achieved by exploiting the attribute dependencies that are present in the data distribution;
the theory of probabilistic graphical models offers tools to efficiently examine these dependencies.
Our first approach is an exact approach, in that it requires each data holder to share noisy versions of the algorithm's sufficient statistics;
we also provide a detailed theoretical analysis of this approach.
The other two approaches are based on heuristic techniques, inspired from popular ideas from the distributed machine learning literature.
Finally, once we have the privacy-preserving approximation of the data distribution in hand, we are able to perform arbitrary analyses on our data (e.g. classification).

We assume that our distributed database consists of homogeneous records,
and each record consists of $N$ attributes $X_1, ... , X_N$, from a set $\mathcal{A}$.
In statistical machine learning, each attribute is viewed as a random variable,
and, thus, a record $\mathbf{x}$ (or data point - the terms are used interchangeably) can be viewed as a realization of the random vector $\mathbf{X} = [X_1 \ ... \ X_N]^T$.
All attributes are observed, so there exist no hidden variables.
A (relatively small) number of missing values may exist for some attributes; this can be faced using elementary techniques, like mean imputation.

We also assume that we deal with discrete data,
so $\forall i\in \{1,...,N\}$, $X_i$ takes values from the discrete alphabet $\text{domain}(X_i)$.
If continuous attributes exist, we propose to discretize them using any discretization technique;
in doing so, we are able to accurately estimate quantities like the mutual information, that play a major role in our solution.
On the contrary, estimating such quantities for continuous data is a much harder problem.

We next introduce some notation, that will allow us to formally describe our model.

\begin{table}[H]%[!t]
\renewcommand{\arraystretch}{1.3}
\caption{Table of notations}
\label{table notation}
\centering
\begin{tabular}{|c||c|}
\hline
\textbf{Notation} & \textbf{Description}\\
\hline
\hline
$N$ & Data dimension - number of attributes per data point\\
\hline
$X_i$ & Attribute $i\ (i\in\{1,...,N\})$\\
\hline
$\mathcal{A}$ & Set of all attributes: $\mathcal{A}=\{X_1,...,X_N\}$ \\
\hline
$\mathbf{x}$ & Data point: $\mathbf{x} = (x_1,...,x_N)\ (\forall i\in\{1,...,N\}:\ x_i \in \text{domain}(X_i))$ \\
\hline
\hline
$M$ & Number of data holders (degree of distribution)\\
\hline
$D_j$ & Dataset of holder $j\ (j\in\{1,...,M\})$ \\
\hline
$n_j$ & Size of dataset of holder $j\ (j\in\{1,...,M\})$: $n_j = |D_j|$ \\
\hline
$n$ & Size of total dataset: $n = \sum_{j=1}^{M}n_j$ \\
%\hline
%\hline
%$\mathcal{G}$ & Bayesian network structure \\
%\hline
%$\mathbf{Pa}(X_i)$ & Parents of attribute $i$ in the BN structure \\
%\hline
%$\mathbf{CPT}$ & BN parameters (Conditional Probability Tables) \\
%\hline
%$k$ & BN in-degree \\
%\hline
%\hline
%$\Prob(\bullet)$ & Probability of event $\bullet$\\
%\hline
%$I(\bullet;\star)$ & Mutual information between $\bullet$ and $\star$\\
\hline
\end{tabular}
\end{table}
Therefore, for some $i\in \{1,...,N\}$, by $X_i$ we refer to the $i$-th attribute (random variable) in $\mathcal{A}$, while $x_i$ denotes a realization of $X_i$.
Accordingly, by $\mathbf{X}$ we refer to the random vector that consists of all $N$ attributes in $\mathcal{A}$, while $\mathbf{x}$ denotes a realization of $\mathbf{X}$ (a data point).
In addition, for some $j \in \{1,...,N\}$ and $j\not=i$,
consider the set $\mathcal{A'} = \{X_i,X_j,...\} \subseteq \mathcal{A}$.
By $\mathbf{X}_{\mathcal{A'}}$ we refer to the random vector that consists of all attributes in $\mathcal{A'}$, while $\mathbf{x}_{\mathcal{A'}}$ denotes a realization of $\mathbf{X}_{\mathcal{A}'}$,
which is in fact a tuple that consists of the values that the attributes in $\mathcal{A'}$ take in data point $\mathbf{x}$.
So, for instance, both $\mathbf{X}$ and $\mathbf{X}_{\mathcal{A}}$ refer to the same random vector, and $\mathbf{x}$, $\mathbf{x}_{\mathcal{A}}$ both refer to realizations of this random vector.

Finally, we introduce the notation that we use for frequency and probability distributions.
We denote by $d_{\mathcal{A'}}$ the Cartesian product of the domains of the attributes in $\mathcal{A'}$:
$$d_{\mathcal{A'}}=\text{domain}(X_i) \times \text{domain}(X_j) \times ...$$
Also, if $|d_{\mathcal{A'}}|=d$, and we define an arbitrary ordering over the set $d_{\mathcal{A'}}$,
then for some $k \in\{1,...,d\}$, $d_{\mathcal{A'}}[k]=(x_i,x_j,...)$ represents the $k$-th element in $d_{\mathcal{A'}}$,
and is, in fact, one of the $d$ possible realizations of the random vector $\mathbf{X}_{\mathcal{A'}}$.
Given a dataset $D$, the joint frequency distribution of the attributes in $\mathcal{A'}$ is defined (and computed) as:
$$\mathbf{c}_{\mathbf{X}_{\mathcal{A'}} }
= \mathbf{c}_{\mathcal{A'}} 
= [c_1\ ...\ c_d]^T 
, \ \text{ where } c_k 
= \sum_{\mathbf{x} \in D} \mathbf{1}( \ \mathbf{x}_{\mathcal{A'}}=d_{\mathcal{A'}}[k] \ )$$
Since we view the attributes as random variables, we make the assumption that there exists an underlying probability distribution (that generated $D$);
we denote the joint probability distribution of the attributes in $\mathcal{A'}$ by:
$$\mathbf{p}_{\mathbf{X}_{\mathcal{A'}} }
= \mathbf{p}_{\mathcal{A'}} 
= [p_1\ ...\ p_d]^T
, \ \text{ where } p_k 
= \Prob(\ \mathbf{X}_{\mathcal{A'}}=d_{\mathcal{A'}}[k] \ )$$
Of course, in practice we only have the data in $D$; $\mathbf{p}_{\mathcal{A'}}$ is just a model that lives in our head. 
The maximum likelihood estimate of $\mathbf{p}_{\mathcal{A'}}$ is computed from $\mathbf{c}_{\mathcal{A'}}$ as: 
$$\mathbf{\hat{p}}_{\mathcal{A'}} = \frac{1}{|D|}\mathbf{c}_{\mathcal{A'}}$$
In general, using $\hat{}$ we refer to the maximum likelihood (empirical) estimate of the underlying quantity, and using $\tilde{}$ we refer to a perturbed (noisy) estimate. \\
Finally, we assume that there exist no zero-probability events, so $p_i > 0, \ \forall i \in \{1,...,d\}$,
although it is possible not to observe an event, so that $c_i = 0$, for some $i \in \{1,...,d\}$.

\section{Differential Privacy in the Distributed Model}
Recall that, in the \emph{traditional differential privacy model},
it is assumed that there exists a trusted entity by the data owners, whose sensitive data are in the dataset, called the data holder or curator.
This entity has direct access to the private dataset and analyzes it,
ensuring that any output produced by the analysis satisfies differential privacy.
Therefore, the data holder and the data analyst is a common (and trusted) entity.

This is not the case in several real-world applications;
the data owners often do not trust the entity that collects and analyzes their data, so the traditional model needs to be modified.
The \emph{local model}, introduced by Kasiviswanathan et al. \cite{kasiviswanathan2011can}, is motivated by such situations, and is, in fact, a generalization of randomized response.
Each data owner maintains its own database (of size 1 - a single data point), and provides the analyst with noisy answers to queries about it (satisfying differential privacy).

In our model, there are two additional changes.
Firstly, the data holder and the data analyst are \emph{separate entities}.
The former is trusted by the data owners (e.g. hospitals), whereas the latter is not.
Secondly, we assume that there exist \emph{multiple data holders} (all trusted), each maintaining the sensitive data that belong to a subset of the data owners.
We also assume that these subsets are non-overlapping, so each data owner's record is stored by exactly one data holder. 
Therefore, although the local model could apply to our model, we can avoid having each data owner perturb its own record, and make the data holders responsible for the perturbation.
In particular, each trusted data holder collects its subset of the sensitive data from the data owners and either responds to queries, or performs arbitrary analyses on them;
critically, the answers given must satisfy differential privacy (using the standard definition), and the overall privacy budget consumed must meet the privacy requirements. 
Then, the analyst only gets to see these answers, and since differential privacy is immune to post-processing, the owners' privacy is preserved. \\

Several efforts in the differential privacy literature have addressed this, or closely related models.
A line of work attempts to combine differential privacy with \emph{secure multiparty computation}.
For example, Pathak et al. \cite{pathak2010multiparty} propose a method for the analyst to learn a global, differentially private classifier from locally trained classifiers, which are perturbed and aggregated using a secure protocol.
Alhadidi et al. \cite{alhadidi2012secure} address the problem of privacy-preserving data publishing, in the special case that the data are horizontally partitioned among two parties;
they develop a two-party protocol for the exponential mechanism, and their solution relies on generalization.
Goryczka et al. \cite{goryczka2013secure} consider the problem of secure sum aggregation in the distributed model (using various secure multiparty computation protocols and encryption schemes),
while preserving differential privacy for the aggregated data (using distributed versions of the Laplace mechanism).

The main limitation of these hybrid solutions is that they sacrifice some of the most significant advantages of differential privacy;
for example, they introduce the need of having a secret key, and they require making assumptions about the computational power of the adversary.
To avoid such limitations, we focus on achieving \emph{pure differential privacy}.
We identify two paths we could follow to address our model:
\begin{itemize}
\item[-] \textbf{Distributed data mining with differential privacy.}
The (untrusted) analyst develops a distributed algorithm, so it communicates with the (trusted) data holders during its execution.
The data holders respond to the analyst's queries, ensuring that their answers satisfy differential privacy. \\
This approach can be viewed as a collaborative approach, since a global model is jointly learned, and its major advantage is that the entire dataset is directly utilized.
\item[-] \textbf{Data publishing with differential privacy.}
Each (trusted) data holder constructs a model (e.g. a probabilistic graphical model) or synopsis (e.g. a histogram) of its dataset, which it then publishes, or uses to generate and publish a synthetic dataset.
If the published model/synopsis/dataset satisfies differential privacy,
then any analysis performed on it will also guarantee differential privacy,
since differential privacy is immune to post-processing.
The (untrusted) analyst collects all models/synopses/datasets, merges them, and runs a centralized, non-private algorithm on the merged result. \\
Clearly, the major advantage of this approach is that the published datasets can be used for arbitrary analyses.
Zheng \cite{zheng2015differential} investigates (among others) the implications of publishing the model  versus publishing a synthetic dataset (generated using the model).
\end{itemize}

\subsection{Distributed Data Mining with Differential Privacy}
Most efforts on distributed data mining with differential privacy are task-specific, so the result of the distributed, differentially private algorithm is, for instance, a privacy-preserving \emph{classifier}.

The parallelizable nature of \emph{stochastic gradient descent} facilitates the design of distributed machine learning algorithms that are based on this optimization method.
Rajkumar et al. \cite{rajkumar2012differentially} build a global classifier, not by combining locally trained classifiers (like Pathak et al. \cite{pathak2010multiparty}),
but instead, by directly optimizing the overall multiparty objective;
their solution is based on Gaussian objective perturbation and guarantees $(\eps,\delta)$-differential privacy.
Shokri et al. \cite{shokri2015privacy} propose a method that allows multiple data owners to jointly learn a neural network model, without even sharing their sensitive datasets;
each data owner only shares small subsets of its models parameters during training.
Direct privacy leakage is prevented, since no sensitive dataset is shared, 
and indirect privacy leakage, which could occur by sharing the parameters, is also prevented by using the sparse vector technique (a primitive differentially private mechanism).

Some authors consider a slightly modified model;
they assume the existence of a \emph{public dataset}, originating from data owners who are willing to share their data,
The public dataset is used to enhance the utility of their analyses, which are all classification-related. 
Ji et al. \cite{ji2014differentially} develop a method to train a differentially private logistic regression model from distributed data.
Their solution is based on properly modifying the Newton-Raphson method, which they use to solve the underlying optimization problem.
Xie et al. \cite{xie2016data} propose an ensemble learning method to aggregate locally trained binary classifiers and regressors.
The local models are trained in a differentially private manner (using the expected risk minimization technique of Chaudhuri et al. \cite{chaudhuri2011differentially}), and the aggregation method is based on using the public data.
Hamm et al. \cite{hamm2016learning} also investigate the problem of aggregating locally trained classifiers (that guarantee differential privacy), but they assume that the public dataset consists of auxiliary unlabeled data.
Their main contribution is in the aggregation method;
they demonstrate the limitations of using majority voting,
and therefore propose a new risk, weighted by class probabilities (estimated from
the ensemble).

Wahab et al. \cite{abdel2014darm} examine a different data mining task,
and in particular, they propose a distributed \emph{association rules mining} framework.
Their solution is based on having the data owners anonymize their data using a simple differentially private scheme.

\subsection{Data Publishing with Differential Privacy} \label{section data publishing dp}
As we already noted, the main advantage of this approach is that it produces a general and query/task independent result that can be used for arbitrary analyses.
Hence, a vast literature has been developed on data publishing with differential privacy.
An important line of work is based on constructing and publishing differentially private \emph{synopses} of the input dataset.
Barak et al. \cite{barak2007privacy} examine the release of contingency tables (frequency distributions) via Fourier decompositions.
Hay et al. \cite{hay2010boosting} and Xu et al. \cite{xu2013differentially} investigate the similar problem of releasing histograms,
whereas Ding et al. \cite{ding2011differentially} work with data cubes.
Other synopses have also been used, like spatial decompositions (Cormode et al. \cite{cormode2012differentially}), and wavelet transforms, which allow to more accurately answer range queries (Xiao et al. \cite{xiao2010differential}).
Cormode et al. \cite{cormode2012differentially} examine more sophisticated summarization techniques, based on sampling, filtering and sketching.

The first connection between differential privacy and \emph{probabilistic inference} is due to Williams et al. \cite{williams2010probabilistic};
they apply probabilistic inference to the noisy data, and, taking into account that the perturbation process is known, they attempt to estimate the parameters of the model that generated the data.
Dimitrakakis et al. \cite{dimitrakakis2014robust} examine the connections between differential privacy and Bayesian inference by introducing a differentially private mechanism based on posterior sampling.
Several subsequent efforts build upon their ideas (e.g. Zhang et al. \cite{zhang2016differential}, Foulds et al. \cite{foulds2016theory}, Bernstein et al. \cite{bernstein2017differentially}).

The state of the art solution in data publishing with differential privacy is \emph{PrivBayes}, introduced by Zhang et al. \cite{zhang2014privbayes}, \cite{zhang2017privbayes}.
The authors identify that the main problem in publishing high-dimensional data (that consist of a relatively large number of attributes $N=|\mathcal{A}|$) with differential privacy
is that the perturbation required inevitably overlaps the signal in the data.
By high-dimensional, we refer to data whose domain size $|d_{\mathcal{A}}|$ is comparable with the total number of data points $n$.
The proposed solution, namely PrivBayes, is inspired from the theory of probabilistic graphical models,
and is based on learning the Bayesian Network (directed graphical model) that best fits the data, while satisfying differential privacy.
The learned Bayesian Network provides an approximation of the high-dimensional data distribution as a product of low-order conditional and marginal distributions;
these low-order distributions contain much more compact signal that is not severely damaged by the required perturbation.
Finally, a synthetic dataset is published by sampling tuples (data points) from the approximate distribution.
As we already noted in the introduction of this chapter, our solution is also based on this idea, and is strongly inspired by PrivBayes.

Following PrivBayes, Chen et al. \cite{chen2015differentially} develop a sampling-based framework to explore the dependencies among all attributes and build a dependency graph, and then approximate the data distribution based on the junction tree algorithm.
However, the version of the sparse vector technique they use to satisfy differential privacy is shown to be flawed.
Ping et al. \cite{ping2017datasynthesizer} implement PrivBayes as part of their proposed DataSynthesizer, a software tool that takes a sensitive dataset as input and outputs a synthetic, statistically similar dataset.

\subsection{Our Approach: Distributed Data Publishing with Differential Privacy}
We aim to combine the two approaches, in that:
\begin{itemize}
\item[-] we jointly learn a global, differentially private model (specifically a Bayesian Network) utilizing the entire dataset, and
\item[-] we publish either the model itself, or a dataset generated from the model, so that it can be used for arbitrary analyses.
\end{itemize}
Thus, our approach can be characterized as a \emph{distributed data publishing with differential privacy} approach.
One of the solutions that we offer is based on a purely distributed algorithm, in that the data holders incrementally respond to the analyst's queries during the execution of the algorithm, and a single (global) model is learned in a distributed fashion.
The other two solutions are hybrid, as the information that each data holder shares is a synopsis/model of its local dataset, which the analyst utilizes to learn a global model.

In a recent work, Su et al. \cite{su2016differentially} also develop a distributed version of PrivBayes, in order to privately learn (at once) a (global) Bayesian Network from distributed data, and then use it to publish a synthetic dataset.
However, their solution does not meet the requirements we have posed, in two ways.
Firstly, they assume the existence of a semi-trusted curator, an intermediate entity that assists the data holders to collectively learn the global model.
In particular, the data holders and the curator collaboratively identify the Bayesian Network that best fits the integrated dataset $D$ in a sequential
manner;
their key contribution is in the construction of the search frontier, which consists of the set of candidate edges to add to the Bayesian Network in the next update step.
Secondly, their solution is based on the distributed (multi-party) version of the Laplace mechanism (Pathak et al. \cite{pathak2010multiparty}), which requires the use of cryptography.

\section{Bayesian Networks}
In this section, we provide the necessary background for Bayesian Networks, which constitute the basic building block of our solution.

\subsection{Basic Definitions}
A Bayesian Network (Pearl \cite{pearl1986fusion}) is a probabilistic graphical model, and in particular a directed graphical model, that defines a family of joint probability distributions over a set of attributes/random variables.
More formally, we give the following definitions, following Koller and Friedman \cite{koller2009probabilistic}

\begin{defn}[Bayesian Network structure] \label{def_BN_structure}
A Bayesian Network structure $\mathcal{G}$ is a directed acyclic graph whose nodes represent attributes $X_1,...,X_N$ from a set $\mathcal{A}$.
Let $\mathbf{Pa}(X_i)$ denote the parents of $X_i$ in $\mathcal{G}$ and $\mathbf{NonDesc}(X_i)$ denote the set of attributes that are not descendants of $X_i$ in $\mathcal{G}$.
Then $\mathcal{G}$ encodes the following set of conditional independence assumptions:
$$ ( \ X_i \perp \mathbf{NonDesc}(X_i) \ ) \ | \ \mathbf{Pa}(X_i) \ , \ \ \forall i \in \{1,...,N\}$$
\end{defn}

\begin{figure}%[H]%[ht!]
    \centering
    \includegraphics[scale=0.5]{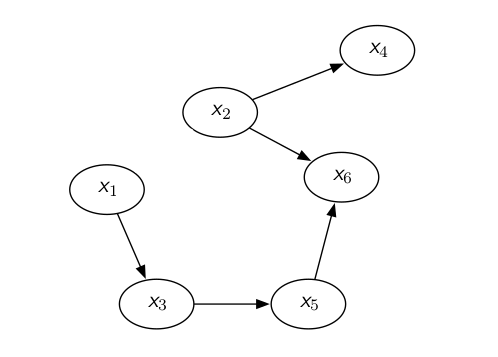}
    \caption{Example Bayesian Network structure}
	\label{FIG_example_bn}
\end{figure}

An example Bayesian network structure is illustrated in Figure \ref{FIG_example_bn}.
Using this example, we identify the types of connections between nodes, and the (in)dependence relations they imply for the corresponding attributes:
\begin{itemize}
\item[-] $X_1 \rightarrow X_3$: direct dependence, $X_3 \not \perp X_1$.
\item[-] $X_1 \rightarrow X_3 \rightarrow X_5$: indirect causal effect, $X_5 \not \perp X_1$, but $(X_5 \perp X_1) | X_3$. 
\item[-] $X_5 \leftarrow X_3 \leftarrow X_1$: indirect evidential effect, $X_1 \not \perp X_5$, but $(X_1 \perp X_5) | X_3$.
\item[-] $X_6 \leftarrow X_2 \rightarrow X_4$: common cause, $X_4 \not \perp X_6$, but $(X_4 \perp X_6) | X_2$.
\item[-] $X_5 \rightarrow X_6 \leftarrow X_2$: common effect, $X_5 \perp X_2$, but $(X_5 \not \perp X_2) | X_6$.
\end{itemize} 

\begin{defn}[Factorization] \label{def_BN_factor}
Let $\mathcal{G}$ be a Bayesian Network structure over a set of attributes $\mathcal{A}=\{X_1,...,X_N\}$.
The distribution $\mathbf{p}_{\mathcal{A}}$ factorizes according to $\mathcal{G}$ if $\mathbf{p}_{\mathcal{A}}$ can be expressed as a product:
$$ \mathbf{p}_{\mathcal{A}} = \prod_{i=1}^N \mathbf{p}_{X_i | \mathbf{Pa}(X_i)} $$
Each individual factor $\mathbf{p}_{X_i | \mathbf{Pa}(X_i)}$ represents the set of all conditional probability distributions of $X_i$, one for each realization of the attributes in $\mathbf{Pa}(X_i)$,
and is called a conditional probability table. \\
We denote the set of all conditional probability tables by: $\Theta = \{ \Theta_1 , ... , \Theta_N \}$, where $\Theta_i = \mathbf{p}_{X_i | \mathbf{Pa}(X_i)}, \ \forall i \in \{1,...,N\}$.
\end{defn}

\begin{defn}[Bayesian Network] \label{def_BN}
A Bayesian Network over a set of attributes $\mathcal{A}$ is a pair $\mathcal{B}=(\mathcal{G},\Theta)$, where $\mathbf{p}_{\mathcal{A}}$ factorizes according to $\mathcal{G}$ and is specified based on the conditional probability distributions in $\Theta$.
\end{defn}

\begin{defn}[Bayesian Network degree] \label{def_BN_degree}
Let $\mathcal{B}=(\mathcal{G},\Theta)$ be a Bayesian Network over a set of attributes $\mathcal{A}=\{X_1,...,X_N\}$.
We define the degree $k$ of $\mathcal{B}$ as the in-degree $\mathcal{G}$:
$$ k = \max_{i \in \{1,...,N\}} |\mathbf{Pa}(X_i)|$$
\end{defn}

Based on Definitions \ref{def_BN_structure}, \ref{def_BN_factor}, \ref{def_BN}, we see that \emph{any} joint probability distribution over a set of attributes $\mathcal{A}$ can be represented using a Bayesian Network (by the \emph{chain rule}).
In practice, we are particularly interested in distributions that factorize according to \emph{sparse graphs}, that lead to compact representations and efficient inference. \\
\textbf{Example.} Consider a probability distribution $\mathbf{p}$ over $N$ attributes $\{X_1,...,X_N\}$, such that $\text{domain}(X_i) = d,\ \forall i \in \{1,...,N\}$.
Then, $\mathbf{p}$ is originally represented using $d^N$ parameters, each corresponding to the probability of a specific realization of the random vector $\mathbf{X}$.
A Bayesian Network allows us to represent $\mathbf{p}$ using only
$$ \sum_{i=1}^N |\Theta_i| = \sum_{i=1}^N d^{|\mathbf{Pa}(X_i)|+1} = \mathcal{O}(N d^k ) $$
parameters, which can be substantially smaller (usually $k<<N$).

The time complexity of both making \emph{inferences} using $\mathbf{p}$, and \emph{sampling} from $\mathbf{p}$, can significantly decrease.
More specifically,
\begin{itemize}
\item[-] By inference, we basically refer to two tasks.
Assume that we observe a subset $\mathcal{A}_1$ of the attributes in $\mathcal{A}$, and let $\mathcal{A}_2 = \mathcal{A} \setminus \mathcal{A}_1$.
The first task is the calculation of posterior probabilities, namely $\mathbf{p}_{\mathcal{A}_2|\mathcal{A}_1}$.
The second task is the calculation of the most probable configuration (realization) for the unobserved attributes, namely $\text{argmax}_{\mathbf{x_{\mathcal{A}_2}} \in d_{\mathcal{A}_2}} \mathbf{p}_{\mathcal{A}_2|\mathcal{A}_1}$.
As an example, in case the unobserved attributes correspond to a set of class attributes,
calculating the most probable configuration provides a solution for the corresponding classification problem.
\item[-] As far as sampling is concerned, efficient algorithms exist, and probably the simplest among them is \emph{prior sampling} (which we utilize throughout this work).
Instead of sampling from the $N$-dimensional joint distribution $\mathbf{p}$,
we separately draw a sample for each attribute $X_i \ (\forall i \in \{ 1,...,N \})$ from the proper conditional distribution in $\Theta_i$.
To be able to do so, we need to ensure that, by the time we draw a sample for $X_i$, we have already sampled for all attributes in $\mathbf{Pa}(X_i)$;
this is achieved by topologically ordering the attributes based on $\mathcal{G}$.
\end{itemize}

The power of Bayesian Networks is twofold.
Besides the \emph{computational gain} they offer, by allowing us to approximate a high-dimensional distribution as a product of low-order conditional and marginal distributions,
Bayesian Networks are also highly \emph{interpretable} models,
and can hence be used for knowledge discovery.

\subsection{Learning Bayesian Networks} \label{section_learning_BNs}
In real-world applications, we only have access to data, which we assume that follow a distribution $\mathbf{p}$ over a set $\mathcal{A}=\{X_1,...,X_N\}$ of attributes, that can be encoded by a Bayesian Network.
Sometimes the Bayesian Network structure is known; for example, it may be given by an expert, or it may be determined by the physical properties of the application.
However, our focus is on the case that the structure is unknown, and our objective is to find both the structure and the parameters of the Bayesian Network that best fits our dataset $D$.
As we already noted in the introduction of this chapter, we make the assumption that our data are fully-observed, in the sense that no hidden variables exist.

In general, there are three approaches to learning Bayesian Networks of unknown structure from data, namely:
\begin{itemize}
\item[-] score-based, where the Bayesian Network learning problem is viewed as a model selection problem and is solved using optimization methods,
\item[-] constraint-based, where the Bayesian Network is viewed as a representation of independences and the learning process is based on independence tests, and
\item[-] Bayesian model averaging, where an ensemble of possible structures is learned and aggregated.
\end{itemize}
We focus on the score-based approach, which we adopt in our work.

The first step in \emph{score-based Bayesian Network learning} is to assign each possible Bayesian Network $\mathcal{B}$ a score.
A natural choice for \emph{scoring function} is the likelihood function $\mathcal{L}$, which measures the probability of observing the data in $D$ (assuming that individual data points are i.i.d.), given a model.
In our case, the model is a Bayesian Network $\mathcal{B}=(\mathcal{G},\Theta)$, 
so, denoting by $\mathbf{p}_{\mathcal{B}}$ the distribution encoded by $\mathcal{B}$, we conclude that the score of $\mathcal{B}$ is:
$$ \mathcal{L}(\mathcal{B} ; D) 
=  \prod_{\mathbf{x} \in D} \mathbf{p}_{\mathcal{B}} (\mathbf{x})
=  \prod_{\mathbf{x} \in D} \prod_{i=1}^N \mathbf{p}_{X_i|\mathbf{Pa}(X_i)} (x_i,\mathbf{x}_{\mathbf{Pa}(X_i)})
=  \prod_{i=1}^N \prod_{\mathbf{x} \in D} \Theta_i (x_i,\mathbf{x}_{\mathbf{Pa}(X_i)})
$$
which illustrates the decomposability of the global likelihood into local likelihoods (one for each parameter $\Theta_i$), based on the Bayesian Network structure.
If we instead use the logarithm of $\mathcal{L}$ as scoring function, it turns out that:
$$ \ell (\mathcal{B} ; D) 
= \log \mathcal{L}(\mathcal{B} ; D) 
= n \sum_{i=1}^N \hat{I}(X_i;\mathbf{Pa}(X_i)) - n  \sum_{i=1}^N \hat{H}(X_i)
$$
where by $\hat{H}$ and $\hat{I}$ we refer to the empirical entropy and mutual information respectively.

The next step is to find the Bayesian Network structure $\mathcal{G}$ that achieves the highest score.
Noting that the only term in the log-likelihood score that depends on $\mathcal{G}$ is the empirical mutual information, gives that:
$$ \max_{\mathcal{B}} \ \ell (\mathcal{B} ; D) 
= \max_{\mathcal{G}} \underbrace{ \max_{\Theta} \ \ell (\mathcal{G},\Theta ; D)}_{\text{MLE of } \Theta \text{ given } \mathcal{G}} 
= \max_{\mathcal{G}} \ \ell (\mathcal{G},\hat{\Theta} ; D)
= \max_{\mathcal{G}} \sum_{i=1}^N \hat{I}(X_i;\mathbf{Pa}(X_i))$$
which demonstrates that the structure $\mathcal{G}^*$ that maximizes the log-likelihood score is the one that has the maximum sum over all attributes, of the mutual information between each attribute and its parents in $\mathcal{G}^*$,
when the maximum likelihood parameters $\hat{\Theta}$ are used for $\mathcal{G}^*$.
From an information theoretic point of view, it can be shown that $\mathcal{G}^*$ minimizes the Kullback-Leibler divergence between the actual data distribution, and the distribution encoded by the Bayesian Network structure $\mathcal{G}^*$.

Solving the aforementioned optimization problem turns out to be hard in general.
Specifically, taking into account that our graph consists of $N$ nodes,
there are $\mathcal{O}(2^{N^2})$ potential structures that form our search space (super-exponential in the number of attributes).
Consequently, heuristic (local-search) algorithms are employed in practice, like hill-climbing and simulated annealing.
Nevertheless, for the special case that $k=1$ (Bayesian Networks with degree 1), the well-known Chow-Liu algorithm \cite{chow1968approximating} (Algorithm \ref{Chow-Liu Algorithm}) allows us to greedily find the optimal structure.

\begin{algorithm}%[H]
\DontPrintSemicolon

\KwIn{Dataset $D$}

\KwOut{BN structure $\mathcal{G}$}

Initialize $\mathcal{G}_0$ to a fully connected (undirected) graph \;

\For{i=1\ \KwTo\ N}{

	Estimate (and store) $\mathbf{\hat{p}}_{X_i}$ from $D$ \;

	\For{j=1\ \KwTo\ i-1}{
	
		Estimate (and store) $\mathbf{\hat{p}}_{X_j,X_i}$ from $D$ \;
	
		Compute weight $\hat{I}(X_j;X_i)$ of edge $(X_j,X_i)$ in $\mathcal{G}_0$ \;
	} 
}
	
$\mathcal{G} = \text{MaximumWeightSpanningTree}(\mathcal{G}_0)$ \;
	
Give directions to edges in $\mathcal{G}$ \;
	
Return $\mathcal{G}$ \;

\caption{Chow-Liu Algorithm\label{Chow-Liu Algorithm}}
\end{algorithm}

An important limitation of the likelihood score is that it favors more complex structures over simpler ones.
In fact, if we do not artificially constrain the number of parents allowed for each attribute (as in the case of the Chow-Liu algorithm), 
then any algorithm using the likelihood score will almost always return a fully connected structure.
This results from the fact that adding an additional parent to any attribute will almost always increase the score, since for any random variables $X,Y,Z$, we have $\hat{I}(X; Y, Z) \geq \hat{I}(X, Y )$ (information never hurts).
Therefore, in practice, we either learn fixed-degree Bayesian Networks (this is the case in PrivBayes), or we penalize more complex structures depending on the dataset size (e.g. BIC score).

Once the Bayesian Network structure $\mathcal{G}$ is known, we are left with a simple parameter estimation task; we have to estimate the parameters in $\Theta$.
A common approach to achieve this is to (once again) use the maximum likelihood principle.
The decomposability of the likelihood function, which we demonstrated earlier,
allows us to maximize each local likelihood $\mathcal{L}_i(\Theta_i ; D) = \prod_{\mathbf{x} \in D} \Theta_i (x_i,\mathbf{x}_{\mathbf{Pa}(X_i)})$ separately,
and then combine the solutions to get the (global) maximum likelihood estimate $\hat{\Theta}$ for $\Theta$:
$$\hat{\Theta} = \{\hat{\Theta}_1,...,\hat{\Theta}_N \}, \text{ where } \hat{\Theta}_i=\text{argmax}_{\Theta_i} \mathcal{L}_i(\Theta_i ; D)$$

We described a purely frequentist approach to learning Bayesian Networks from data, in that we utilize the likelihood score to learn the structure and then estimate the parameters that maximize the likelihood function.
An alternative path would be to adopt a Bayesian approach, and treat the candidate structures (in the structure learning phase) or the parameters (in the parameter learning phase) as random;
we would assign them prior distributions, and we would maximize the posterior distribution of the data, instead of the likelihood.
We do not further discuss the Bayesian approach here,
but we mention that its main advantage over the frequentist approach is that it avoids overfitting.

%\subsection{Distributed Machine Learning}
%We summarize some common ideas and methods for distributed machine learning.
%For a much more detailed presentation, we refer the interested reader to Barral et al. [cit], and the references therein.
%
%survey for a detailed analysis of the most popular methods in distributed machine learning, focus on classification.
%
%combine models vs combine outputs.
%
%1 decision rules: soft output (posterior probabilities of assigning a data point to a particular class) then product rule, max rule, etc.
%VS hard output then majority voting 
%
%2 stacked generalization: train local classifiers using a subset of the data (training set),
%then train a global classifier that takes as input the outputs of local classifiers on another subset of the data (meta-level training set), instead of combining them using fixed rules
%
%3 meta-learning: inspired by stacked generalization, meta-level training set is formed in more sophisticated ways
%
%4 knowledge probing: interpretability of meta-learning, 
%
%\subsection{Learning Bayesian Networks from Distributed Data}
%
%few authors address this problem

\section{PrivBayes}
In this section, we provide an overview of PrivBayes.
We also highlight the changes that need to be performed in our model, as well as the assumptions we make.
Apparently, a first major change originates from the fact that PrivBayes addresses the traditional differential privacy model, where the data holder and the analyst are the same, trusted entity.
Therefore, the original solution depends on a centralized algorithm (the single dataset $D$ is analyzed by a trusted entity),
and the only requirement is that the resulting Bayesian Network satisfies differential privacy (both the structure and the parameters).

PrivBayes consists of three phases, which we briefly present here in order to be able argue about the privacy guarantees of the overall algorithm.
In the next subsections, we present each of the first two phases separately;
the third phase is trivial, so we fully develop it here.
\begin{itemize}
\item[-] \textbf{Structure learning phase.}
During this phase, the analyst accesses the sensitive data in $D$.
It uses an $\eps_1$-differentially private algorithm to learn the structure of a $k$-degree Bayesian Network that accurately encodes the conditional independencies that are present in the underlying data distribution.

\item[-] \textbf{Parameter learning phase.}
During this phase, the analyst utilizes the learned structure, and again accesses the sensitive data in $D$.
It uses an $\eps_2$-differentially private algorithm to estimate the parameters of the learned $k$-degree Bayesian Network.

\item[-] \textbf{Synthetic data generation phase.}
During this phase, no access to the sensitive data is performed.
The analyst utilizes the learned model to sample $n'$ tuples (data points) from the approximate distribution (encoded by the Bayesian Network).
The sampling step is performed using prior sampling,
and does not access the sensitive data - only the output of the first two phases is used.
Since (as we argue in theorem \ref{thm_privbayes_overall}) the output of the first two phases satisfies differential privacy, according to the post-processing theorem, so does the result of the third phase,
and no additional perturbation is required. \\
We note that the value of $n'$ depends on the application, but a reasonable choice would be to set $n'=n$.
Also, in case we are interested in the model, and not in publishing a synthetic dataset, the data generation phase may even be skipped.
\end{itemize}

\begin{thm} \label{thm_privbayes_overall}
Let $\eps_1$ and $\eps_2$ be the privacy budget consumed during each of the first two phases of PrivBayes respectively.
Then, the overall algorithm satisfies $(\eps_1+\eps_2)$-differential privacy.
\end{thm}
Theorem \ref{thm_privbayes_overall} directly follows from the composition theorem of differential privacy.
The choice of $\eps_1$ and $\eps_2$ determines the balance between the quality of the learned structure and the learned parameters.
Assuming that the total available privacy budget is $\eps$, and by setting $\eps_1=\beta \eps$ and $\eps_1=(1-\beta) \eps$,
the authors of PrivBayes experimentally conclude that the optimal split is achieved for some $\beta \in [0.2,0.5]$. 
For simplicity, in our work we assume that, whenever the privacy budget has to be split between the two phases, $\eps_1=\eps_2=\frac{\eps}{2}$,
so the overall algorithm satisfies $\eps$-differential privacy.

\subsection{Structure Learning Phase}
The structure learning phase of PrivBayes is based on a greedy extension of the Chow-Liu algorithm (Algorithm \ref{Chow-Liu Algorithm}) for higher degree Bayesian Networks, which we present in Algorithm \ref{Greedy BN Structure Learning}.
The authors of PrivBayes propose the aforementioned approach, instead of the local-search algorithms that are commonly used for learning the structure of higher degree Bayesian Networks,
because these algorithms perform too many data accesses and incur a high cost in terms of sensitivity;
as a result, a prohibitive amount of noise is added.

\begin{algorithm}%[H]
\DontPrintSemicolon
\KwIn{Dataset $D$, BN degree $k$, Privacy budget $\eps_1$}

\KwOut{BN structure $\mathcal{G}$}

Initialize $\mathcal{G}=\emptyset$ and $V = \emptyset$ \;

\For{each $\mathcal{A'} \subseteq \mathcal{A}$ such that $|\mathcal{A'}| \leq k+1$}{

	Estimate (and store) $\mathbf{\hat{p}}_{\mathcal{A'}}$ from $D$ \;

}

Arbitrarily select an attribute $X$ from $\mathcal{A}$ \; 
Add $(X,\emptyset)$ to $\mathcal{G}$ and $X$ to $V$ \;

\For{i=1 \KwTo N-1}{

	Initialize $\Omega=\emptyset$ \;
	
	\For{each $X \in \mathcal{A} \setminus V$}{
	
		\For{each $\mathbf{Pa}(X) \subseteq V$ such that $|\mathbf{Pa}(X)| \leq k$}{
	
			Compute $\hat{I}(X;\mathbf{Pa}(X))$ using $\mathbf{\hat{p}}_{X}$, $\mathbf{\hat{p}}_{\mathbf{Pa}(X)}$, and $\mathbf{\hat{p}}_{X,\mathbf{Pa}(X)}$ \;
			
			Add $(X,\ \mathbf{Pa}(X),\ \hat{I}(X;\mathbf{Pa}(X)))$ to $\Omega$ \;
		
		}
	}
	Sample a tuple {\scriptsize $(X,\mathbf{Pa}(X),\hat{I}(X;\mathbf{Pa}(X)))$} from $\Omega$ w/ prob. $\propto$ $\frac{ \hat{I}(X;\mathbf{Pa}(X)) \eps_1 }{ 2 (N-1) \Delta \hat{I} }$ \; \label{Greedy BN Structure Learning_expo}
	
	Add $(X,\mathbf{Pa}(X))$ to $\mathcal{G}$ and $X$ to $V$\;
}
Return $\mathcal{G} $

\caption{Greedy BN Structure Learning\label{Greedy BN Structure Learning}}
\end{algorithm}

Lemma \ref{lemma_sensitivity_mutual_info} quantifies the sensitivity of the empirical mutual information; the proof can be found in the full PrivBayes paper \cite{zhang2017privbayes}.
We remark that the sensitivity is computed using the definition of adjacent datasets that leads to \emph{bounded} differential privacy,
so $\Delta \hat{I}$ expresses the maximum change in the empirical mutual information that is caused by changing the value of one data point.
\begin{lemma} \label{lemma_sensitivity_mutual_info}
For any random variables $X$ and $Y$,
the sensitivity of $\hat{I}(X;Y)$ is:
\[ \Delta \hat{I} = \left\{
\begin{array}{ll}
      \frac{1}{n} \log n + \frac{n-1}{n} \log \frac{n}{n-1} & \text{if } X \text{ or } Y \text{ is binary}\\
      \frac{2}{n} \log \frac{n+1}{2} + \frac{n-1}{n} \log \frac{n+1}{n-1} & \text{otherwise}
\end{array}
\right.
\]
\end{lemma}

The output $\mathcal{G}$ of Algorithm \ref{Greedy BN Structure Learning} consists of the $N-1$ attribute-parent pairs
each of which was added using the exponential mechanism (line \ref{Greedy BN Structure Learning_expo}) with the empirical mutual information as scoring function and with privacy budget $\frac{\eps_1}{N-1}$.
Thus, Theorem \ref{thm_privbayes_struct} directly follows (by the composition theorem).
\begin{thm} \label{thm_privbayes_struct}
Algorithm \ref{Greedy BN Structure Learning} satisfies $\eps_1$-differential privacy.
\end{thm}

\textbf{How to pick $k$.}
Based on our earlier discussion on Bayesian Networks, it is not hard to notice that, excluding the first $k$ attributes that are inserted in $\mathcal{G}$,
all other $N-k$ attributes will have exactly $k$ parents (see Section \ref{section_learning_BNs} - information never hurts).
A natural question is how to pick $k$ in practice.
To answer this question, the authors of PrivBayes introduce the notion of $\theta$-\emph{usefulness}, which enables PrivBayes to automatically select $k$.
In particular, the user-specified parameter $\theta$ expresses the minimum-allowed ratio of average scale of signal to average scale of noise
in each entry in the conditional probability tables of the resulting $k$-degree Bayesian Network.
So, for example, when the privacy budget is small, PrivBayes will favor smaller values for $k$, so that the dimension of the resulting conditional probability distributions is small enough and the signal contained is not dominated by the noise.
Although this notion could apply to our model as well, we focus on different aspects of PrivBayes, and, hence, assume that $k$ is known in advance. 

\textbf{Improving the scoring function.}
A key contribution of PrivBayes is that the authors identify a significant limitation in using the empirical mutual information as scoring function for the exponential mechanism.
Specifically, although $\lim_{n \rightarrow \infty} \Delta \hat{I} = 0$ (for both binary and non-binary domains), $\Delta \hat{I} > \frac{\log n}{n}$,
so the empirical mutual information can be quite large compared to the range of $\hat{I}$,
and, as a result, the exponential mechanism cannot distinguish between high-scoring and low-scoring attribute-parent pairs.
Recall that the probability of sampling a particular pair is $\propto \frac{ \hat{I}(X;\mathbf{Pa}(X)) \eps_1 }{ 2 (N-1) \Delta \hat{I} }$, so when $\hat{I}(X;\mathbf{Pa}(X))$ and $\Delta \hat{I}$ are comparable for many candidate pairs $(X;\mathbf{Pa}(X))$, their sampling probabilities tend to become uniform.
PrivBayes tackles this limitation by introducing two alternative scoring functions,
that have small sensitivities compared to their ranges, but, at the same time, their behavior is similar with that of the empirical mutual information.
\begin{itemize}
\item[-] The first, namely $F$, relies on the notion of maximum joint distributions and has sensitivity $\Delta F = \frac{1}{n}$. It can be efficiently computed only for binary domains.
\item[-] The second, namely $R$, relies on the $L_1$ distance between $\mathbf{\hat{p}}_{X,\mathbf{Pa}(X)}$ and $\mathbf{\hat{p}}_{X} \mathbf{\hat{p}}_{\mathbf{Pa}(X_)}$ (the distribution that minimizes the empirical mutual information), and has sensitivity $\Delta R = \frac{3}{n} + \frac{2}{n^2}$. It can be efficiently computed for arbitrary domains.
\end{itemize}
This optimization directly applies to all our proposed solutions.
Nevertheless, we choose to maintain the empirical mutual information as scoring function (in all the solutions we examine), because, firstly, it is a well-understood function, and secondly, it facilitates our theoretical analysis.

\subsection{Parameter Learning Phase}
The parameter learning phase of PrivBayes also follows the frequentist approach we described in Section \ref{section_learning_BNs}.

\begin{algorithm}%[H]
\DontPrintSemicolon

\KwIn{Dataset $D$, BN structure $\mathcal{G}$, Privacy budget $\eps_2$}

\KwResult{BN parameters $\Theta$}

Initialize $\Theta_i=\emptyset, \ \forall \ i \in \{1,...,N\}$ \;

\For{$i=1\ \KwTo\ N$}{

	Estimate $\mathbf{\hat{p}}_{X_i,\mathbf{Pa}(X_i)}$ \;
	
	Compute noisy $\mathbf{\tilde{p}}_{X_i,\mathbf{Pa}(X_i)} = \mathbf{\hat{p}}_{X_i,\mathbf{Pa}(X_i)} + \text{Laplace}(0,\frac{2N}{n\eps_2})$ \; \label{BN Param Learning laplace}
	
	Set negative values in $\mathbf{\tilde{p}}_{X_i,\mathbf{Pa}(X_i)}$ to 0 and normalize \;
	
	Compute $\tilde{\Theta}_i = \mathbf{\tilde{p}}_{X_i|\mathbf{Pa}(X_i)}$ (by marginalizing $\mathbf{\tilde{p}}_{X_i,\mathbf{Pa}(X_i)}$) \;
	
	Add $\tilde{\Theta}_i$ to $\tilde{\Theta}$ \;
}

Return $\tilde{\Theta}$ \;

\caption{BN Parameter Learning\label{BN Parameter Learning}}
\end{algorithm}

The output $\tilde{\Theta}$ of Algorithm \ref{BN Parameter Learning} consists of the $N$ conditional probability tables $\tilde{\Theta}_i$, $i \in \{1,...,N\}$,
each of which represents the set of conditional distributions of an attribute, given all realizations of its parents.
Notice that each $\tilde{\Theta}_i$ is constructed using a noisy version of the maximum likelihood estimate of the joint probability distribution of attribute $X_i$ and its parents.
The sensitivity of the maximum likelihood estimate (using $n$ data points) is $\Delta \mathbf{\hat{p}} = \frac{2}{n}$;
the proof is almost identical with that of Theorem \ref{thm_noisy_ss_priv}, where we argue about the sensitivity of the frequency distribution over a set of attributes (which is the basis of the maximum likelihood estimate).
In total, we estimate $N$ noisy distributions using the Laplace mechanism (line \ref{BN Param Learning laplace}), each with privacy budget $\frac{\eps_2}{N}$,
so (again) by the composition theorem:
\begin{thm} \label{thm_privbayes_param}
Algorithm \ref{BN Parameter Learning} satisfies $\eps_2$-differential privacy.
\end{thm}

\textbf{Consistency.}
We remark that, although the distributions in $\tilde{\Theta}$ have no consistency issues (since each of them is the conditional distribution of a different attribute, given a different realization of its parents),
the intermediate joint distributions $\mathbf{\tilde{p}}$ computed by Algorithm \ref{BN Parameter Learning} are (generally) not consistent.
Specifically, an arbitrary attribute $X$ is likely to be involved in several joint distributions, and we have no guarantees that the marginals of $X$ derived using different joint distributions will be the same.
Following PrivBayes, we also ignore these consistency issues, as they have no impact on the final result.

\subsection{PrivBayes in the Distributed Model}
As we already argued, PrivBayes follows a score-based approach to learning the target Bayesian Network,
and, in particular, PrivBayes finds a greedy solution to the optimization problem which we described in Section \ref{section_learning_BNs}.
We re-formulate this optimization problem, taking into account that in our case, the data are distributed among $M$ data holders.
\begin{eqnarray*}
\max_{\mathcal{B}} \ell (\mathcal{B} ; D_1,...,D_M)
& = & \max_{\mathcal{G}} \sum_{i=1}^N \hat{I}(X_i;\mathbf{Pa}(X_i)) \\
& = &\max_{\mathcal{G}} \ 
\sum_{i=1}^{N} \sum_{
\substack{x \in  d_{X_i} \\
\mathbf{p} \in d_{\mathbf{Pa}(X_i)}}} \ \bigg( \ 
\frac{\sum_{j=1}^M{\mathbf{c}_{X_i,\mathbf{Pa}(X_i)}^{(j)}(x,\mathbf{p})}}{n} \\
&& \ \ \ \ \ \times\ \log_2 
\frac{n \sum_{j=1}^M
{\mathbf{c}_{X_i,\mathbf{Pa}(X_i)}^{(j)}(x,\mathbf{p})}}
{\sum_{j=1}^M{\mathbf{c}_{X_i}^{(j)}(x)} \sum_{j=1}^M{\mathbf{c}_{\mathbf{Pa}(X_i)}^{(j)}(\mathbf{p})}} \ \bigg)
\end{eqnarray*}
The main observation is that the scoring function which we optimize is \emph{non-linear} with respect to the frequency distributions $\mathbf{c}$,
and, consequently, we cannot compose the global value of the scoring function from its local values.
Although here we examine the empirical mutual information, our argument is true for all scoring functions introduced in PrivBayes.
Notice that the \emph{sufficient statistic} to estimate the empirical mutual information between an attribute and its parents is their joint frequency distribution,
and, hence, to solve our optimization problem we need to have all the $(k+1)$-dimensional frequency distributions (where $k$ is the degree of the target Bayesian Network).

The key question that needs to be answered is what information each data owner shares with the (untrusted) analyst;
we refer to this question as the \emph{What to share?} question. 
Furthermore, depending on the answer to this question, we need to figure out how to combine the information shared by different data holders.
In the next sections, we examine different answers to the \emph{What to share?} question. \\

Once the Bayesian Network structure $\mathcal{G}$ is known, we again follow a frequentist approach and adjust Algorithm \ref{BN Parameter Learning} to the distributed model.
As we will see, depending on the structure learning approach used,
the analyst may need or need not re-access the data to estimate the Bayesian Network parameters.
The latter may be the case if the analyst has already retrieved the required local distributions, or is (somehow) able to estimate them.
Therefore, we introduce the boolean parameter $retrieved$ that indicates whether the analyst already possesses the distributions $\mathbf{\tilde{p}}_{X_i,\mathbf{Pa}(X_i)}, \ \forall i \in \{1,...,N\}$.

\begin{algorithm}%[H]
\DontPrintSemicolon

\KwIn{Datasets $D_1,...,D_M$, BN structure $\mathcal{G}$, Boolean $retrieved$ }

\KwResult{BN parameters $\Theta$}

Initialize $\Theta_i=\emptyset, \ \forall \ i \in \{1,...,N\}$ \;

\For{$i=1\ \KwTo\ N$}{

	\If{ $retrieved == \text{False}$ }{

		\For{$j=1\ \KwTo\ M$}{
			\textbf{QUERY}($D_j$): retrieve local $\mathbf{\tilde{c}}^{(j)}_{X_i,\mathbf{Pa}(X_i)}$ \;
		}
		Estimate global $\mathbf{\tilde{p}}_{X_i,\mathbf{Pa}(X_i)} = \frac{1}{n} \sum_{j=1}^M{ \mathbf{\tilde{c}}_{X_i,\mathbf{Pa}(X_i)}^{(j)} }$
	
	}
	
	Compute $\tilde{\Theta}_i = \mathbf{\tilde{p}}_{X_i|\mathbf{Pa}(X_i)}$ (by marginalizing $\mathbf{\tilde{p}}_{X_i,\mathbf{Pa}(X_i)}$) \;
	
	Add $\tilde{\Theta}_i$ to $\tilde{\Theta}$ \;
}

Return $\tilde{\Theta}$ \;

\caption{Distributed BN Parameter Learning\label{Distributed BN Parameter Learning}}
\end{algorithm}

Notice that, in contrast to Algorithm \ref{BN Parameter Learning}, no perturbation is performed in Algorithm \ref{Distributed BN Parameter Learning}.
Since the analyst is not trusted, it is the data holders' responsibility to properly perturb their local frequency distributions, and handle any negative frequencies that may appear, prior to sharing them.
Although it is a straightforward application of the Laplace mechanism, we explain in detail how each data holder properly perturbs its local frequency distribution in Section \ref{noisy_ss_priv}.

\section{Sharing the Noisy Sufficient Statistics}
The first approach to answering the \emph{What to share?} question is based on asking each data holder to share its part of the sufficient statistics,
that is, all $(k+1)$-dimensional frequency distributions.
In doing so, the analyst is able to compose the global sufficient statistics (since the empirical frequency distribution is composable by simply summing the counts), and evaluate the scoring function for all candidate structures.
In that sense, this approach, which we call \emph{Sharing the Noisy Sufficient Statistics} (Algorithm \ref{Sharing the Noisy Sufficient Statistics}), is an exact approach.

\begin{algorithm}%[H]

\caption{Sharing the Noisy Sufficient Statistics\label{Sharing the Noisy Sufficient Statistics}}

\DontPrintSemicolon

\KwIn{Datasets $D_1,...,D_M$, BN degree k}

\KwOut{BN structure $\mathcal{G}$}

Initialize $\mathcal{G}=\emptyset$ and $V = \emptyset$ \;

\For{each $\mathcal{A'} \subseteq \mathcal{A}$ such that $|\mathcal{A'}|=k+1$}{

	\For{$j=1\ \KwTo\ M$}{
			\textbf{QUERY}($D_j$): retrieve local $\mathbf{\tilde{c}}^{(j)}_{\mathcal{A'}}$ \;
	}
	Estimate global $\mathbf{\tilde{p}}_{\mathcal{A'}} = \frac{1}{n} \sum_{j=1}^M{ \mathbf{\tilde{c}}_{\mathcal{A'}}^{(j)} }$
}

Arbitrarily select an attribute $X$ from $\mathcal{A}$ \;
Add $(X,\emptyset)$ to $\mathcal{G}$ and $X$ to $V$ \;

\For{i=1 \KwTo N-1}{

	Initialize $\Omega=\emptyset$ \;
	
	\For{each $X \in \mathcal{A} \setminus V$}{
	
		\For{each $\mathbf{Pa}(X) \subseteq V$ such that $|\mathbf{Pa}(X)| \leq k$}{
		
			Compute $\mathbf{\tilde{p}}_{X,\mathbf{Pa}(X)}$, $\mathbf{\tilde{p}}_{X}$, $\mathbf{\tilde{p}}_{\mathbf{Pa}(X)}$ by marginalizing the proper distributions, and then $\tilde{I}(X;\mathbf{Pa}(X))$ \;
	
			Add $(X,\ \mathbf{Pa}(X),\ \tilde{I}(X;\mathbf{Pa}(X)))$ to $\Omega$ \;
		
		}
	}
	Select $(X,\mathbf{Pa}(X))$ with the highest $\tilde{I}(X;\mathbf{Pa}(X))$ \;
	
	Add $(X,\mathbf{Pa}(X))$ to $\mathcal{G}$ and $X$ to $V$ \;
}
Return $\mathcal{G} $ \;
\end{algorithm}

In order to satisfy differential privacy, each data holder responds to the queries on its dataset using the Laplace mechanism.
We provide a more detailed analysis of the privacy guarantees of Algorithm \ref{Sharing the Noisy Sufficient Statistics} in Section \ref{noisy_ss_priv}.

Once the analyst collects the frequency distributions, it does not need to access the data again. 
This leads to the following two significant advantages:
\begin{itemize}
\item[-] First, with the (noisy) sufficient statistics to estimate the empirical mutual information at hand, any algorithm (e.g. local search) can be used to address the structure learning optimization problem.
Nevertheless, as shown in Algorithm \ref{Sharing the Noisy Sufficient Statistics}, we keep using the greedy extension of the Chow-Liu Algorithm (to fairly compare the approaches we examine).
\item[-] Second, once the structure learning phase is completed, the parameter learning Algorithm \ref{Distributed BN Parameter Learning} utilizes the already-retrieved distributions, and hence the entire privacy budget can be consumed in the structure learning phase.
\end{itemize}

On the downside, each data holder has to share ${N}\choose{k+1}$ frequency distributions, which may be prohibitive for high-degree Bayesian Networks, in terms of both perturbation and communication cost. 
To attack this problem, we could examine more sophisticated techniques in sharing the required frequency distributions, like the ones we presented in Section \ref{section data publishing dp}.
Nevertheless, as Xu et al. \cite{xu2013differentially} demonstrate, even the naive application of the Laplace mechanism performs sufficiently well compared to such techniques for a variety of datasets.

In the next two subsections, we provide a detailed privacy and accuracy analysis of Algorithm \ref{Sharing the Noisy Sufficient Statistics}.

\subsection{Privacy Analysis} \label{noisy_ss_priv}
We first examine the privacy guarantees of Algorithm \ref{Sharing the Noisy Sufficient Statistics}.
The following theorem quantifies the amount of noise each data holder must add to each frequency distribution it shares in order to preserve differential privacy.
Recall that the frequency distribution $\mathbf{c}_{\mathcal{A'}}$ of the attributes in $\mathcal{A'}$ can be viewed as a $d$-dimensional vector, where $d=|d_{\mathcal{A'}}|$.

\begin{thm} \label{thm_noisy_ss_priv}
Let $b = \frac{2{{N}\choose{k+1}}}{\eps}$.
If, $\forall \mathcal{A'} \subseteq \mathcal{A}$ such that $|\mathcal{A'}|=k+1$, each data holder shares $\mathbf{\tilde{c}}_{\mathcal{A'}} \ = \ \mathbf{c}_{\mathcal{A'}} \ + \ \boldsymbol{\eta}$,
where $\boldsymbol{\eta} = [\eta_1 \ \eta_2 \ ... \ \eta_d]^T$ is a random vector of i.i.d. $Laplace(0,b)$ entries,
then Algorithm \ref{Sharing the Noisy Sufficient Statistics} preserves $\eps$-differential privacy for any dataset $D_j \ (j \in \{1,...,M\})$.
\end{thm}

\begin{tcolorbox}[breakable]
\begin{proof}
Let $D,\ D'=D \cup \{\mathbf{x'}\} \setminus \{\mathbf{x}\}$ be two adjacent datasets of the same size, possessed by an arbitrary data holder.
Assume that $\mathbf{x}_{\mathcal{A'}}=d_{\mathcal{A'}}[j]$ for some $j \in \{1,...,d\}$, that is, the attributes $\mathcal{A'}$ in $\mathbf{x}$ take the $j$-th value from their joint domain $d_{\mathcal{A'}}$.
Accordingly, assume that $\mathbf{x'}_{\mathcal{A'}}=d_{\mathcal{A'}}[j']$ for some $j' \in \{1,...,d\}$.
The sensitivity of the frequency distribution $\mathbf{c}$ is:
\begin{eqnarray*}
\Delta \mathbf{c}
& = & \max_{D,D'} ||\mathbf{c}(D) - \mathbf{c}(D') ||_1 
\ = \ \max_{D,D'} \sum_{i=1}^d |c_i(D)-c_i(D')| \\
& = & |c_j(D) - (c_j(D) - 1)| + |c_{j'}(D) - (c_{j'}(D) + 1)|
\ = \ 2
\end{eqnarray*}
since removing $\mathbf{x}$ from the dataset will cause the count of $d_{\mathcal{A'}}[j]$ to decrease by one,
and adding $\mathbf{x'}$ to the dataset will cause the count of $d_{\mathcal{A'}}[j']$ to increase by one.
By simple application of the Laplace mechanism, it follows that adding i.i.d. $\text{Laplace}(0,\frac{\Delta \mathbf{c}}{\eps'})$ noise to each count will preserve $\eps'$-differential privacy. \\

The untrusted analyst interacts with the data only by viewing the underlying frequency distributions.
In total, each data holder shares ${N}\choose{k+1}$ frequency distributions, and to compute each of them, a new data access is required.
Therefore, by the composition theorem, if $\eps'=\frac{\eps}{ {{N}\choose{k+1}} }$, the overall algorithm preserves $\eps$-differential privacy.
\end{proof}
\end{tcolorbox}

\subsection{Accuracy Analysis}
To argue about the accuracy of Algorithm \ref{Sharing the Noisy Sufficient Statistics}, we examine in detail part of the pipeline it implements.

For each $\mathcal{A'} \subseteq \mathcal{A}$ such that $|\mathcal{A'}|=k+1$,
each data holder computes the (local) joint frequency distribution of the attributes in $\mathcal{A'}$.
Therefore, if $\mathbf{c}^{(j)}$ is the distribution that holder $j\in \{1,...,M\}$ computes,
then, assuming that $|d_{\mathcal{A'}}|=d$,
holder $j$ shares the following $d$-dimensional vector: 
$$\mathbf{\tilde{c}}^{(j)} = \mathbf{c}^{(j)} + \boldsymbol{\eta}^{(j)} 
= [c_1^{(j)} \ c_2^{(j)} \ ... \ c_d^{(j)}]^T + [\eta_1^{(j)} \ \eta_2^{(j)} \ ... \ \eta_d^{(j)}]^T
= [\tilde{c}_1^{(j)} \ \tilde{c}_2^{(j)} \ ... \ \tilde{c}_d^{(j)}]^T$$
where $\boldsymbol{\eta}^{(j)}$ is picked as described in Theorem \ref{thm_noisy_ss_priv}.
Then, the analyst collects the noisy vectors $\mathbf{\tilde{c}}^{(1)}, ... , \mathbf{\tilde{c}}^{(M)}$, it merges them as:
$$\mathbf{\tilde{c}} = \sum_{j=1}^M \mathbf{\tilde{c}}^{(j)} = \sum_{j=1}^M \mathbf{c}^{(j)} + \boldsymbol{\eta}^{(j)} = \mathbf{c} + \sum_{j=1}^M \boldsymbol{\eta}^{(j)}$$
and it estimates the corresponding probability distribution $\mathbf{\tilde{p}} = \frac{1}{n'}\mathbf{\tilde{c}}$. 
Notice that:
$$n' = \sum_{i=1}^d \sum_{j=1}^M \tilde{c}_i^{(j)} = \sum_{i=1}^d \sum_{j=1}^M c_i^{(j)} + \eta_i^{(j)} = n + \sum_{i=1}^d \sum_{j=1}^M \eta_i^{(j)} $$
so $n'$ is the sum of the actual total dataset size $n$, plus the sum of $d \times M$ zero-mean Laplace random variables, and, hence, it is also random with $\E[n']=n$.
To simplify our analysis, we assume that the analyst divides by $n$ and we ignore the consistency issues this causes to the corresponding probability distribution.
The resulting vector $\mathbf{\tilde{p}}=\mathbf{\tilde{p}}_{\mathcal{A'}}$ (we have so far skipped the subscript $\mathcal{A'}$ for simplicity in presentation), is subsequently utilized in the following two ways:
\begin{itemize}
\item[-] To compute the joint entropy of the attributes in $\mathcal{A'}$, $\tilde{H}(\mathcal{A'})$.
\item[-] To compute the joint probability distributions of attributes in subsets of $\mathcal{A'}$ by marginalization, and then, their joint entropies. 
\end{itemize}
Once the analyst has computed all the required entropies, it may compute the mutual information between an attribute $X$ and its candidate parent set $\mathbf{Pa}(X)$, utilizing the well-known formula: 
$$\tilde{I}(X;\mathbf{Pa}(X)) = \tilde{H}(X) + \tilde{H}(\mathbf{Pa}(X)) - \tilde{H}(X,\mathbf{Pa}(X))$$

The pipeline we described is (partly) depicted in the diagram below. \\

\begin{tikzcd}
c_i^{(1)} \arrow[r, "+\eta_i^{(1)}"] 
& \tilde{c}_i^{(1)} \arrow[rdd] 
& 
& \dots
& \tilde{p}_1 \arrow[rdd] 
& 
& \\
\vdots
&
& 
& 
& \vdots
& 
& \\
c_i^{(j)} \arrow[r, "+\eta_i^{(j)}"] 
& \tilde{c}_i^{(j)} \arrow[r]
& + \arrow[r]  % \oplus
& \tilde{c}_i \arrow[r, "\times \frac{1}{n}"]
& \tilde{p}_i \arrow[r] 
& \boxed{H(\bullet)} \arrow[r] 
& \tilde{H}(\mathcal{A'})\\
\vdots
&
& 
& 
& \vdots
& 
& \\
c_i^{(M)} \arrow[r, "+\eta_i^{(M)}"] 
& \tilde{c}_i^{(M)} \arrow[ruu]
& 
& \dots
& \tilde{p}_d \arrow[ruu] 
& 
& \\
\end{tikzcd}

%For example, assume that $k=2$ and $N>3$.
%At some point, the analyst will retrieve (from the data holders) $\mathbf{\tilde{p}}_{X_1,X_2,X_3}$, and use it to estimate $H(X_1,X_2,X_3)$.
%Then, the analyst will marginalize $\mathbf{\tilde{p}}_{X_1,X_2,X_3}$ to get $\mathbf{\tilde{p}}_{X_1}$ and $\mathbf{\tilde{p}}_{X_2,X_3}$.
%These will be used to compute $H(X_1)$ and $H(X_2,X_3)$ 

The aforementioned process is executed for all ${N}\choose{k+1}$ sets $\mathcal{A'}$.
Thus, when the analyst needs to compute the mutual information between $X$ and $\mathbf{Pa}(X)$, 
it first selects a set $\mathcal{A'}$ such that $\{X\} \cup \mathbf{Pa}(X) \subseteq \mathcal{A'}$.
Then, it performs the necessary marginalizations, to extract $\mathbf{\tilde{p}}_X,\mathbf{\tilde{p}}_{\mathbf{Pa}(X)},\mathbf{\tilde{p}}_{X,\mathbf{Pa}(X)}$;
the last distribution needs to be computed only if $\{X\} \cup \mathbf{Pa}(X) \subset \mathcal{A'}$ - otherwise the analyst already has it.
Finally, after computing the entropies of these three distributions ($\tilde{H}(X)$, $\tilde{H}(\mathbf{Pa}(X))$, and -potentially- $\tilde{H}(X,\mathbf{Pa}(X))$),
the analyst utilizes them to estimate $I(X;\mathbf{Pa}(X))$.
Of course, in practice, the analyst would not need to compute the entropies, as the mutual information can be derived directly from the probability distributions.
Nevertheless, we adopt this approach to facilitate our theoretical analysis of the algorithm.

Our main argument is that, if the noisy entropy estimate $\tilde{H}(\mathcal{A'})$ for an arbitrary set $\mathcal{A'}$ is close to the true entropy $H(\mathcal{A'})$,
then so will be the mutual information estimates computed over the course of the algorithm,
and therefore, Algorithm \ref{Sharing the Noisy Sufficient Statistics} will construct a BN structure that accurately encodes the data distribution. \\

Before starting our analysis, we point out that we use the Big $\mathcal{O}$ for multiple variables as described by Cormen et al. \cite{cormen2009introduction}, despite the inconsistencies demonstrated by Howell \cite{howell2008asymptotic},
so for a vector $\mathbf{x} \in \mathbb{R}^n$ and two functions $f,g$ defined on some subset of $\mathbb{R}^n$:
$$ f(\mathbf{x}) = \mathcal{O}(g(\mathbf{x})) \text{ as } \mathbf{x} \rightarrow \infty $$
$$ \ \Leftrightarrow \ \exists M , \exists C > 0 \text{ such that } \forall \mathbf{x} \text{ with } ||\mathbf{x}||_{\infty} \geq M \ , \ |f(\mathbf{x}| \leq C |g(\mathbf{x})|$$
In practice, we may be interested in settings where some of the variables are fixed (for instance, we may not care about having an arbitrarily large number of data holders).
If this is the case, we may view these variables as constants, which would simplify the resulting expressions.

We start our analysis with a few useful lemmas.
Proving the first  of them (Lemma \ref{lemma_multinomial}) involves tedious computations;
equivalent results were obtained by Harris \cite{harris1975statistical}.
\begin{lemma} \label{lemma_multinomial}
Let $\mathbf{c}=[c_1 \ ... \ c_d]^T$ be a random vector that follows a multinomial distribution with parameters $n,\mathbf{p}$, where $\mathbf{p}=[p_1 \ ... \ p_d]^T$ and $\sum_{i=1}^d{p_i} = 1$. Then, $\forall i \in \{1,...,d\}$, the first 6 moments of $c_i$ are: 
\begin{eqnarray*}
\E [c_i] & = & n p_i \\ 
\E [c_i^2] & = & n^2 p_i^2 + n (-p_i^2 + p_i) \\ 
\E [c_i^3] & = & n^3 p_i^3 + n^2 (-3p_i^3 + 3p_i^2) + n(2p_i^3-3p_i^2+p_i) \\ 
\E [c_i^4] & = & n^4 p_i^4 + n^3 (-6p_i^4 + 6p_i^3) + n^2(11p_i^4-18p_i^3+7p_i^2) \\
&& \ + n(-6p_i^4+12p_i^3-7p_i^2+p_i) \\ 
\E [c_i^5] & = & n^5 p_i^5 + n^4 (-10p_i^5 + 10p_i^4) + n^3(35p_i^5-60p_i^4+25p_i^3) \\
&& \ + n^2(-50p_i^5+110p_i^4-75p_i^3+15p_i^2) \\
&& \ + n(24p_i^5-60p_i^4+50p_i^3-15p_i^2+p_i) \\ 
\E [c_i^6] & = & n^6 p_i^6 + n^5 (-15p_i^6 + 15p_i^5) + n^4(85p_i^6-150p_i^5+65p_i^4) \\
&& \ + n^3(-225p_i^6+525p_i^5-390p_i^4+90p_i^3) \\
&& \ + n^2(274p_i^6-750p_i^5+715p_i^4-270p_i^3+31p_i^2) \\
&& \ + n(-120p_i^6+360p_i^5-390p_i^4+180p_i^3-31p_i^2+p_i)
\end{eqnarray*}
In addition, for $j \not = i$:
\begin{eqnarray*}
\E [c_i c_j] &=& n (n-1) p_i p_j \\
\E [c_i^2 c_j] &=& n (n-1) p_i p_j + n (n-1) (n-2) p_i^2 p_j
\end{eqnarray*}
\end{lemma}

In proving the next Lemma \ref{lemmma_laplace} we utilize the multinomial theorem, the linearity of expectations and the independence assumption between the Laplace random variables.
\begin{lemma} \label{lemmma_laplace}
Let $\eta_1, ... , \eta_M$ be i.i.d. $\text{Laplace}(0,b)$ random variables. Then, $\forall i \in \{1,...,M\}$ and for some $\kappa \in \mathbb{N}$:
\[ 
\E [ \eta_i^{\kappa} ] = \left\{
\begin{array}{ll}
      0 & \text{if } \kappa \text{ is odd}\\
      b^{\kappa} \kappa! & \text{if } \kappa \text{ is even} \\
\end{array} 
\right. 
\]
Also, taking into account that:
$$ \E [ ( \ \sum_{i=1}^M{\eta_i} \ )^{\kappa} ] = 
\sum_{\substack{ (\kappa_1,\kappa_2,...,\kappa_M): \\ \kappa_1+\kappa_2+...+\kappa_M=\kappa }
}{
\frac{\kappa!}{\kappa_1!\kappa_2!...\kappa_M!}\prod_{i=1}^{M}{\E [ \eta_i^{\kappa_i} ]}
} $$
it follows that:
\begin{equation*}
\begin{aligned}
\E [ \sum_{i=1}^M{\eta_i} ] & = 0 
& , \ \  \E [ ( \ \sum_{i=1}^M{\eta_i} \ )^2 ] & = 2 M b^2 \\
\E [ ( \ \sum_{i=1}^M{\eta_i} \ )^3 ] & = 0 
& , \ \  \E [ ( \ \sum_{i=1}^M{\eta_i} \ )^4 ] & = 12 (M^2+M)b^4 \\
\E [ ( \ \sum_{i=1}^M{\eta_i} \ )^5 ] & = 0 
& , \ \  \E [ ( \ \sum_{i=1}^M{\eta_i} \ )^6 ] & = 240 (M^3+2M)b^6
\end{aligned}
\end{equation*}
\end{lemma}

Lemma \ref{lemma_cauchy_scwharz} is an application of the well-known Cauchy-Schwarz inequality on the expectations of random variables.
\begin{lemma} \label{lemma_cauchy_scwharz}
For any two random variables $X$ and $Y$:
$$ | \E[ XY ] | \leq \sqrt{ \E[X^2] \E[Y^2]}$$
where equality holds if and only if $X=cY$ for some constant $c \in \mathbb{R}$.
\end{lemma}

The next Lemma \ref{lemma_p_min} will allow us to bound sums that appear in our analysis.
\begin{lemma} \label{lemma_p_min}
Let $\mathbf{p}=[p_1 \ ... \ p_d]^T$ be a probability distribution over a discrete alphabet of size $d$,
and let $0<p_{min} \leq p_i, \ \forall i \in \{1,...,d\}$.
Then: 
\begin{eqnarray*}
& (i) &
\sum_{i=1}^d \frac{1}{(p_i)^{\kappa}} \leq \frac{d}{(p_{min})^{\kappa}} = \mathcal{O}(\frac{d}{(p_{min})^{\kappa}}) \ \ (\text{for some } \kappa \in \mathbb{N}) \\
\end{eqnarray*}
If, in addition, $p_{min} < \frac{1}{e^2}$, then:
\begin{eqnarray*}
& (ii) & 
\sum_{i=1}^d (1+\log p_i)^2 \leq d (1+\log p_{min})^2 = \mathcal{O}(d \log^2p_{min}) \\
& (iii) & 
\sum_{i=1}^d \frac{|1+\log p_i|}{p_i} \leq d \frac{|1+\log p_{min}|}{p_{min}} = \mathcal{O}(\frac{d \log p_{min}}{p_{min}}) \\
& (iv) & 
\sum_{i=1}^d \frac{|1+\log p_i|}{(p_i)^2} \leq d \frac{|1+\log p_{min}|}{(p_{min})^2} = \mathcal{O}(\frac{d \log p_{min}}{(p_{min})^2})
%& (ii) & 
%\sum_{i=1}^d (p_i)^{\kappa} \log^2 (p_i) = \mathcal{O}(d) \ \ (\text{for some } \kappa \in \mathbb{N})
\end{eqnarray*}
The asymptotic expressions hold for $d \rightarrow \infty, \ p_{min} \rightarrow 0$.
\end{lemma}

We make the following remarks concerning the proofs of the inequalities that appear in Lemma \ref{lemma_p_min}:
\begin{itemize}
\item[-] Inequality $(i)$:
the proof is trivial.
\item[-] Inequalities $(ii),(iii),(iv)$:
the function $f(x)=|1+\log(x)|, \ x \in (0,1]$, has the following behavior:
\begin{itemize}
\item[] $\frac{1}{e^2} \leq x \leq 1 \ \Leftrightarrow \ f(x) \leq 1$
\item[] $0 < x < \frac{1}{e^2} \ \Leftrightarrow \ f(x) > 1$ and f is monotonically decreasing.
\end{itemize}
Thus, if $p_{min} < \frac{1}{e^2}$, then $\max_i f(p_i) = f(p_{min})$.
Taking this into account, we derive the desired results by simple algebra.
%\item[-] Inequality $()$:
%for any $\kappa$, $(p_i)^{\kappa}$ goes to zero faster than $\log^2(p_i)$ goes to infinity, so their product is $o(1)$.
\end{itemize}
For example, we apply inequalities $(i)$ and $(ii)$ for a uniform distribution (assuming $d \geq 8 \Leftrightarrow p_{min} = \frac{1}{d} < \frac{1}{e^2}$):
\begin{itemize}
\item[] $\sum_{i=1}^d \frac{1}{(p_i)^{\kappa}} = \sum_{i=1}^d \frac{1}{(\frac{1}{d})^{\kappa}} = d^{\kappa+1}$
\item[] $ \sum_{i=1}^d (1+\log p_i)^2 = d (1+ \log \frac{1}{d})^2 $
\end{itemize}

The last Lemma \ref{lemma_entropy} is due to Basharin \cite{basharin1959statistical} who first examined the properties of the empirical entropy estimator,
and Harris \cite{harris1975statistical}, who performed a much more detailed analysis.
\begin{lemma} \label{lemma_entropy}
Let $\mathbf{p}=[p_1 \ ... \ p_d]^T$ be a probability distribution over a discrete alphabet of size $d$,
and let $H(\mathbf{p}) = - \sum_{i=1}^d p_i \log_2 p_i$ be the entropy functional of $\mathbf{p}$.
The empirical entropy entropy estimator is computed as:
$$ \hat{H} = H(\mathbf{\hat{p}}) = - \sum_{i=1}^d \hat{p}_i \log_2 \hat{p}_i = - \log_2(e)\sum_{i=1}^d \hat{p}_i \log \hat{p}_i $$
where $\mathbf{\hat{p}}$ is the maximum likelihood estimate of $\mathbf{p}$ using $n$ independent data points. \\
Ignoring the $\log_2(e)$ scale factor, the bias and mean squared error of $\hat{H}$ are:
\begin{equation*}
\begin{aligned}
& \E [ \hat{H} - H ] &=& \ -\frac{d-1}{2n} + \frac{1-\sum_{i=1}^d \frac{1}{p_i}}{12n^2} + \mathcal{O}(\frac{1}{n^3}) \\
& \E [ (\hat{H} - H)^2 ] &=& \ \frac{\sum_{i=1}^d p_i \log_2^2 p_i - H^2}{n} + \frac{d^2-1}{4n^2} + \mathcal{O}(\frac{1}{n^3})
\end{aligned}
\end{equation*}
\end{lemma}
Following all previous authors on the topic, we also ignore the scale factor in our analysis,
and work with the entropy defined using the natural logarithm.
This has no essential effect on the computation, and at the end, we can multiply everything through by $\log_2(e)$ to change the base back to base two.

%\color{red}{
%\textbf{FIX THESE!}
%
%\begin{thm}
%If $\eps \ = \ \omega(\frac{N^2\sqrt{Md}}{n})$, then the bias and mean-squared error of the $\eps$-differentially private entropy estimator $\tilde{H}$ converges asymptotically to that of the non-private empirical entropy estimator.
%\end{thm}
%
%\begin{thm}
%If Theorem 2 holds, then the distributed and $\eps$-differentially private entropy estimator $\tilde{H}$ used by Algorithm \ref{Sharing the Noisy Sufficient Statistics} achieves:
%$$ \E [\tilde{H}-H] \ = \ -\frac{d-1}{2n}
%\ + \ \mathcal{O}(\frac{d}{n^2})
%\ - \ \frac{Mb^2}{n^2}\sum_{i=1}^{d}{\frac{1}{p_i}}
%\ - \ \theta$$
%
%$$  \ = \ \frac{1}{n}(\sum_{i=1}^{d}{p_i\log^2{p_i}}-H^2 ) 
%\ + \ \mathcal{O}(\frac{d^2}{n^2}) 
%\ + \ \frac{2Mb^2}{n^2}\sum_{i=1}^{d}{(1+\log{p_i})^2} 
%\ + \ \theta $$
%where $H$ is the true entropy and $\theta$ consists of dominated noise terms.
%\end{thm}
%}
%\color{black}

Taking these into account, we provide two theorems that give the conditions under which our distributed, differentially private entropy estimator accurately approximates the true entropy.
In the first (Theorem \ref{thm_noiy_ss_bias}), we examine the estimator's bias.

\begin{thm} \label{thm_noiy_ss_bias}
The absolute bias of the distributed, differentially private entropy estimator $\tilde{H}$ used by Algorithm \ref{Sharing the Noisy Sufficient Statistics} is:
\begin{eqnarray*}
| \ \E [ \tilde{H} - H ] \ | & = & \frac{d-1}{2n}
+ \mathcal{O}(\ \frac{dMb^2}{p_{min} n^2 }
+ \frac{dMb^2}{(p_{min})^2 n^3}
+ \frac{d M^2 b^4}{(p_{min})^3 n^4}
\ )
\end{eqnarray*}
If $p_{min} = \omega(\frac{1}{n})$ and we pick $\eps = \omega(\sqrt{\frac{M}{p_{min}n}}N^{k+1})$, then the noise terms get dominated by the $\mathcal{O}(\frac{d}{n})$ term, and the bias of $\tilde{H}$ converges asymptotically to that of the empirical entropy estimator $\hat{H}$ (Lemma \ref{lemma_entropy}).
\end{thm}

\begin{tcolorbox}[breakable]
\begin{proof}
Let $k$ be the degree of the target Bayesian Network.
We fix a set of attributes $\mathcal{A'} \subseteq \mathcal{A}$ such that $|\mathcal{A'}|=k+1$, and assume $|d_{\mathcal{A'}}|=d$.
For simplicity in presentation, we skip the subscripts and denote the joint frequency and probability distributions of the attributes in $\mathcal{A'}$ by $\mathbf{c}$ and $\mathbf{p}$ respectively.

Recall that, based on the algorithm's pipeline and our assumptions:
$$\tilde{\mathbf{c}} = \mathbf{c} + \sum_{j=1}^M \boldsymbol{\eta}^{(j)}$$
so $\tilde{\mathbf{c}}$ is the sum of vector $\mathbf{c} \sim \text{Multinomial}(n,\mathbf{p})$ and $M$ i.i.d. $\text{Laplace}(0,b)$ random vectors (with independent entries). 
Each entry in $\tilde{\mathbf{c}}$ is computed as $\tilde{c}_i = c_i + \sum_{j=1}^M \eta_i^{(j)}$, $\forall i \in \{1,...,d\}$.
We utilize Lemmas \ref{lemma_multinomial} and \ref{lemmma_laplace}, and we use red fonts to emphasize the additive terms that emerge due to the perturbation:
\begin{eqnarray*}
\E [\tilde{c}_i] & = & n p_i \\ 
\E [\tilde{c}_i^2] & = & n^2 p_i^2 + n (-p_i^2 + p_i) + \textcolor{red}{2Mb^2} \\ 
\E [\tilde{c}_i^3] & = & n^3 p_i^3 + n^2 (-3p_i^3 + 3p_i^2) + n(2p_i^3-3p_i^2+p_i+\textcolor{red}{6p_i Mb^2}) \\ 
\E [\tilde{c}_i^4] & = & n^4 p_i^4 + n^3 (-6p_i^4 + 6p_i^3) + n^2(11p_i^4-18p_i^3+7p_i^2+\textcolor{red}{12p_i^2 Mb^2}) \\
&& \ + n[-6p_i^4+12p_i^3-7p_i^2+p_i + \textcolor{red}{12(-p_i^2+p_i) Mb^2}] \\ 
&& \ + \textcolor{red}{12 (M^2+M) b^4} \\
\E [\tilde{c}_i^5] & = & n^5 p_i^5 + n^4 (-10p_i^5 + 10p_i^4) + n^3(35p_i^5-60p_i^4+25p_i^3+\textcolor{red}{20p_i^3 Mb^2}) \\
&& \ + n^2[-50p_i^5+110p_i^4-75p_i^3+15p_i^2+\textcolor{red}{60(-p_i^3+p_i^2) Mb^2}] \\
&& \ + n [ 24p_i^5-60p_i^4+50p_i^3-15p_i^2+p_i \\
&& \ \ \ \ \ + \textcolor{red}{(40p_i^3-60p_i^2+20p_i) M b^2} + \textcolor{red}{60 p_i (M^2+M) b^4} ] \\ 
\E [\tilde{c}_i^6] & = & n^6 p_i^6 + n^5 (-15p_i^6 + 15p_i^5) + n^4(85p_i^6-150p_i^5+65p_i^4 + \textcolor{red}{30 p_i^4 M b^2)} \\
&& \ + n^3[-225p_i^6+525p_i^5-390p_i^4+90p_i^3 + \textcolor{red}{180 (-p_i^4 + p_i^3) M b^2}] \\
&& \ + n^2 [274p_i^6-750p_i^5+715p_i^4-270p_i^3+31p_i^2 \\
&& \ \ \ \ \ + \textcolor{red}{(330 p_i^4 - 540 p_i^3+ 210 p_i^2) M b^2} + \textcolor{red}{180 p_i^2 (M^2+M) b^4}] \\
&& \ + n [-120p_i^6+360p_i^5-390p_i^4+180p_i^3-31p_i^2+p_i \\
&& \ \ \ \ \ + \textcolor{red}{ 30 (-p_i^2 +p_i) M b^2 } + \textcolor{red}{180 (-p_i^2 + p_i) (M^2+M) b^4 } ]\\
&& \ + \textcolor{red}{ 240 (M^3+2M) b^6 }
\end{eqnarray*}
In addition, for $j \not = i$:
\begin{eqnarray*}
\E [\tilde{c}_i \tilde{c}_j] &=& n (n-1) p_i p_j \\
\E [\tilde{c}_i^2 \tilde{c}_j] &=& n^3 p_i^2 p_j + n^2 (-3p_i^2 p_j+p_i p_j) +n(2p_i^2 p_j-p_i p_j+\textcolor{red}{2p_jMb^2})
\end{eqnarray*}
Noting that
$\tilde{p}_i=\frac{\tilde{c}_i}{n} \ \Rightarrow \ \E [\tilde{p}_i^{\kappa}] =\frac{\E [\tilde{c}_i^{\kappa}]}{n^{\kappa}} \ (\forall \kappa \in \mathbb{N})$,
and thus, $\E [\tilde{p}_i]=\frac{np_i}{n}=p_i$, 
we compute the central moments of each $\tilde{p}_i$:
\begin{eqnarray*}
\E [\tilde{p}_i - p_i] & = & 0 \\ 
\E [(\tilde{p}_i - p_i)^2] & = & \frac{1}{n} (-p_i^2 + p_i) + \frac{1}{n^2}\textcolor{red}{2Mb^2} \\ 
\E [(\tilde{p}_i - p_i)^3] & = & \frac{1}{n^2}(2p_i^3-3p_i^2+p_i) \\ 
\E [(\tilde{p}_i - p_i)^4] & = & \frac{1}{n^2}(3p_i^4-6p_i^3+3p_i^2) \\
&& \ + \frac{1}{n^3}[-6p_i^4+12p_i^3-7p_i^2+p_i + \textcolor{red}{12(-p_i^2+p_i) Mb^2}] \\ 
&& \ + \frac{1}{n^4}\textcolor{red}{12 (M^2+M) b^4} \\
\E [(\tilde{p}_i - p_i)^5] & = & \frac{1}{n^3}(-20p_i^5+50p_i^4-40p_i^3+10p_i^2) \\
&& \ + \frac{1}{n^4} [ 24p_i^5-60p_i^4+50p_i^3-15p_i^2+p_i \\
&& \ \ \ \ \ + \textcolor{red}{(40p_i^3-60p_i^2+20p_i) M b^2} ] \\ 
\E [(\tilde{p}_i - p_i)^6] & = & \frac{1}{ n^3 }(-15p_i^6+45p_i^5-45p_i^4+15p_i^3) \\
&& \ + \frac{1}{n^4} [130p_i^6-390p_i^5+415p_i^4-180p_i^3+25p_i^2 \\
&& \ \ \ \ \ + \textcolor{red}{(90 p_i^4 - 180 p_i^3 + 90 p_i^2) M b^2} ] \\
&& \ + \frac{1}{n^5} [-120p_i^6+360p_i^5-390p_i^4+180p_i^3-31p_i^2+p_i \\
&& \ \ \ \ \ + \textcolor{red}{ 30 (- p_i^2 + p_i) M b^2 } + \textcolor{red}{180 (-p_i^2 + p_i) (M^2+M) b^4 } ]\\
&& \ + \frac{1}{n^6}\textcolor{red}{ 240 (M^3+2M) b^6 }
\end{eqnarray*}
In addition, for $j \not = i$:
\begin{eqnarray*}
\E [(\tilde{p}_i-p_i)(\tilde{p}_j-p_j)] &=& \frac{1}{n} p_i p_j \\
\E [(\tilde{p}_i-p_i)^2 (\tilde{p}_j-p_j)] &=& \frac{1}{n^2} (2p_i^2 p_j-p_i p_j)
\end{eqnarray*}

We take a third order Taylor expansion (with remainder) of $\tilde{H}$ around the true entropy $H$:
\begin{eqnarray*}
\tilde{H} & = & H - \sum_{i=1}^d (1+\log p_i)(\tilde{p_i}-p_i) 
- \frac{1}{2} \sum_{i=1}^d \frac{(\tilde{p_i}-p_i)^2}{p_i}
+ \frac{1}{6} \sum_{i=1}^d \frac{(\tilde{p_i}-p_i)^3}{p_i^2} \\
& & \ - \frac{1}{12} \sum_{i=1}^d \frac{(\tilde{p_i}-p_i)^4}{[ (1-c) p_i + c \tilde{p_i}]^3}
\end{eqnarray*}
where $c \in [0,1]$.
We next take $\E [\tilde{H}]$, and by linearity of expectations, we examine the expectation of each term in the Taylor expansion separately:
\begin{eqnarray*}
\sum_{i=1}^d (1+\log p_i) \E [\tilde{p_i}-p_i]
& = &  0 \\
\sum_{i=1}^d \frac{\E [(\tilde{p_i}-p_i)^2 ]}{p_i}
& = & \sum_{i=1}^d \frac{1}{n}\frac{-p_i^2+p_i}{p_i} + \frac{1}{n^2}2Mb^2\frac{1}{p_i} \\
& = & \frac{d-1}{n}+\frac{2Mb^2 \sum_{i=1}^d\frac{1}{p_i}}{n^2} \\
& = & \frac{d-1}{n} + \mathcal{O}( \frac{dMb^2}{p_{min} n^2 } ) \\
\sum_{i=1}^d \frac{\E [(\tilde{p_i}-p_i)^3]}{p_i^2}
& = & \sum_{i=1}^d \frac{1}{n^2}\frac{2p_i^3-3p_i^2+p_i}{p_i^2} \\
& = & \frac{2-3d+\sum_{i=1}^d\frac{1}{p_i}}{n^2} \\
& = & \mathcal{O}( \frac{d}{p_{min}n^2} ) \\
\sum_{i=1}^d \E [ \frac{(\tilde{p_i}-p_i)^4}{[ (1-c) p_i + c \tilde{p_i}]^3} ]
& \leq &  \sum_{i=1}^d \frac{\E [ (\tilde{p_i}-p_i)^4 ]}{(1-c)^3 p_i^3}\\
& = & \frac{3-6d+3\sum_{i=1}^d\frac{1}{p_i}}{(1-c)^3 n^2} \\
&& \ + \frac{-6+12d-7\sum_{i=1}^d\frac{1}{p_i}+\sum_{i=1}^d\frac{1}{p_i^2}}{(1-c)^3 n^3} \\
&& \ + \frac{12Mb^2(-\sum_{i=1}^d\frac{1}{p_i}+\sum_{i=1}^d\frac{1}{p_i^2})}{(1-c)^3 n^3} \\
&& \ + \frac{12(M^2+M)b^4\sum_{i=1}^d\frac{1}{p_i^3}}{(1-c)^3 n^4} \\
& = & \mathcal{O}( \frac{d}{p_{min}n^2} + \frac{dMb^2}{(p_{min})^2 n^3} + \frac{d M^2 b^4}{(p_{min})^3 n^4})
\end{eqnarray*}

To derive the asymptotic expressions, we utilize lemma \ref{lemma_p_min}.

Let us review the variables that appear in our expression:
\begin{itemize}
\item[-] $n$: total number of data points ($n \rightarrow \infty$).
\item[-] $d$: joint domain size of attributes whose entropy we estimate ($d \rightarrow \infty$).
\item[-] $M$: number of data holders ($M \rightarrow \infty$).
\item[-] $b$: noise scale ($b \rightarrow \infty$).
Recall that $b=\frac{{{N}\choose{k+1}}}{\eps}=\mathcal{O}(\frac{N^{k+1}}{\eps})$, so:
\begin{itemize}
\item[-] $N$: total number of attributes ($N \rightarrow \infty$).
\item[-] $\eps$: is the privacy budget ($\eps \rightarrow 0$).
\end{itemize}
\item[-] $p_{min}$: minimum entry in joint probability distribution of attributes whose entropy we estimate ($p_{min} \rightarrow 0$).
\end{itemize}
So, for example, the $\mathcal{O}( \frac{d}{p_{min}n^2} )$ terms get dominated by the $\mathcal{O}( \frac{dMb^2}{p_{min} n^2 } ) $ term.

Combining the above yields:
\begin{eqnarray*}
| \ \E [ \tilde{H} - H ] \ | & = & \frac{d-1}{2n}
+ \mathcal{O}(\ \frac{dMb^2}{p_{min} n^2 }
+ \frac{dMb^2}{(p_{min})^2 n^3}
+ \frac{d M^2 b^4}{(p_{min})^3 n^4}
\ )
\end{eqnarray*}

Therefore, to ensure that the bias of $\tilde{H}$ converges asymptotically to that of $\hat{H}$, the following must hold:
$$
\left\{
\begin{array}{ll}
\frac{dMb^2}{p_{min} n^2 } = o(\ \frac{d}{n} \ ) \\
\frac{dMb^2}{(p_{min})^2 n^3}  = o(\ \frac{d}{n} \ ) \\
\frac{d M^2 b^4}{(p_{min})^3 n^4} = o(\ \frac{d}{n} \ )
\end{array}
\right.
\Leftrightarrow
%\left\{
%\begin{array}{ll}
%\frac{Mb^2}{p_{min} n } = o(\ 1 \ ) \\
%\frac{1}{p_{min} n}  = o(\ 1 \ )
%\end{array}
%\right.
%\Leftrightarrow
\left\{
\begin{array}{ll}
\eps = \omega(\sqrt{\frac{M}{p_{min}n}}N^{k+1}) \\
p_{min} = \omega(\frac{1}{n})
\end{array}
\right.
$$
which completes the proof.
\end{proof}
\end{tcolorbox}

We next examine the estimator's mean squared error (Theorem \ref{thm_noiy_ss_mse}).
\begin{thm} \label{thm_noiy_ss_mse}
Assume that the data distribution $\mathbf{p}$ satisfies $p_{min} < \frac{1}{e^2}$, $p_{min} = \omega(\frac{1}{\sqrt{n}})$ and  $\sigma^2 = \mathcal{O}(d^2)$, where $\sigma^2 = (\sum_{i=1}^d p_i\log^2 p_i)-H^2$. \\
Then, if we pick $\eps = \omega(\sqrt{\frac{M}{n}}N^{k+1}|\log(p_{min})|)$,
the mean squared error of the distributed, differentially private entropy estimator $\tilde{H}$ used by Algorithm \ref{Sharing the Noisy Sufficient Statistics} is:
$$ \E [(\tilde{H}-H)^2] = \frac{\sigma^2}{n} + o(\frac{\sigma^2}{n})$$
and it converges asymptotically to that of the empirical entropy estimator $\hat{H}$ (Lemma \ref{lemma_entropy}).
\end{thm}

\begin{tcolorbox}[breakable]
\begin{proof}
The proof of Theorem \ref{thm_noiy_ss_mse} follows that of Theorem \ref{thm_noiy_ss_bias}, but involves much more complicated computations.
We therefore focus on the asymptotic behavior of the quantities involved in our analysis, and we analytically compute only the dominant ones.
We begin by deriving asymptotic expressions for the central moments of each $\tilde{p}_i$ (as $n,M,b \rightarrow \infty$), introducing the notation $\theta=\frac{Mb^2}{n^2}$:
\begin{eqnarray*}
\E [\tilde{p}_i - p_i] & = & 0 \\ 
%% 2
\E [(\tilde{p}_i - p_i)^2] & = & \mathcal{O}(\frac{1}{n} + \frac{Mb^2}{n^2}) \ = \ \mathcal{O}(\frac{1}{n} + \theta) \\ 
%% 3
\E [(\tilde{p}_i - p_i)^3] & = & \mathcal{O}(\frac{1}{n^2}) \\ %% \ = \ o(\frac{p_i}{n})\\ 
%% 4
\E [(\tilde{p}_i - p_i)^4] & = & \mathcal{O}(\frac{1}{n^2} + \frac{M b^2}{n^3} + \frac{M^2 b^4}{n^4} )  \ = \ \mathcal{O}(\frac{1}{n^2} + \frac{\theta}{n} + \theta^2) \\
%% 5
\E [(\tilde{p}_i - p_i)^5] & = & \mathcal{O} ( \frac{1}{n^3} + \frac{M b^2}{n^4} )  \ = \ \mathcal{O}(\frac{1}{n^3} + \frac{\theta}{n^2}) \\ 
%% 6
\E [(\tilde{p}_i - p_i)^6] & = & \mathcal{O} ( \frac{1}{ n^3 } + \frac{M b^2}{n^4} + \frac{M^2 b^4}{n^5} + \frac{M^3 b^6}{n^6} ) \\
& = & \mathcal{O}(\frac{1}{n^3} + \frac{\theta}{n^2} + \frac{\theta^2}{n} + \theta^3)
\end{eqnarray*}
In addition, for $j \not = i$:
\begin{eqnarray*}
\E [(\tilde{p}_i-p_i)(\tilde{p}_j-p_j)] &=&  \mathcal{O}(\frac{1}{n}) \\
\E [(\tilde{p}_i-p_i)^2 (\tilde{p}_j-p_j)] &=&  \mathcal{O}(\frac{1}{n^2})
\end{eqnarray*}
We are also interested in higher order cross moments, so we apply Lemma \ref{lemma_cauchy_scwharz}:
\begin{eqnarray*}
\E [(\tilde{p}_i-p_i)^2(\tilde{p}_j-p_j)^2] & \leq & \sqrt{ \E [(\tilde{p}_i - p_i)^4] \E [(\tilde{p}_j - p_j)^4] } \\
& = & \sqrt{ \mathcal{O}(\frac{1}{n^2} + \frac{\theta}{n} + \theta^2 )^2 } \\
& = & \mathcal{O}(\frac{1}{n^2} + \frac{\theta}{n} + \theta^2 ) \\
\E [(\tilde{p}_i-p_i) (\tilde{p}_j-p_j)^3] & \leq & \sqrt{ \E [(\tilde{p}_i - p_i)^2] \E [(\tilde{p}_j - p_j)^6] } \\
& = & \sqrt{ \mathcal{O}(\frac{1}{n} + \theta) \ \mathcal{O} ( \frac{1}{ n^3 } + \frac{\theta}{n^2} + \frac{\theta^2}{n} + \theta^3 ) } \\
& = & \mathcal{O}(\frac{1}{ n^2 } + \frac{\theta^{\frac{1}{2}}}{n^{\frac{3}{2}}} + \frac{\theta}{n} + \frac{\theta^{\frac{3}{2}}}{n^{\frac{1}{2}}} + \theta^2 ) \\
\E [(\tilde{p}_i-p_i)^2(\tilde{p}_j-p_j)^3] & \leq & \sqrt{ \E [(\tilde{p}_i - p_i)^4] \E [(\tilde{p}_j - p_j)^6] } \\
& = & \sqrt{ \mathcal{O}(\frac{1}{n^2} + \frac{\theta}{n} + \theta^2) \ \mathcal{O}(\frac{1}{n^3} + \frac{\theta}{n^2} + \frac{\theta^2}{n} + \theta^3) } \\
& = & \mathcal{O}( \frac{1}{ n^{\frac{5}{2}} } + \frac{\theta^{\frac{1}{2}}}{n^2} + \frac{\theta}{n^{\frac{3}{2}}} + \frac{\theta^{\frac{3}{2}}}{n} + \frac{\theta^2}{n^{\frac{1}{2}}} + \theta^{\frac{5}{2}}  ) \\
\E [(\tilde{p}_i-p_i)^3 (\tilde{p}_j-p_j)^3] & \leq & \sqrt{ \E [(\tilde{p}_i - p_i)^6] \E [(\tilde{p}_j - p_j)^6] } \\
& = & \sqrt{ \mathcal{O}(\frac{1}{n^3} + \frac{\theta}{n^2} + \frac{\theta^2}{n} + \theta^3)^2 } \\
& = & \mathcal{O}(\frac{1}{n^3} + \frac{\theta}{n^2} + \frac{\theta^2}{n} + \theta^3)
\end{eqnarray*}

We take a second order Taylor expansion (with remainder) of $\tilde{H}$ around the true entropy $H$, and examine the quantity $(\tilde{H}-H)^2$:
\begin{eqnarray*}
(\tilde{H}-H)^2 & = & 
[ \ H - \sum_{i=1}^d (1+\log p_i)(\tilde{p_i}-p_i) 
- \frac{1}{2} \sum_{i=1}^d \frac{(\tilde{p_i}-p_i)^2}{p_i} \\
&& \ + \frac{1}{6} \sum_{i=1}^d \frac{(\tilde{p_i}-p_i)^3}{[ (1-\theta) p_i + \theta \tilde{p_i}]^2} - H \ ]^2 \\
& = & [ \ \sum_{i=1}^d (1+\log p_i)(\tilde{p_i}-p_i) \ ]^2 \\
&& \ + \frac{1}{4} [ \ \sum_{i=1}^d \frac{(\tilde{p_i}-p_i)^2}{p_i} \ ]^2 \\
&& \ + \frac{1}{36} [ \ \sum_{i=1}^d \frac{(\tilde{p_i}-p_i)^3}{[ (1-c) p_i + c \tilde{p_i}]^2} \ ]^2 \\
&& \ + [ \ \sum_{i=1}^d (1+\log p_i)(\tilde{p_i}-p_i) \ ] \ [ \ \sum_{i=1}^d \frac{(\tilde{p_i}-p_i)^2}{p_i} \ ] \\
&& \ - \frac{1}{3} [ \ \sum_{i=1}^d (1+\log p_i)(\tilde{p_i}-p_i) \ ] \ [ \ \sum_{i=1}^d \frac{(\tilde{p_i}-p_i)^3}{[ (1-c) p_i + c \tilde{p_i}]^2} \ ] \\
&& \ - \frac{1}{6} [ \ \sum_{i=1}^d \frac{(\tilde{p_i}-p_i)^2}{p_i} \ ] \ [ \ \sum_{i=1}^d \frac{(\tilde{p_i}-p_i)^3}{[ (1-c) p_i + c \tilde{p_i}]^2} \ ]
\end{eqnarray*}
where $c \in [0,1]$.
We next take $\E [(\tilde{H}-H)^2]$, and by linearity of expectations, we examine the expectation of each term separately.

The first term, $\E [ \ [ \ \sum_{i=1}^d (1+\log p_i)(\tilde{p_i}-p_i) \ ]^2 \ ]$, involves the $\mathcal{O}(\frac{1}{n})$ terms $\E [(\tilde{p}_i - p_i)^2]$ and $\E [(\tilde{p}_i-p_i)(\tilde{p}_j-p_j)]$,
so we compute it analytically:
\begin{eqnarray*}
%% TERM 1
\E [ \ [ \ \sum_{i=1}^d (1+\log p_i)(\tilde{p_i}-p_i) \ ]^2 \ ]
& = & \sum_{i=1}^d (1+\log p_i)^2 \E [ (\tilde{p_i}-p_i)^2 ] \\
&& \ + 2 \sum_{i=1}^{d-1} \sum_{j=i+1}^{d} (1+\log p_i)(1+\log p_j) \\
&& \ \ \ \E [(\tilde{p_i}-p_i)(\tilde{p_j}-p_j) ] \\
& = & \frac{1}{n} [(\sum_{i=1}^d p_i\log^2 p_i)-H^2] \\
&& \ + \frac{1}{n^2} 2 M b^2 \sum_{i=1}^d (1+\log p_i)^2\\
& = & \frac{1}{n} [(\sum_{i=1}^d p_i\log^2 p_i)-H^2] \\
&& \ + \mathcal{O}( d \log^2 (p_{min}) \theta ) \\
& = & \mathcal{O}(\frac{\sigma^2}{n}) + \mathcal{O}( d \log^2 (p_{min}) \theta )
\end{eqnarray*}

To ensure that the mean squared error of $\tilde{H}$ converges to that of $\hat{H}$, we want the second term to be dominated by the first, that is:
$$ d \log^2 (p_{min}) \theta = o(\frac{\sigma^2}{n})$$
Given that $\sigma^2 = \mathcal{O}(d^2)$, the aforementioned condition is satisfied when:
$$\log^2 (p_{min}) \theta = o(\frac{1}{n}) 
\ \Leftrightarrow \ 
\eps = \omega(\sqrt{\frac{M}{n}}N^{k+1}|\log(p_{min})|)$$
Note that the condition $\frac{\theta}{p_{min}} = o(\frac{1}{n})$, which appeared in Theorem \ref{thm_noiy_ss_bias}, implies the condition we want to satisfy here.

We proceed with the remaining terms, based on Lemma \ref{lemma_p_min}, 
and on the assumption that $\log^2 (p_{min}) \theta = o(\frac{1}{n})$, which implies that $\theta = o(\frac{1}{n})$.
\begin{eqnarray*}
%% TERM 2
\E [ \ [ \ \sum_{i=1}^d \frac{(\tilde{p_i}-p_i)^2}{p_i} \ ]^2 \ ]
& = & \sum_{i=1}^d \frac{\E [ (\tilde{p_i}-p_i)^4 ]}{p_i^2} \\
&& \ + 2 \sum_{i=1}^{d-1} \sum_{j=i+1}^{d} \frac{\E [(\tilde{p_i}-p_i)^2 (\tilde{p_j}-p_j)^2 ]}{p_ip_j} \\
& = & \mathcal{O}(\frac{d^2}{(p_{min})^2}) \ \mathcal{O}( \frac{1}{n^2} + \frac{\theta}{n} + \theta^2 ) \\
& = & \mathcal{O}(\frac{d^2}{(p_{min})^2}) \ [ \mathcal{O}( \frac{1}{n^2} ) + o( \frac{1}{n^2})] \\\\
%% TERM 3
\E [ \ [ \ \sum_{i=1}^d \frac{(\tilde{p_i}-p_i)^3}{[(1-c)p_i+c \tilde{p}_i]^2} \ ]^2 \ ]
& \leq & \E [ \ [ \ \sum_{i=1}^d \frac{(\tilde{p_i}-p_i)^3}{(1-c)^2 p_i^2} \ ]^2 \ ] \\
& = & \sum_{i=1}^d \frac{\E [ (\tilde{p_i}-p_i)^6 ]}{(1-c)^4 p_i^4} \\
&& \ + 2 \sum_{i=1}^{d-1} \sum_{j=i+1}^{d} \frac{\E [(\tilde{p_i}-p_i)^3 (\tilde{p_j}-p_j)^3 ]}{(1-c)^4 p_i^2 p_j^2} \\
& = & \mathcal{O}(\frac{d^2}{(p_{min})^4}) \ \mathcal{O}(\frac{1}{n^3} + \frac{\theta}{n^2} + \frac{\theta^2}{n} + \theta^3) \\
& = & \mathcal{O}(\frac{d^2}{(p_{min})^4}) \ [ \mathcal{O}( \frac{1}{n^3} ) + o( \frac{1}{n^3})]
\end{eqnarray*}

\begin{eqnarray*}
%% TERM 4
& \E & [ \ \sum_{i=1}^d (1+\log p_i)(\tilde{p_i}-p_i) \ \sum_{i=1}^d \frac{(\tilde{p_i}-p_i)^2}{p_i} \ ] \\
& = & \sum_{i=1}^d \frac{(1+\log p_i)}{p_i} \E [ (\tilde{p_i}-p_i)^3 ]
 + \sum_{(i,j):i\not = j} \frac{1+\log p_i}{p_j} \E [ (\tilde{p_i}-p_i)(\tilde{p_j}-p_j)^2 ]\\
& = & \mathcal{O}(\frac{d^2 \log(p_{min})}{p_{min}}) \ \mathcal{O}(\frac{1}{n^2}) \\\\
%% TERM 5
& \E & [ \ \sum_{i=1}^d (1+\log p_i)(\tilde{p_i}-p_i) \ \sum_{i=1}^d \frac{(\tilde{p_i}-p_i)^3}{[(1-c)p_i+c \tilde{p}_i]^2} \ ] \\
& \leq & \E [ \ \sum_{i=1}^d (1+\log p_i)(\tilde{p_i}-p_i) \ \sum_{i=1}^d \frac{(\tilde{p_i}-p_i)^3}{(1-c)^2 p_i^2} \ ] \\
& = & \sum_{i=1}^d \frac{1+\log p_i}{(1-c)^2 p_i^2} \E [ (\tilde{p_i}-p_i)^4 ]
 + \sum_{(i,j):i\not = j} \frac{1+\log p_i}{(1-c)^2 p_j^2} \E [ (\tilde{p_i}-p_i)(\tilde{p_j}-p_j)^3 ]\\
& = & \mathcal{O}(\frac{d^2 \log(p_{min})}{(p_{min})^2}) \ \mathcal{O}(\frac{1}{n^2} + \frac{\theta^{\frac{1}{2}}}{n^{\frac{3}{2}}} + \frac{\theta}{n} + \frac{\theta^{\frac{3}{2}}}{n^{\frac{1}{2}}} + \theta^2) \\
& = & \mathcal{O}(\frac{d^2 \log(p_{min})}{(p_{min})^2}) \ [ \mathcal{O}( \frac{1}{n^2} ) + o( \frac{1}{n^2})] \\\\
%% TERM 6
& \E & [ \ \sum_{i=1}^d \frac{(\tilde{p_i}-p_i)^2}{p_i} \ \sum_{i=1}^d \frac{(\tilde{p_i}-p_i)^3}{[(1-c)p_i+c \tilde{p}_i]^2} \ ] \\
& \leq & \E [ \ \sum_{i=1}^d \frac{(\tilde{p_i}-p_i)^2}{p_i} \ \sum_{i=1}^d \frac{(\tilde{p_i}-p_i)^3}{(1-c)^2 p_i^2} \ ] \\
& = & \sum_{i=1}^d \frac{\E [ (\tilde{p_i}-p_i)^5 ]}{(1-c)^2 p_i^3} 
 + \sum_{(i,j):i\not = j} \frac{\E [ (\tilde{p_i}-p_i)^2(\tilde{p_j}-p_j)^3 ]}{p_i (1-c)^2 p_j^2} \\
& = & \mathcal{O}(\frac{d^2}{(p_{min})^3}) \ \mathcal{O}(\frac{1}{n^{\frac{5}{2}}} + \frac{\theta^{\frac{1}{2}}}{n^2} + \frac{\theta}{n^{\frac{3}{2}}} + \frac{\theta^{\frac{3}{2}}}{n} + \frac{\theta^2}{n^{\frac{1}{2}}} + \theta^{\frac{5}{2}}) \\
& = & \mathcal{O}(\frac{d^2}{(p_{min})^3}) \ [ \mathcal{O}( \frac{1}{n^{\frac{5}{2}}} ) + o( \frac{1}{n^{\frac{5}{2}}})]
\end{eqnarray*}

We want all the resulting terms to go to zero faster than the dominant $\mathcal{O}(\frac{\sigma^2}{n})$ term.
Thus, we need to ensure that all five terms are $o(\frac{\sigma^2}{n})$,
which gives us five conditions to satisfy.
It is easy to see that the stronger condition, which implies the remaining four ones is:
$$\frac{1}{(p_{min})^2 n} = o(1) \ \Leftrightarrow \ p_{min} = \omega(\frac{1}{\sqrt{n}})$$ 
which completes the proof.
\end{proof}
\end{tcolorbox}

\section{Noisy Majority Voting}
As we argued, a major limitation of the Sharing the Sufficient Statistics approach is the high communication cost it incurs;
recall that the joint frequency distribution of the attributes in a set $\mathcal{A'}$  is a vector of dimension $|d_{\mathcal{A'}}|$, which can be quite large.
To tackle this problem, we may be tempted to ask each data holder to instead share the local (noisy) value of the scoring function for each candidate attribute-parent pair.
However, this approach is also problematic:
\begin{itemize}
\item[-] Each data holder has to share an even larger number of scores, and specifically $\sum_{i=1}^{k+1} {{N}\choose{i}}$ scores.
\item[-] The analyst has to re-access the data to the learn the parameters.
\item[-] The empirical mutual information is non-linear, so the analyst cannot compose the global score from local scores.
A naive solution (which does not work well in practice) would be to take the mean or median of the local scores.
An alternative solution, which we leave as an open question, would be to use a linear scoring function.
\end{itemize}
Motivated by these observations, we propose a second answer (Algorithm \ref{Noisy Majority Voting}) to the \emph{What to share?} question, which is based on the notion of majority voting from the distributed machine learning literature.
Specifically, each data holder incrementally reports (a noisy version of) the highest mutual information attribute-parent pair that it would add to the Bayesian Network.
The analyst collects all votes, and adds the most-voted pair to the structure.

\begin{algorithm}%[H]
\DontPrintSemicolon

\KwIn{Datasets $D_1,...,D_M$, BN degree k}

\KwOut{BN structure $\mathcal{G}$}

Initialize $\mathcal{G}=\emptyset$ and $V = \emptyset$

Arbitrarily select an attribute $X$ from $\mathcal{A}$; add $(X,\emptyset)$ to $\mathcal{G}$ and $X$ to $V$

\For{i=1 \KwTo N-1}{

	Initialize multi-set $votes=\emptyset$
	
	\For{$j=1\ \KwTo\ M$}{
	
		\textbf{QUERY}($D_j$): get $(X,\mathbf{Pa}(X))$ with the highest $\tilde{I}(X,\mathbf{Pa}(X))$ , subject to: $X \in A \setminus V$ and $\mathbf{Pa}(X) \subseteq V$
		
		Add $(X,\mathbf{Pa}(X))$ to votes
	}
	
	Find most-voted $(X,\mathbf{Pa}(X))$ in $votes$ (break ties arbitrarily)
	
	Add $(X,\mathbf{Pa}(X))$ to $\mathcal{G}$ and $X$ to $V$
}
Return $\mathcal{G} $
\caption{Noisy Majority Voting\label{Noisy Majority Voting}}
\end{algorithm}

In order to satisfy differential privacy, each data holder responds to the queries on its dataset using the exponential mechanism with the empirical mutual information as scoring function and with privacy budget $\frac{\eps_1}{N-1}$.
This is, in fact, exactly what we describe in Algorithm \ref{Greedy BN Structure Learning} (line \ref{Greedy BN Structure Learning_expo}).

Once the Bayesian Network structure is known, Algorithm \ref{Distributed BN Parameter Learning} is used to learn the parameters;
the analyst now has to retrieve the $N$ required frequency distributions to estimate the parameters ($retrieved=\text{False}$),
and the data holders respond to its queries using the Laplace mechanism, with privacy budget $\frac{\eps_2}{N}$.

\begin{thm} \label{thm_noisy_mv_priv}
Let $\eps_1$ and $\eps_2$ be the total privacy budget that each data holder uses in responding to the analyst's queries, during the structure learning phase (Algorithm \ref{Noisy Majority Voting}) and the parameter learning phase (Algorithm \ref{Distributed BN Parameter Learning} with input $retrieved=\text{False}$), respectively.
Then, the overall algorithm satisfies $(\eps_1+\eps_2)$-differential privacy for any dataset $D_j \ (j \in \{1,...,M\})$.
\end{thm}

\section{Sharing the Noisy Model}
In the final approach we examine, the data holders first learn a local (noisy) model, and then share it with the analyst. 
Therefore, the answer to the \emph{What to share?} question is \emph{the model}, and this is, in fact, equivalent with having each data holder locally run PrivBayes and publish the learned model.
The analyst collects the local models and aggregates them using a technique inspired by knowledge probing;
it generates a synthetic dataset using each local model (of size proportional to that of the local dataset that was used to learn the local model),
and then it learns a global model based on the synthetic data.

\begin{algorithm}%[H]
\DontPrintSemicolon

\KwIn{Datasets $D_1,...,D_M$, BN degree k}

\KwOut{Global BN structure $\mathcal{G}$}

Initialize $D_{synth}=\emptyset$

\For{$k=1\ \KwTo\ M$}{

	\textbf{QUERY}($D_j$): get local model structure $\mathcal{G}^{(j)}$, parameters $\Theta^{(j)}$ and dataset size $n_j$
	
	Generate $D_{j,synth}$ using Prior Sampling such that $|D_{j,synth}|=n_j$
	
	Add $D_{j,synth}$ to $D_{synth}$

}

Run Algorithm \ref{Greedy BN Structure Learning} with input $D_{synth},\ k$ and without using the exponential mechanism in line \ref{Greedy BN Structure Learning_expo}, and return resulting $\mathcal{G}$

\caption{Sharing the Noisy Model\label{Sharing the Noisy Model}}
\end{algorithm}

Each data holder constructs its local model using Algorithms \ref{Greedy BN Structure Learning} and \ref{BN Parameter Learning}, which were shown to jointly satisfy $\eps$-differential privacy (Theorem \ref{thm_privbayes_overall}).
Since differential privacy is immune to post-processing, we have the following theorem.

\begin{thm} \label{thm_noisy_mv_priv}
Assume that each data holder shares an $\eps$-differentially private local model.
Then, the global model that results by aggregating the local models according to Algorithm \ref{Sharing the Noisy Model} also satisfies $\eps$-differential privacy for any dataset $D_j \ (j \in \{1,...,M\})$.
\end{thm}

Notice that, once the analyst collects the local models, it does not need to access the data again.
In particular, the analyst learns both the structure and the parameters of the global model using the synthetic data it generates.
Thus, Algorithm \ref{Sharing the Noisy Model} has the exact same two advantages as Algorithm \ref{Sharing the Noisy Sufficient Statistics};
any algorithm can be used to address the structure learning optimization problem,
and the parameter learning Algorithm \ref{Distributed BN Parameter Learning} utilizes frequency distributions originating from the synthetic data.

\section{Experimental Evaluation}
In this section, we experimentally evaluate our algorithms, both on synthetic and on real-world data.
To conduct our experiments, we build a system in Python programming language, utilizing popular libraries (namely: NumPy, pandas and NetworkX), as well as novel libraries we develop (prob\_utils.py, pgm\_utils.py).

We use two \emph{baselines} for comparison.
\begin{itemize}
\item[-] As a lower bound for performance, we use the maximum likelihood estimate of the distribution that assumes all attributes to be independent (naive-Bayes), with no privacy constraints.
\item[-] As an upper bound, we use a probabilistic graphical model learned with the greedy extension of the Chow-Liu algorithm (Algorithm \ref{Greedy BN Structure Learning}), by centrally collecting all data and with no privacy constraints.
For the case of synthetic data, we also evaluate for comparison the actual underlying Bayesian Network, which is known.
\end{itemize}

\subsection{Evaluation Metric}
The metric we use in our experiments is the \emph{cross entropy} between the actual data distribution, $\mathbf{p}$, and the distribution encoded by the learned Bayesian Network, $\mathbf{p}_{\mathcal{B}}$.
The cross entropy between the aforementioned two distributions, over the set of attributes $\mathcal{A}$ (with $|d_{\mathcal{A}}|=d$) is defined as:
\begin{eqnarray*}
H(\mathbf{p},\mathbf{p}_{\mathcal{B}})
& = & \E_{\mathbf{p}} [-\log_2 \mathbf{p}_{\mathcal{B}}] \\
& = & - \sum_{i=1}^{d} p_i \log_2 p_{\mathcal{B}i} \\
& = & - \sum_{i=1}^{d} p_i \log_2 p_i + p_i \log_2 \frac{ p_{\mathcal{B}i} }{ p_i } \\
& = & H(\mathbf{p}) + D_{KL}(\mathbf{p}||\mathbf{p}_{\mathcal{B}})
\end{eqnarray*}
where:
\begin{itemize}
\item[-] $H(\mathbf{p})$ is the entropy of $\mathbf{p}$,
which measures the expected number of bits needed to encode an element from $d_{\mathcal{A}}$, using a coding scheme that is optimized for $\mathbf{p}$,
\item[-] $D_{KL}(\mathbf{p}||\mathbf{p}_{\mathcal{B}})$ is the Kullback-Leibler divergence (relative entropy) of $\mathbf{p}_{\mathcal{B}}$ from $\mathbf{p}$,
which measures the expected number of extra bits needed to encode an element from $d_{\mathcal{A}}$, using a coding scheme that is optimized for $\mathbf{p}_{\mathcal{B}}$ rather than $\mathbf{p}$,
\end{itemize}
and, thus, $ H(\mathbf{p},\mathbf{p}_{\mathcal{B}})$ measures the expected number of bits needed to encode an element from $d_{\mathcal{A}}$,
using a coding scheme that is optimized for the learned distribution $\mathbf{p}_{\mathcal{B}}$,
rather than the actual distribution $\mathbf{p}$.

In our experiments with synthetic data, the actual data distribution $\mathbf{p}$ is known, so we are able to use the aforementioned formula.
However, when experimenting with real-world data, $\mathbf{p}$ is unknown, so we instead use a Monte Carlo estimate of the cross entropy, namely the \emph{empirical cross entropy} $\hat{H}(D_{test},\mathbf{p}_{\mathcal{B}})$.
To compute $\hat{H}(D_{test},\mathbf{p}_{\mathcal{B}})$, we reserve a fraction ($20 \%$) of the total dataset $D$ which we do not use when learning the Bayesian Network model $p_{\mathcal{B}}$.
Therefore, the reserved, test dataset $D_{test}$ consists of independent data points that were randomly drawn from $D$, and hence can be assumed to follow the same distribution $\mathbf{p}$.
We estimate the empirical cross entropy as:
$$\hat{H}(D_{test},\mathbf{p}_{\mathcal{B}})
= - \frac{1}{|D_{test}|} \sum_{\mathbf{x} \in D_{test}} \log_2 \mathbf{p}_{\mathcal{B}}(\mathbf{x})  $$

Before proceeding, we make an additional, important remark.
In theory, we have assumed that there exist no \emph{zero-probability events} in $\mathbf{p}$, so that all elements in $d_{\mathcal{A}}$ have non-zero probability and, hence, $p_i>0, \ \forall i \in \{1,...,d\}$.
In practice, however, the learned distribution $\mathbf{p}_{\mathcal{B}}$ may contain zero-probability events,
since it is the maximum likelihood estimate of $\mathbf{p}$, computed using a noisy version $\mathbf{\tilde{c}}$ of the observed frequency distribution $\mathbf{c}$.
Even if all elements in $d_{\mathcal{A}}$ are observed (which is unlikely when $d$ is large), so that $c_i>0, \ \forall i \in \{1,...,d\}$,
the noise added by the Laplace mechanism may cause $\tilde{c}_i=0$ for some $i$ (it actually may cause $\tilde{c}_i \leq 0$, but we replace negative frequencies with $0$), and, thus, $p_{\mathcal{B}i} = 0$.
If this is the case, it is not hard to see that both $H(\mathbf{p},\mathbf{p}_{\mathcal{B}}) = \infty$ and $\hat{H}(D_{test},\mathbf{p}_{\mathcal{B}})=\infty$.
We tackle this problem by using \emph{additive (add-one) smoothing}, so $\forall i$, we estimate the smoothed probability $\tilde{p}'_{\mathcal{B} i}$ based on the smoothed frequency distribution:
$$ \tilde{c}'_i = \tilde{c}_i + 1 \ \Rightarrow \ \tilde{p}'_{\mathcal{B} i} = \frac{\tilde{c}'_i}{n+d}$$

\subsection{Synthetic Data}
\begin{figure}%[H]%[ht!]
    \centering
    \includegraphics[width=0.75\columnwidth]{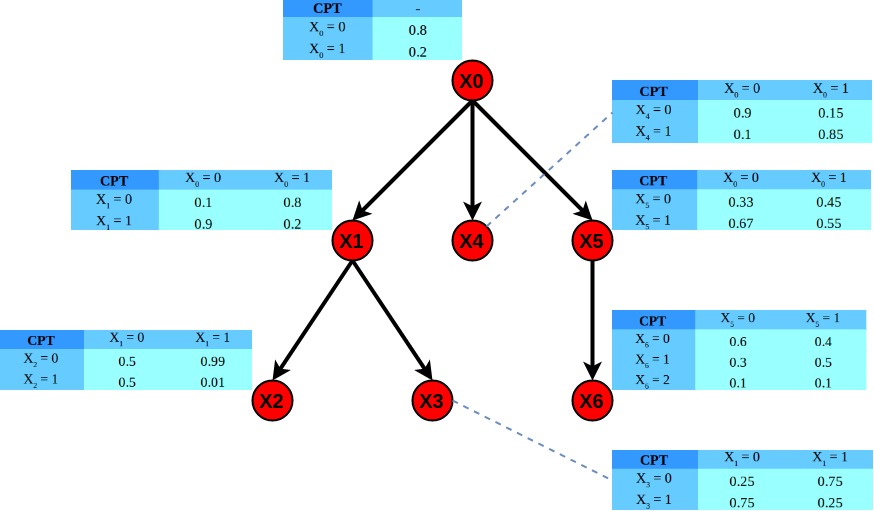}
    \caption{Actual data distribution}
	\label{FIG_synthetic_graph_structure}
\end{figure}

In our first set of experiments, we use a \emph{synthetic dataset}, where each data point consists of a (relatively small) set of attributes $\mathcal{A}=\{X_0,...,X_6\}$.
The first six attributes ($X_0-X_5$) are binary, whereas the seventh attribute ($X_6$) is ternary,
so the joint domain of the 7 attributes has size $d=|d_{\mathcal{A}}|=2^6 \times 3 = 192$.
The actual data distribution $\mathbf{p}_{\mathcal{A}}=\mathbf{p}$ can be encoded by the Bayesian Network illustrated in Figure \ref{FIG_synthetic_graph_structure}.

A tedious (but straightforward) computation gives that $H(\mathbf{p})=5.12$ bits,
whereas for a uniform distribution $\mathbf{p'}$, $H(\mathbf{p;}) = -\log_2(\frac{1}{192}) = 7.58$ bits.
This illustrates that $\mathbf{p}$ is far from perfectly random,
and, by identifying the dependencies that are present in $\mathbf{p}$, we may be able to save a significant amount of bits!
If we manage to perfectly approximate $\mathbf{p}$, then we expect $H(\mathbf{p},\mathbf{p}_{\mathcal{B}}) = H(\mathbf{p},\mathbf{p}) = H(\mathbf{p}) + D_{KL}(\mathbf{p}||\mathbf{p}) = 5.12 + 0 = 5.12$ bits.

In all our experiments, we average each measurement over $100$ independent repetitions.
In each repetition, we use \emph{prior sampling} to generate a synthetic dataset of $n \approx 40,000$ data points, according to the distribution we described.
We partition the $n$ data points among the $M$ data holders using a \emph{round-robin partitioning} scheme, so that each data holder has $\lfloor \frac{n}{M} \rfloor$ data points.

\begin{figure}%[H]%[ht!]
    \centering
    \includegraphics[width=0.75\columnwidth]{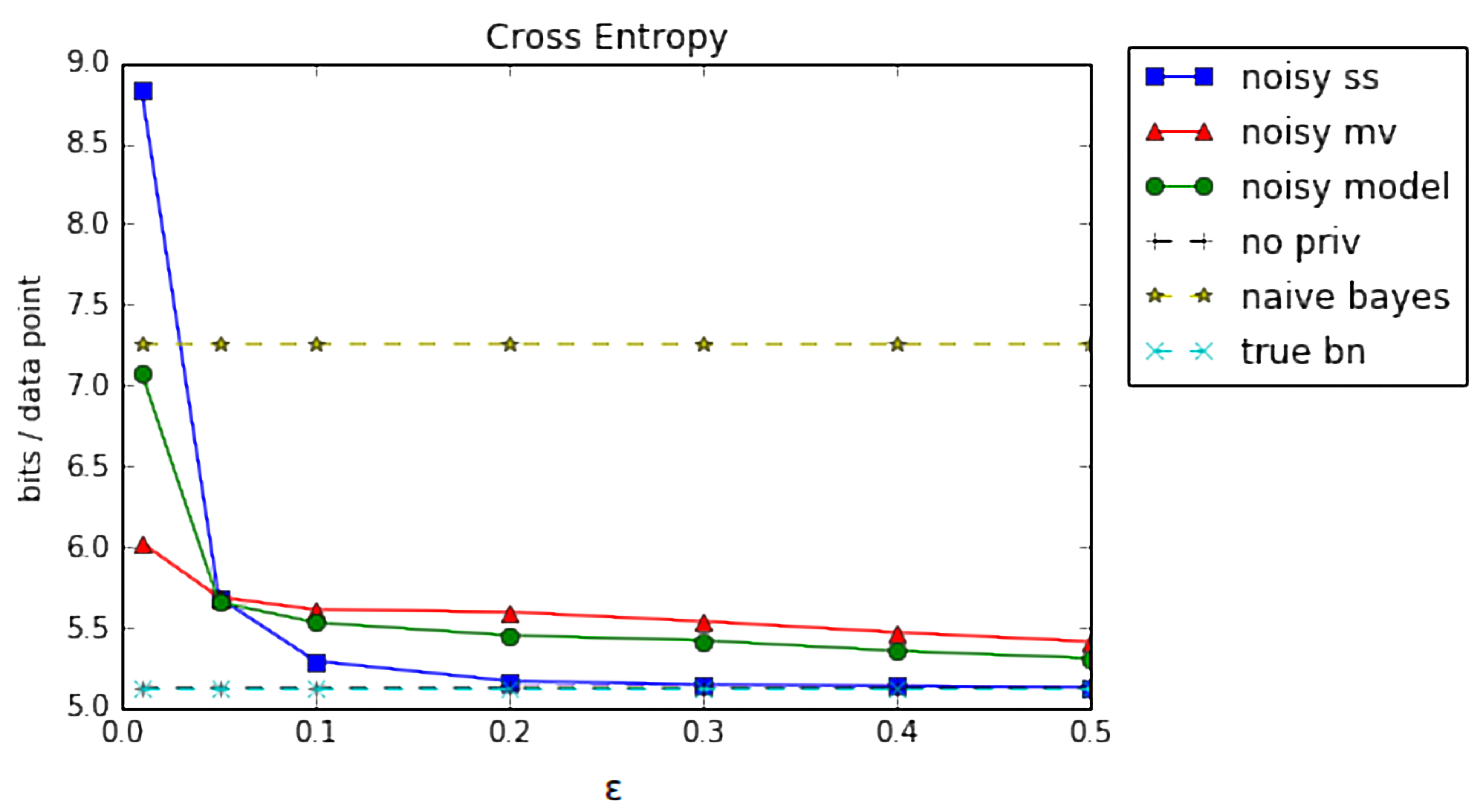}
    \caption{Cross entropy for varying privacy budget $\eps$.\\Number of data holders: $M=3$. Bayesian Network degree: $k=1$.}
	\label{FIG_exp1}
\end{figure}

\begin{figure}%[ht!]
    \centering
    \includegraphics[width=0.75\columnwidth]{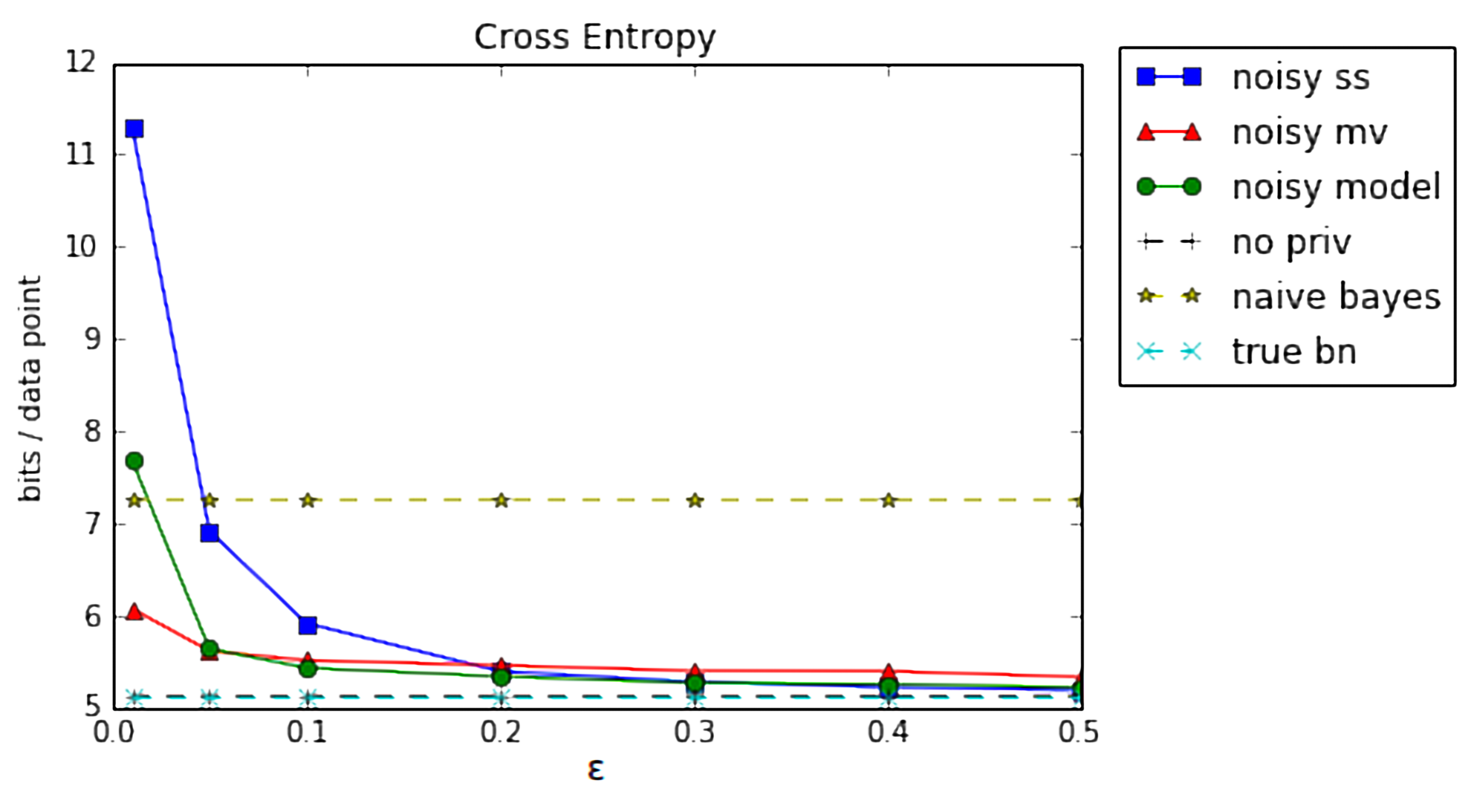}
    \caption{Cross entropy for varying privacy budget $\eps$.\\Number of data holders: $M=3$. Bayesian Network degree: $k=2$.}
	\label{FIG_exp1b}
\end{figure}

In Figures \ref{FIG_exp1} and \ref{FIG_exp1b} we observe that all algorithms perform significantly better than the naive-Bayes, and approach the optimum value of bits per data point of the true Bayesian Network as $\eps$ increases;
Algorithm \ref{Sharing the Noisy Sufficient Statistics} (noisy ss) asymptotically outperforms the other algorithms, but is more sensitive to noise.
We also notice that the non-private algorithm accurately learns the true Bayesian Network.

\begin{figure}%[H]%[ht!]
    \centering
    \includegraphics[width=0.75\columnwidth]{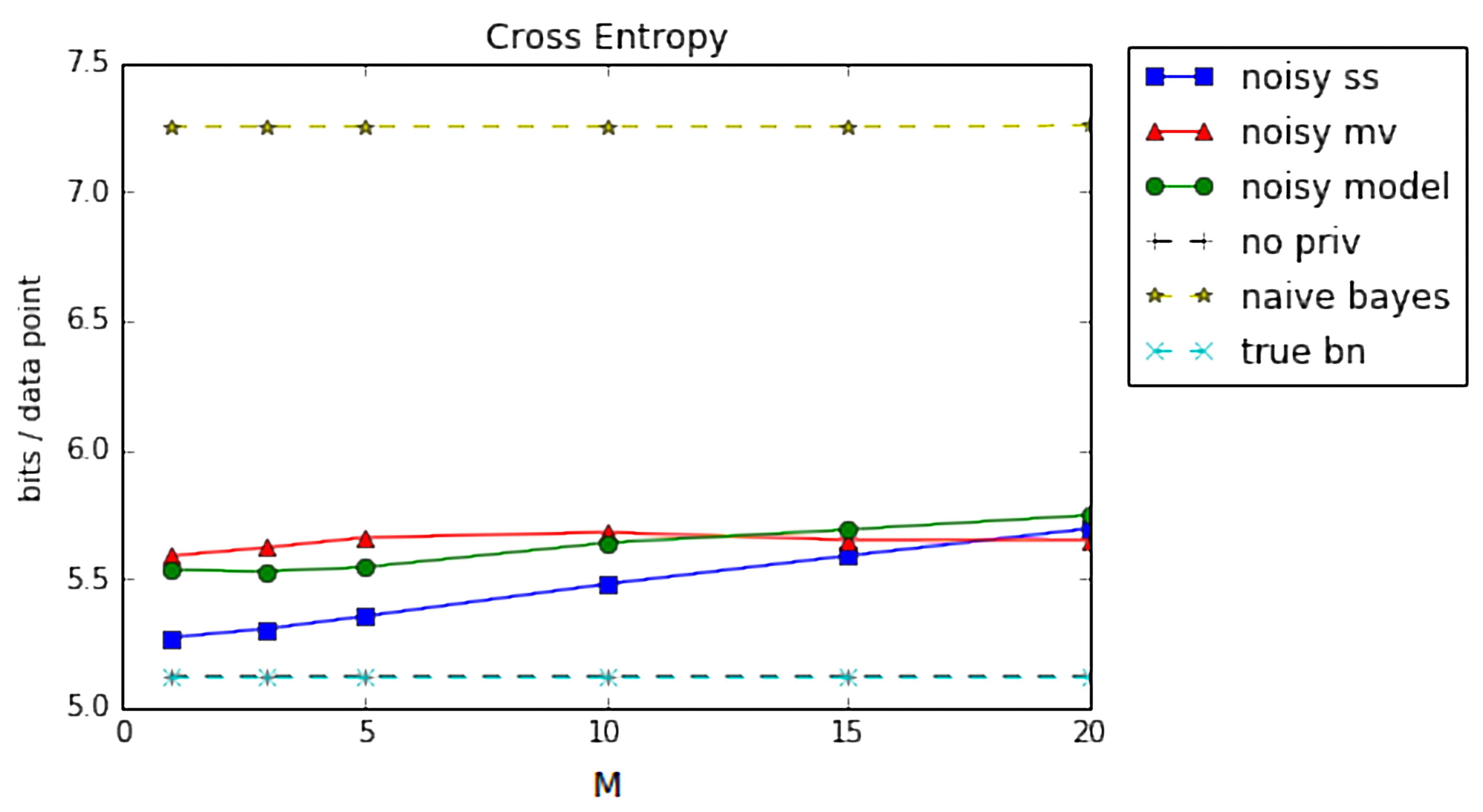}
    \caption{Cross entropy for varying number of data holders $M$.\\Privacy budget: $\eps=0.1$. Bayesian Network degree: $k=1$.}
	\label{FIG_exp2}
\end{figure}

The key thing to notice in Figure \ref{FIG_exp2} is the robustness of Algorithm \ref{Noisy Majority Voting} to the number of data holders.
On the contrary, the performance of Algortihms \ref{Sharing the Noisy Sufficient Statistics} and \ref{Sharing the Noisy Model} deteriorates as the number of data holders increases.

\begin{figure}%[H]%[ht!]
    \centering
    \includegraphics[width=0.75\columnwidth]{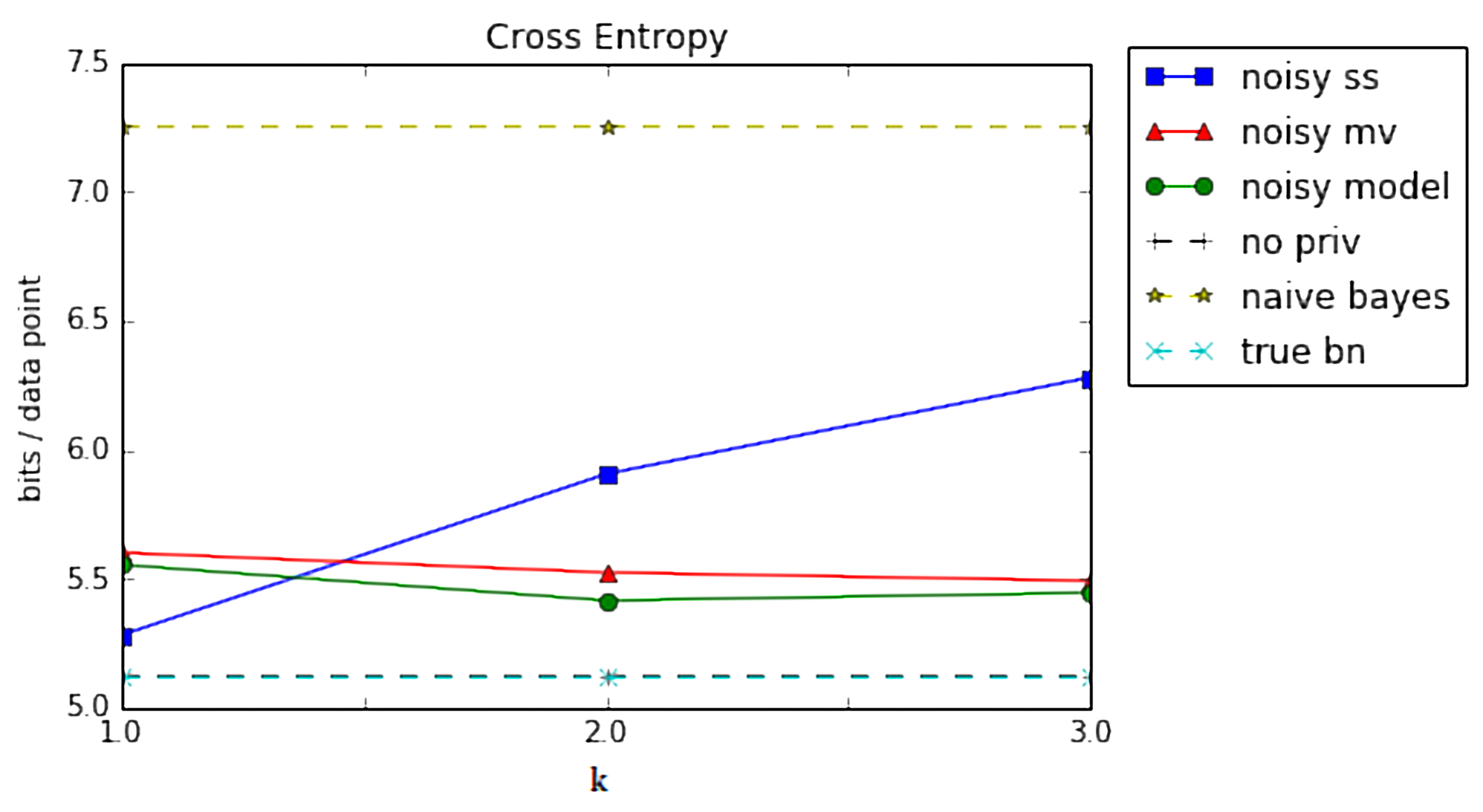}
    \caption{Cross entropy for varying Bayesian Network degree.\\Privacy budget: $\eps=0.1$. Number of data holders: $M=3$.}
	\label{FIG_exp5}
\end{figure}

Finally, comparing Figures \ref{FIG_exp1} and \ref{FIG_exp1b}, and as we also observe in Figure \ref{FIG_exp5},
as the Bayesian Network degree increases, the performance of Algorithm \ref{Sharing the Noisy Sufficient Statistics} (noisy ss) worsens.
This is expected, since a higher degree Bayesian Network implies the use and perturbation of higher dimensional distributions.
In contrast, Algorithms \ref{Noisy Majority Voting} (noisy mv) and \ref{Sharing the Noisy Model} (noisy model) slightly improve as the degree increases.

\subsection{Real-World Data}
In our second set of experiments, we utilize \emph{real-world data} from the Diabetes dataset (\cite{strack2014impact}) of the UCI Machine Learning Repository.

The dataset was extracted from a large database that represents 10 years of clinical care at US hospitals and includes over 50 attributes representing patient and hospital outcomes.
In particular, a total of 101,766 records (data points) were selected, based on the following inclusion criteria:
\begin{itemize}
\item[-] It is an inpatient encounter (a hospital admission).
\item[-] It is a diabetic encounter, that is, one during which any kind of diabetes was entered to the system as a diagnosis.
\item[-] The length of stay was at least 1 day and at most 14 days.
\item[-] Laboratory tests were performed during the encounter.
\item[-] Medications were administered during the encounter.
\end{itemize}
Next, attribute (feature) selection was performed (by clinical experts) and only attributes that were potentially associated with the diabetic condition or management were retained (55 attributes, describing the diabetic encounters, including demographics, diagnoses, diabetic medications, number of visits in the year preceding the encounter, and payer information).
The final attributes in each record include (among others): patient id, race, gender, age, admission type, time in hospital, medical specialty of admitting physician, number of lab test performed, HbA1c test result, diagnosis, number of medication, diabetic medications, number of outpatient, inpatient, emergency visits in the year before the hospitalization, etc.

We perform additional pre-processing to the dataset, as we describe in the next three steps.
\begin{itemize}
\item[-] First, we remove identification-related attributes (e.g. patient id) and attributes with a high percentage of missing values (e.g. weight - 97 percent of values are missing).
We are left with 43 attributes.
\item[-] Second, we identify 3 diagnosis-related attributes with large domains (almost 1,000 values per attribute) and compress their domains into 9 general groups of diagnoses, as described in \cite{strack2014impact}.
The final domain sizes vary from 2 to 118; we have a total of almost $10^{32}$ possible tuples, so (inevitably) we will not observe a huge number of events. 
\item[-] Third, we remove records that refer to the same patient.
Such records cannot be considered statistically independent, and, also, differential privacy is guaranteed at a record level, so the privacy of a patient that has multiple records is violated.
We note that one solution that would allow each patient to have multiple records is the group differential privacy guarantee.
We are left with approximately 70,000 records.
\end{itemize}

\begin{figure}%[H]%[ht!]
    \centering
    \includegraphics[width=0.75\columnwidth]{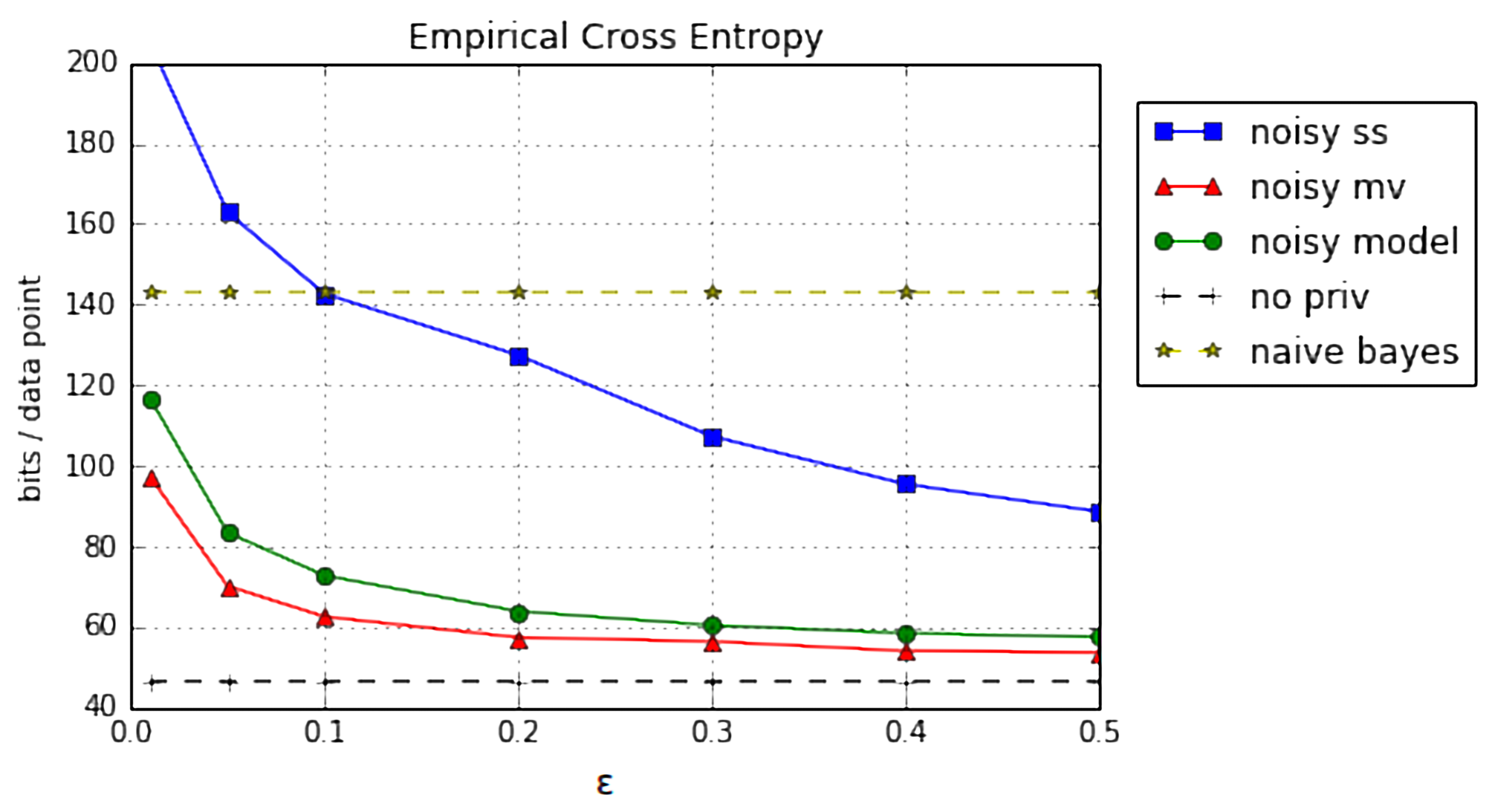}
    \caption{Cross entropy for varying privacy budget $\eps$.\\Number of data holders: $M=3$. Bayesian Network degree: $k=1$.}
	\label{FIG_exp3}
\end{figure}

In all our experiments, we average each measurement over $20$ independent repetitions. We again split the data among the data holders using a round-robin partitioning scheme.
We first experiment for varying privacy budget $\eps$ (Figure \ref{FIG_exp3}).
We observe that Algorithm \ref{Sharing the Noisy Sufficient Statistics} (noisy ss) is significantly worse; it appears that it is highly sensitive to the data dimension and the domain sizes of the attributes,
which are significantly increased compared to the synthetic data we experimented with.
Algorithm \ref{Noisy Majority Voting} (noisy mv) outperforms the other two algorithms and approaches the performance of the non-private algorithm.

\begin{figure}%[H]%[ht!]
    \centering
    \includegraphics[width=0.75\columnwidth]{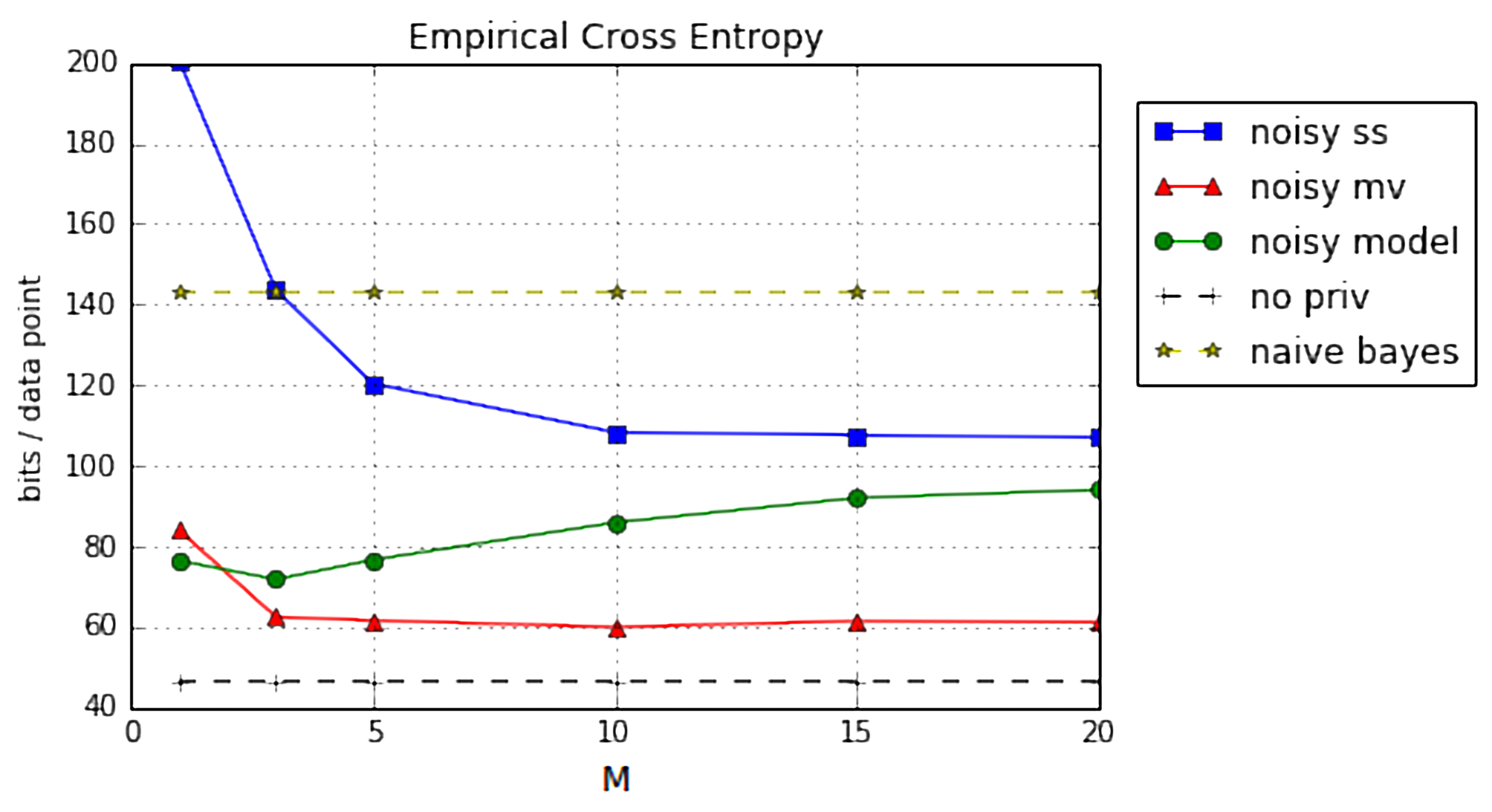}
    \caption{Cross entropy for varying number of data holders $M$.\\Privacy budget: $\eps=0.1$. Bayesian Network degree: $k=1$.}
	\label{FIG_exp4}
\end{figure}

We next experiment for varying number of data holders.
Figure \ref{FIG_exp4} confirms the robustness of Algorithm \ref{Noisy Majority Voting} (noisy mv) to the number of data holders,
and validates that -for complex datasets- Algorithm \ref{Noisy Majority Voting} (noisy mv) is the clear winner.
The performance of Algorithm \ref{Sharing the Noisy Model} (noisy model) significantly worsens as number of holders increases.
When the dataset is split among a larger number of data holders, each holder possesses a smaller dataset, and therefore the quality of the local models deteriorates.
Finally, the behavior of Algorithm \ref{Sharing the Noisy Sufficient Statistics} (noisy ss) may at first seem strange, as it differs from the algorithm's behavior in the corresponding experiment with synthetic data (Figure \ref{FIG_exp2}).
This behavior illustrates that when the data dimension and the domain sizes of the attributes are both large, the impact of the number of data holders is not critical.
A possible explanation for the improvement in performance is that, as the number of data holders increases, more zero-mean noise terms are added to the frequency distributions, and hence the noise cancels out.

\section{Discussion}
To summarize, in this chapter we addressed the problem of distributed learning of Bayesian Networks with differential privacy.
After formally describing our model and identifying the challenges that arise when moving from a centralized to a distributed environment,
we examined three solutions, namely Sharing the Noisy Sufficient Statistics, Noisy Majority Voting and Sharing the Noisy Model.

Our detailed experimental evaluation indicates that Noisy Majority Voting is robust to all the parameters we examined (privacy budget, number of data holders, Bayesian Network degree),
and significantly outperforms the other algorithms when the data are high-dimensional.
We also note that the behavior of Sharing the Noisy Sufficient Statistics is consistent with our theoretical analysis,
which predicted that:
\begin{itemize}
\item[-] When the privacy budget $\eps$ exceeds a certain threshold, our differentially private entropy estimates converge to the non-private ones,
and thus the learned distribution accurately approximates the non-private one.
\item[-] When either the number of data holders $M$ or the Bayesian Network degree $k$ increases, our differentially private entropy estimates worsen, and so does the performance of the algorithm.
\end{itemize}

\chapter{Pan-Private Stream Density Estimation}
We work in the \emph{data stream model} (\cite{garofalakis2016data}, \cite{cormode2011synopses}).
The input is represented as a finite sequence of tuples (from some finite domain) which is not available for random access, but instead arrives one at a time in a stream.
We thus deal with dynamic data, and we assume that each input tuple corresponds to some user from a universe $\mathcal{U}$, who is identified by a unique key (user id).

More specifically, in what follows, we consider the \emph{cashier-register} streaming model:
each update is of the form \textit{(user id, update value)}, where the update values are strictly positive integers.
For further simplicity, we permit only unit updates, so the stream can be viewed as a sequence of user id's.
Each user is mapped to a state that accumulates the user's updates.
We note that, although our work does not apply to the (most general) turnstile streaming model, where negative updates are also allowed,
it can be extended to the special case where both unit insertions and unit deletions are allowed, 
and, at any time step, each user's state can be either 1 (user is present) or 0 (user is absent).

In general, \emph{streaming algorithms} should be able to operate in a single pass (each tuple is examined at most once in fixed arrival order) and in real-time (each tuple's processing time must be low), and should require small space.
For most problems, it is not possible to offer exact answers while satisfying these requirements, so a common approach is to offer $(\alpha,\beta)$-approximate answers, 
that is, with probability at least $1-\beta$ the computed answer is within an additive / multiplicative factor $\alpha$ of the actual answer. \\

The particular problem we examine is the following:
given an input data stream, we want to estimate its \emph{density}, 
that is, the fraction of $\mathcal{U}$ that actually appears in the stream.
This statistic is closely related to the well-studied distinct count (a.k.a. $0^{th}$ frequency moment and Hamming norm), which expresses the number of distinct users that appear in the stream.
The density is more convenient in that it is normalized and takes values in the $[0,1]$ interval.

Additionally, given that the focus of this thesis is the design of privacy-preserving algorithms, we also want to protect the \emph{privacy} of the users that appear in the stream.
We would like to offer a strong privacy guarantee, such as differential privacy, adjusted to the streaming model. \\

Before proceeding with the technical part of this chapter, we briefly present two potential applications, that motivate the study of this problem.
\begin{itemize}
\item[-] We borrow this example from Dwork \cite{dwork2010differential}.
Consider a website for an \emph{epidemic disease self-assessment}.
Users interact with the website to figure out whether the symptoms they are experiencing may be indicative of the disease.
An epidemiological study would be interested in answering queries, like the following:
{\small
\begin{multicols}{2}
\begin{verbatim}
SELECT COUNT( DISTINCT ip_addr )
FROM data_stream
\end{verbatim}
\columnbreak
\begin{verbatim}
SELECT COUNT( DISTINCT ip_addr )
FROM data_stream
WHERE symptom1 = "YES", ...
\end{verbatim}
\end{multicols}
}
An individual's medical record is undoubtedly an extremely sensitive piece of information,
so knowing even the presence of a user in such a website would constitute a major privacy breach.

\item[-] We mentioned that our work can be extended to support both unit insertions and deletions. 
This extension is especially useful in the context of \emph{dynamic graphs}; user id's are replaced by edge id's, and each update either inserts or deletes an edge.
An example sequence of updates is the following: $$( \ (edge_i,insert),\ (edge_j,insert),\ (edge_i,delete),\ ... \ )$$

The density of an undirected graph $\mathcal{G}=(V,E)$ is defined as: $d(\mathcal{G}) = \frac{2|E|}{|V|(|V|-1)}$ and expresses the fraction of edges that are present.
For a dynamic graph, created through a stream $\mathcal{S}$, it is not hard to see that $d(\mathcal{G}) = d(\mathcal{S})$.

If our focus is, for example, on social graphs, which depict personal relations of Internet users, 
then any analysis performed on the graph (such as the density estimation) must not leak information about the existence of individual edges.
\end{itemize}

\begin{table}[H]%[!t]
\renewcommand{\arraystretch}{1.3}
\caption{Table of notations}
\label{table notation}
\centering
\begin{tabular}{|c||c|}
\hline
\textbf{Notation} & \textbf{Description}\\
\hline
\hline
$\mathcal{U}$ & Universe of users, w.l.o.g.: $\mathcal{U} = \{1,...,|\mathcal{U}|\}$\\
\hline
$T$ & Stream length\\
\hline
$s_t$ & Stream update at time step $t$: $t\in\{1,...,T\},\ s_t \in \mathcal{U}$\\
\hline
$\mathcal{S}$ & Input stream: $\mathcal{S}=(s_1,...,s_T)$ \\
\hline
$a_u(t)$ & State of user $u\in\mathcal{U}$ after t updates: $a_u(t) = \sum_{i=1}^t{\mathbf{1}(s_i=u)}$ \\
\hline
$\mathbf{a}(t)$ & State vector after t updates: $\mathbf{a}(t)=[a_1(t)\ ... \ a_{|\mathcal{U}|}(t)]$ \\
\hline
$F_0(\mathcal{S})$ & Number of distinct users in $\mathcal{S}$: $F_0(\mathcal{S}) = ||\mathbf{a}(T)||_0$\\
\hline
$d(\mathcal{S})$ & Density of $\mathcal{S}$: $d(\mathcal{S})=\frac{F_0(\mathcal{S})}{|\mathcal{U}|}$ \\
\hline
\end{tabular}
\end{table}

\section{Differential Privacy in the Streaming Model}
So far, we have considered differential privacy on static datasets, and have assumed that an adversary only has access to the output of the privacy-preserving algorithm.
In the model we examine in this chapter, where the data arrive dynamically in a stream, \emph{two privacy models} (definitions) have been developed (\cite{dwork2010pan},\cite{dwork2010continual},\cite{dwork2010differential}).

\begin{itemize}
\item[-] \textbf{Pan-Privacy.}
In pan-privacy, the internal state of the algorithm is also differentially private, as is the joint distribution of the internal state and the output.
This protects the privacy of the data in case the data holder is subject to compulsory, non-private data release, because of a subpoena, or faces an intrusion from a hacker.

\item[-] \textbf{Continual Observation.}
In continual observation, the goal is to continually monitor and report statistics about events that occur dynamically in discrete time steps, and to ensure that the output sequence is differentially private.
This is the case, for example, when monitoring incidence of influenza, traffic conditions and search trends.
\end{itemize}

The two notions provide orthogonal guarantees, and some work addresses pan-privacy under continual observation. Let us present a simple example:\\\\
\textbf{Example.} Consider an algorithm that takes as input a data stream $\mathcal{S}=(s_1,...,s_T)$, and upon arrival of each element $s_i$, outputs some estimate $f_i = f(s_1,...,s_i)$. 
Pan-privacy ensures that the state of the algorithm is subject to a differential privacy constraint, so privacy is preserved against an adversary that observes the internal state and a (single) output of the algorithm, say $f_j$ for some $j \in \{1,...,T\}$. We can think of this single output as the final output, so that $j=T$.
Differential privacy under continual observation ensures that the entire sequence of outputs $f_1,...,f_T$ satisfies differential privacy.
The two definitions can be combined, by ensuring that the internal state, the output sequence and their joint distribution, all satisfy differential privacy. \\

A very important notion in differential privacy is that of \emph{adjacency};
as we already argued, two (static) datasets are adjacent if they differ on a single record.
But how should we define adjacency in the streaming model?
When are two data streams considered to be adjacent?
We (informally) introduce the following two notions:
\begin{itemize}
\item[-] \textbf{Event-level privacy.}
Two data streams are considered event-level adjacent if they differ on a single update $s_t$ for some $t \in \{1,...,T\}$.
Event-level privacy protects the privacy of $s_t$ and an adversary cannot distinguish whether $s_t$ did or did not appear.
In the epidemic disease self-assessment website, event-level privacy protects the privacy of single visits to the website.
In a dynamic graph, event-level privacy protects the privacy of a single insertion / deletion of an edge between any two nodes.

\item[-] \textbf{User-level privacy.}
Two data streams are considered user-level adjacent if they differ on all (or some) updates that refer to a user $u \in \mathcal{U}$.
User-level privacy protects the privacy of $u$ and an adversary cannot distinguish whether $u$ did or did not ever appear, independently of the actual number of appearances of $u$.
In the epidemic disease self-assessment website, user-level privacy protects the privacy of all the visits of a user (IP-address) to the website.
In a dynamic graph, user-level privacy protects the privacy of any edge, regardless of how many times this edge has been inserted \ deleted.
\end{itemize}
As in the case of static datasets, the word \enquote{differ} in the phrases \enquote{differ on a single update} and \enquote{differ on all (or some) updates that refer to a user} can be interpreted in two ways.
According to the first interpretation, the different updates are present in the one data stream and absent in the other;
in the traditional differential privacy model, this interpretation would lead to unbounded differential privacy.
According to the second interpretation, the different updates are present in both data streams, but refer to two different users;
in the traditional differential privacy model, this interpretation would lead to bounded differential privacy.

The privacy level affects the amount of perturbation used, so a much stronger guarantee like user-level privacy requires excessively more perturbation.
Kellaris et al. \cite{kellaris2014differentially} attempt to bridge the gap between event-level and user-level privacy and develop a framework in-between;
they introduce $w$-event privacy, which protects any event sequence occurring within a window of $w$ time steps. \\

For completeness, and before proceeding with our specific problem (density estimation),
we present the basic definitions for both pan-privacy and continual observation, along with some related work.

\subsection{Pan-Privacy}
As we already argued, the definition of pan-privacy aims to protect against an adversary that can, on rare occasions, observe the internal state of the algorithm.
This is, of course, in addition to the standard -for differential privacy- assumptions about the adversary having arbitrary control over the input, arbitrary prior knowledge and arbitrary computational power.

Before formally defining the model, we remark that ordinary streaming algorithms based on sampling and sketching techniques do not provide the pan-privacy guarantee.
Sampling techniques maintain information about a subset of the users, so an intruder with access to the sample (the internal state of the algorithm) would violate the privacy of the sampled users.
Sketching techniques, like the FM Sketch \cite{flajolet1985probabilistic} and the Count-Min Sketch \cite{cormode2005improved}, which are based on hashing also cannot protect the privacy of the users against an adversary who has access to the hash functions used; the hash functions are part of the internal state.

\begin{defn}[Adjacency]\label{def_adj_1}
Data streams $\mathcal{S}$ and $\mathcal{S'}$ are (user-level) adjacent, denoted as $adj(\mathcal{S},\mathcal{S'})$, if they differ only in the presence or absence of all (or some) occurrences of a single user $u \in \mathcal{U}$.
\end{defn}
Notice that in Definition \ref{def_adj_1}, we use the first interpretation of the word \enquote{differ} in defining adjacent data streams, following Dwork et al. \cite{dwork2010pan}.
Therefore, two adjacent data streams cannot have the same length.
Mir et al. \cite{mir2011pan}, for instance, adopt the second interpretation, as one of the pan-private algorithms they propose requires that the total sum of updates over all users stays the same;
they replace all (or some of) the updates that refer to user $u$ by updates that refer to another user $u'$ (with the same total sum). 

\begin{defn}[Pan-privacy]\label{def_pan_privacy}
Let $\textbf{Alg}$ be a randomized algorithm.
Let $\mathcal{I}$ denote its set of internal states, and let $\mathcal{O}$ denote its set of possible outputs.
Then $\textbf{Alg}$, mapping data streams of finite length $T$ to the range $\mathcal{I} \times \mathcal{O}$, is $\eps$-pan-private against a single intrusion,
if for all sets $I \subseteq \mathcal{I}$ and $O \subseteq \mathcal{O}$, and for all pairs of adjacent data streams $\mathcal{S},\mathcal{S'}$
$$ \Prob[\textbf{Alg}(\mathcal{S}) \in (I,O) ] \ \leq \ e^{\eps} \Prob[\textbf{Alg}(\mathcal{S'}) \in (I,O) ]$$
where the probability space is over the coin flips of $\textbf{Alg}$.
\end{defn}

We remark that Definition \ref{def_pan_privacy} speaks only of a single intrusion. 
To handle multiple intrusions, we must consider interleavings of observations of internal states and output sequences.
We also have to differentiate between announced and unannounced intrusions.
In the former case (e.g. subpoena), the algorithm knows that an intrusion occurred, so it can re-randomize its state and handle multiple announced intrusions.
In the latter case (e.g. hacking), the algorithm can only tolerate a single unannounced intrusion and strong negative results have been proved for even two unannounced intrusions. \\

So far, two works have addressed the challenge of developing pan-private streaming algorithms.
They both examine variants of the same problems, applying different techniques.
Dwork et al. \cite{dwork2010pan} work in the cashier-register streaming model and develop algorithms based on sampling and randomized response.
Mir et al. \cite{mir2011pan} work in the turnstile streaming model and rely on both existing and novel sketches; they also develop a general noise-calibrating technique for sketches. 
The problems examined are the following:
\begin{itemize}
\item[-] Density estimation / Distinct count: the fraction of $\mathcal{U}$ that appears / the number of users with nonzero state.

\item[-] $t$-cropped mean / $t$-cropped first moment: the average / sum, over all users, of the minimum of $t$ and the number of appearances of the user.

\item[-] Fraction of $k$-heavy hitters / $k$-heavy hitters count: the fraction / number of users that appear at least $k$ times.

\item[-] $t$-incidence estimation: the fraction of users that appear exactly $t$ times.
\end{itemize}

\emph{In our work, we aim to offer the user-level pan-privacy guarantee (against a single intrusion).}
Our objective is to estimate the density of the given input stream,
and our algorithms are based on the density estimator of Dwork et al. \cite{dwork2010pan}.
The algorithm proposed by Mir et al. \cite{mir2011pan} for the (similar) distinct count problem relies on the $\ell_0$ Sketch, which is due to Cormode et al. \cite{cormode2002comparing} and utilizes a family of distributions called $p$-stable (Indyk \cite{indyk2000stable}).
Despite being particularly interesting theoretically, the pan-private algorithm of Mir et al. \cite{mir2011pan} is extremely impractical as it involves an application of the exponential mechanism, which requires sampling twice from a space of $2^{S}$ sketches (where $S$ is the bit size of the sketch).
To make matters worse, in order to evaluate the exponential mechanism's scoring function, a Hamming norm minimization problem is solved for every single sketch, which translates to solving $2^S$ problems.

\subsection{Continual Observation}
As we already stated, differential privacy under continual observation addresses the need for privacy in applications that involve repeated computations over dynamic data, and require continually producing updated outputs.
We stick to Definition \ref{def_adj_1} for adjacent data streams, which means that the following definition (\ref{def_cont_obs}) refers to user-level differential privacy under continual observation.

\begin{defn}[Differential privacy under continual observation]\label{def_cont_obs}
Let $\textbf{Alg}$ be a randomized algorithm,
and let $\mathcal{O}$ denote its set of possible output sequences.
Then $\textbf{Alg}$, mapping data streams of finite length $T$ to the range $\mathcal{O}$, is $\eps$-differentially private under continual observation
if for all sets $O \subseteq \mathcal{O}$, for all pairs of adjacent data streams $\mathcal{S},\mathcal{S'}$, and for all $t \in \{1,...,T\}$,
$$ \Prob[\textbf{Alg}(\mathcal{S}(t)) \in O ] \ \leq \ e^{\eps} \Prob[\textbf{Alg}(\mathcal{S'}(t)) \in O ]$$
where the probability space is over the coin flips of $\textbf{Alg}$.
\end{defn}

Although our focus is on algorithms that produce a single output, we remark that Dwork \ref{dwork2010differential} shows how to modify the pan-private density estimator developed by Dwork et al. \cite{dwork2010pan} (which, we repeat, is the baseline for our work), to produce output continually.
The resulting continual observation density estimator guarantees \emph{user-level pan-privacy under continual observation}.
A similar technique can be applied to all the algorithms we develop to allow them to produce continual output while still preserving privacy. \\

Dwork \cite{dwork2010differential} and Dwork et al. \cite{dwork2010continual} initiate the study of differential privacy under continual observation;
they examine the problem of continually releasing differentially private counts (both event-level and user-level), and they also optionally guarantee pan-privacy.
Chan et al. \cite{chan2010private} address the exact same counter problem (independently), but they only focus on event-level privacy, and
Bolot et al. \cite{bolot2013private} extend their work by proposing decayed sums, which emphasize on recent data more than the data of the past (in addition to proposing a new notion of decayed privacy).
Chan et al. \cite{chan2012differentially} report the heavy hitters instead of the counter values in various models (e.g. untrusted aggregator), also ensuring event-level privacy.

A few works examine problems beyond counting.
Fan et al. \cite{fan2012real} propose a framework to release real-time aggregate statistics under differential privacy, based on filtering and adaptive sampling; they focus on user-level privacy.
The next authors examine event-level privacy.
Friedman et al. \cite{friedman2014privacy} develop a framework for privacy-preserving distributed stream monitoring based on the geometric method \cite{sharfman2007geometric}, which enables monitoring arbitrary functions over statistics derived from distributed data streams.
Bonomi et al. \cite{bonomi2016differentially} compute the longest increasing subsequence in a differentially private manner, and
Upadhyay \cite{upadhyay2014differentially} examines the connections between $(\eps,\delta)$-differential privacy and linear algebra in the streaming model, for low-rank approximation, linear regression and matrix multiplication.

\section{Dwork's Density Estimator}
Dwork et al. \cite{dwork2010pan} proposed the first user-level pan-private algorithm for the density estimation problem (Algorithm \ref{DworkDE}). In this section, we provide a detailed presentation and analysis of their algorithm, which we call Dwork's Density Estimator.

Their (randomized) algorithm takes as input the data stream $\mathcal{S}$ whose density we wish to estimate, as well as the desired privacy budget and accuracy parameters.
For simplicity, we assume that the length of the stream is known in advance (that is, we know that the output will be requested after $T$ updates), but this assumption does not affect the analysis of the algorithm, and can easily be dropped, so that the algorithm outputs the stream density when it receives a special signal.

The first random choice is a sampling step - the algorithm maintains information only about a random subset of the users, in order to keep its state small.
The state of the algorithm is a bitarray with one entry per sampled user. 
The entries are random bits, generated as described below; this is the second random choice of the algorithm.
For users that have not appeared in $\mathcal{S}$, their entry is drawn from a uniform Bernoulli distribution, while for users that have appeared, their entry is drawn from a slightly biased (towards $1$) Bernoulli distribution, no matter how many times they have appeared. 
The two distributions should be close enough to guarantee that the state satisfies differential privacy, but far enough to allow collection of aggregate statistics about the fraction of users that appear at least once.
The last random choice is performed via the use of the Laplace mechanism, when outputting the final density estimate, which guarantees that the output also satisfies differential privacy.

\begin{algorithm}%[H]

\caption{Dwork's Density Estimator\label{DworkDE}}

\DontPrintSemicolon

\KwIn{Data stream $\mathcal{S}$, Privacy budget $\eps$, Accuracy parameters $(\alpha,\beta)$}

\KwOut{Density $\tilde{d}(\mathcal{S})$}

Pick $m = \mathcal{O}(\frac{1}{\eps^2 \alpha^2}\log{\frac{1}{\beta}})$
\ \ \textit{(or compute $m^*$ as described in subsection \ref{subSEC_sample_size} and set $m=m^*$)}\;

Sample a random subset $M \subseteq \mathcal{U}$ of $m$ users (without replacement) and define an arbitrary ordering over $M$\;

Create a bitarray $\mathbf{b}=[b_1\ ...\ b_{m}]$ and map $M[i] \rightarrow b_i,\ \forall \ i \in \{1,...,m\}$\;

Initialize $\mathbf{b}$ randomly: $b_i \sim \text{Bernoulli}(\frac{1}{2}), \ \forall \ i \in \{1,...,m\}$\;

\For{t = 1 \KwTo T}{

	\If{$s_t \in M$}{
		
		Find $i:\ M[i]=s_t$\;
		
		Re-sample: $b_i \sim \text{Bernoulli}(\frac{1}{2}+\frac{\eps}{4})$\;
			
	}
}

Return $\tilde{d}(\mathcal{S})\ = \ \frac{4}{\eps}(\frac{1}{m}\sum_{i=1}^{m}{b_i}-\frac{1}{2}) \ + \ \text{Laplace}(0,\frac{1}{\eps m})$\;

\end{algorithm}

We now make two important remarks concerning potential extensions of Algorithm \ref{DworkDE}.
The techniques described in the remarks apply (slightly modified) to all the algorithms we present, so we do not revisit them in our work.

\begin{itemize}
\item[-] Algorithm \ref{DworkDE} can tolerate a single (announced or unannounced) intrusion. Dwork et al. \cite{dwork2010pan} show how to handle multiple announced intrusions, by re-randomizing the state after each intrusion has occurred.

\item[-] Algorithm \ref{DworkDE} works in the cashier-register streaming model, that is, once a user appears in the stream, it cannot be deleted.
However, as we mentioned in the introduction of this chapter, our work also applies to the case where a user $u$ may be both inserted and deleted (later on) from the stream.
In particular, if an update of the form $(u,insert)$ arrives, $u$'s bit is drawn $\text{Bernoulli}(\frac{1}{2}+\frac{\eps}{4})$, whereas if an update of the form $(u,delete)$ arrives, $u$'s bit is re-drawn $\text{Bernoulli}(\frac{1}{2})$.
This allows Algorithm \ref{DworkDE} to perform pan-private graph density estimation as well.
\end{itemize}

\subsection{Privacy Analysis}
We first examine the privacy guarantees of Algorithm \ref{DworkDE}. We have the following theorem.

\begin{thm}\label{thm_DworkDE_priv}
Assume $\eps \leq \frac{1}{2}$. Then Algorithm \ref{DworkDE} satisfies $2\eps$-pan-privacy.
\end{thm}

\begin{tcolorbox}[breakable]
\begin{proof}
Let $\mathcal{S},\mathcal{S'}$ be two adjacent streams that differ on all occurrences of user $u \in \mathcal{U}$. Assume w.l.o.g. that $u \in \mathcal{S}$ and $u \not\in \mathcal{S'}$.

Firstly, the \emph{state} of Algorithm \ref{DworkDE} satisfies $\eps$-differential privacy. All the information that Algorithm \ref{DworkDE} stores as its state is the bitarray $\mathbf{b}$. We distinguish the following two cases:

\begin{itemize}
\item[-] If $u \not\in M$: perfect privacy is guaranteed, as no information is stored on user $u$.

\item[-] If $u \in M$: let (w.l.o.g.) $b_1$ be the entry that corresponds to $u$ in the bitarray. Then, assuming that $u$ has already appeared in the stream when the adversary views $\mathbf{b}$, $b_1(\mathcal{S})$ is drawn from $\text{Bernoulli}(\frac{1}{2}+\frac{\eps}{4})$ and $b_1(\mathcal{S'})$ is drawn from $\text{Bernoulli}(\frac{1}{2})$. We thus have to bound the following probability ratios according to the differential privacy definition:

\begin{eqnarray*}
\frac{\Prob( \ \mathbf{b(\mathcal{S})}=[1 \ \underline{b_2} \ ...\ \underline{b_{m}}]\ )}{\Prob( \ \mathbf{b(\mathcal{S'})}=[1 \ \underline{b_2} \ ...\ \underline{b_{m}}]\ )} &=& \frac{\Prob( \ b_1(\mathcal{S})=1\ )}{\Prob( \ b_1(\mathcal{S'})=1\ )} \\
&=& \frac{\frac{1}{2}+\frac{\eps}{4}}{\frac{1}{2}} \ = \ 1+\frac{\eps}{2} \\
\frac{\Prob( \ \mathbf{b(\mathcal{S})}=[0 \ \underline{b_2} \ ...\ \underline{b_{m}}]\ )}{\Prob( \ \mathbf{b(\mathcal{S'})}=[0 \ \underline{b_2} \ ...\ \underline{b_{m}}]\ )} &=& \frac{\Prob( \ b_1(\mathcal{S})=0\ )}{\Prob( \ b_1(\mathcal{S'})=0\ )} \\
&=& \frac{\frac{1}{2}-\frac{\eps}{4}}{\frac{1}{2}} \ = \ 1-\frac{\eps}{2}
\end{eqnarray*}

Noting that $ e^{-\eps} \leq 1+\frac{\eps}{2} \leq \sum_{k=0}^{\infty}{{\eps^k}{k!}} = e^{\eps} $ and  $ e^{-\eps} \leq 1-\frac{\eps}{2} \leq e^{\eps} \ , \ \forall \ \eps \in [0,\frac{1}{2}]$, it becomes clear that user $u$ is guaranteed $\eps$-differential privacy against an adversary that observes $u$'s entry in $\mathbf{b}$.

\end{itemize}

The \emph{output} of Algorithm \ref{DworkDE} (conditioned on the state) also satisfies $\eps$-differential privacy, as it is computed by independently applying the Laplace mechanism.
Specifically, the sensitivity of $\tilde{d}$ is
$$\Delta \tilde{d} = \max_{\mathcal{S},\mathcal{S'}:\ adj(\mathcal{S},\mathcal{S'})}{|\tilde{d}(\mathcal{S}) - \tilde{d}(\mathcal{S'})|} = \frac{1}{m}$$
so we add noise $\sim \text{Laplace}(0,\frac{1}{\eps m})$.

The \emph{overall} Algorithm \ref{DworkDE} satisfies $2\eps$-pan-privacy. Specifically, for all possible states (bitarrays) $\underline{\mathbf{b}}$ and outputs (estimated densities) $\underline{\tilde{d}}$, it holds that

\begin{eqnarray*}
\Prob( \ \mathbf{b(\mathcal{S})}=\underline{\mathbf{b}}\ ,\ \tilde{d}(\mathcal{S})=\underline{\tilde{d}} \ )
& \ = \ &
\Prob( \ \mathbf{b(\mathcal{S})}=\underline{\mathbf{b}}\ )\ \Prob(\ \tilde{d}(\mathcal{S})=\underline{\tilde{d}} \ | \ \mathbf{b(\mathcal{S})}=\underline{\mathbf{b}} \ ) \\
& \leq & e^{\eps}\ 
\Prob( \ \mathbf{b(\mathcal{S'})}=\underline{\mathbf{b}}\ )\\
&&e^{\eps}\ 
\Prob(\ \tilde{d}(\mathcal{S'})=\underline{\tilde{d}} \ | \ \mathbf{b(\mathcal{S'})}=\underline{\mathbf{b}} \ )\\
& = & e^{2\eps}\ \Prob( \ \mathbf{b(\mathcal{S'})}=\underline{\mathbf{b}}\ ,\ \tilde{d}(\mathcal{S'})=\underline{\tilde{d}} \ )
\end{eqnarray*}

so the definition of pan-privacy is satisfied.
\end{proof}
\end{tcolorbox}

\subsection{Accuracy Analysis}
As far as the accuracy of Algorithm \ref{DworkDE} is concerned, we have two theorems.
The first quantifies the bias and mean squared error of the estimator.
Although Dwork et al. \cite{dwork2010pan} demonstrate that their estimator is unbiased, they do not discuss its mean squared error.

Let $\mathcal{S}_M$ be the subsequence (sub-stream) of the original stream $\mathcal{S}$ that consists only of updates that refer to users in $M$. In particular, $\mathcal{S}_M$ is constructed as:
\begin{eqnarray*}
&&\text{for } i=1 \text{ to } T \\
&&\ \ \ \ \text{if } s_i \in M: \text{ add } s_i \text{ to } \mathcal{S}_M
\end{eqnarray*}

\begin{lemma}\label{lemma_total_var}
A special case of the law of total variance states that, if events $A$ and $B$ partition the whole outcome space of a random variable $b$, then:
\begin{eqnarray*}
\text{var}(b) &=& \text{var}(b|A) \Prob(A) + \text{var}(b|B) \Prob(B) \\
&& + \E[b|A]^2 (1-\Prob(A))\Prob(A) + \E[b|B]^2 (1-\Prob(B))\Prob(B) \\
&& - 2\E[b|A] \Prob(A) \E[b|B] \Prob(B)
\end{eqnarray*}
\end{lemma}

\begin{thm}\label{thm_DworkDE_acc1}
For fixed sample $M$, Algorithm \ref{DworkDE} provides an unbiased estimate $\tilde{d}$ of the density of $\mathcal{S}_M$ and has mean squared error: 
$$\E [ \ (\tilde{d}-d(\mathcal{S}_M))^2 \ ] \ \leq \ \frac{2(2m+1)}{m^2\eps^2}$$
\end{thm}

\begin{tcolorbox}[breakable]
\begin{proof}
We begin with the \emph{bias} computation. We examine the distribution of an arbitrary entry in $\textbf{b}$:
\[ 
b_i \sim \left\{
\begin{array}{ll}
      \text{Bernoulli}(\frac{1}{2}) & M[i] \not\in \mathcal{S}_M\\
      \text{Bernoulli}(\frac{1}{2}+\frac{\eps}{4}) & M[i] \in \mathcal{S}_M \\
\end{array} 
\right. 
\]
Note that the distribution of $b_i$ does not depend on the number of appearances of $M[i]$; it only depends on whether it appeared or not.

Now, let $\hat{d}=\frac{1}{m}\sum_{i=1}^{m}{b_i}$. Then,
\begin{eqnarray*}
\E[\hat{d}] &=&  \frac{1}{m}\sum_{i=1}^{m}{\E[b_i]} \\
&=& \frac{1}{m}\sum_{i=1}^{m} \E[b_i|M[i] \in \mathcal{S}_M] \Prob(M[i] \in \mathcal{S}_M) \\
& & \ \ \ \ \ \ \ \ + \E[b_i|M[i] \not\in \mathcal{S}_M] \Prob(M[i] \not\in \mathcal{S}_M) \\
&=& \frac{1}{m}\sum_{i=1}^{m}{ (\frac{1}{2}+\frac{\eps}{4})d(\mathcal{S}_M) \ + \ \frac{1}{2}(1-d(\mathcal{S}_M)) } \\
&=& \frac{1}{2} + \frac{\eps}{4}d(\mathcal{S}_M)
\end{eqnarray*}
where we interpret the probability that a user is present in the sub-stream as the density of sub-stream. This is true if all users are considered equally likely to appear.

The final estimate (output) $\tilde{d}$ is then computed as  $\tilde{d} = \frac{4}{\eps}(\hat{d}-\frac{1}{2}) + \text{Laplace}(0,\frac{1}{\eps m})$, which gives that
$$ \E[\tilde{d}] \ = \ \frac{4}{\eps}(\E[\hat{d}]-\frac{1}{2}) + \E[\text{Laplace}(0,\frac{1}{\eps m})] \ = \ d(\mathcal{S}_M)$$
so it is indeed an unbiased estimate.\\

We proceed with the \emph{mean squared error}. Given that $\tilde{d}$ is an unbiased estimate of $d(\mathcal{S}_M)$, its mean squared error coincides with its variance.

We make use of Lemma \ref{lemma_total_var} and of the fact that, if $b \sim \text{Bernoulli}(p)$, then $\text{var}(b)=p(1-p)$.
In our setting, we have $m$ Bernoulli random variables $b_i \ (i=1,...,m)$, and for each user $i$, $A$ is the event that $M[i] \in \mathcal{S}_M$ and $B$ is the event that $M[i] \not\in \mathcal{S}_M$. Therefore, $b_i|A \sim \text{Bernoulli}(\frac{1}{2}+\frac{\eps}{4})$ and $b_i|B \sim \text{Bernoulli}(\frac{1}{2})$. By a straightforward computation:

\begin{eqnarray*}
&\text{var}(b_i)& = \ \frac{1}{4} - \frac{\eps^2 d(\mathcal{S}_M)^2}{16} \\
\Rightarrow &\text{var}(\hat{d})& = \ \frac{1}{m^2}\sum_{i=1}^{m}{ \text{var}(b_i)} = \frac{1}{4m} - \frac{\eps^2 d(\mathcal{S}_M)^2}{16m} \\
\Rightarrow &\text{var}(\tilde{d})& = \ (\frac{4}{\eps})^2 \text{var}(\hat{d}) + \text{var}(\text{Laplace}(0,\frac{1}{\eps m})) \\
&& = \ \frac{4}{m \eps^2} - \frac{d(\mathcal{S}_M)^2}{m} + \frac{2}{m^2 \eps^2}\\
&& \leq \frac{2(2m+1)}{m^2\eps^2}
\end{eqnarray*}
which completes the proof. To derive the last inequality we used the fact that $0\leq d(\mathcal{S}_M) \leq 1$.
\end{proof}
\end{tcolorbox}

The next theorem validates that the estimator provides the desired $(\alpha,\beta)$-approximation of the actual stream density. In contrast to Dwork et al. \cite{dwork2010pan}, we parameterize the proof, so we are then able to numerically compute the tightest version of the bound we derive.

Before presenting the theorem, we prove the following useful lemma.

\begin{lemma}\label{lemma_absolute_sum}
For any random variables $X$ and $Y$, and for some $\alpha>0$ and $\delta \in (0,1)$:
\begin{eqnarray*}
\Prob( |X+Y|>\alpha ) \ \leq \ \Prob ( |X|>\alpha \delta ) + \Prob( |Y| >\alpha (1-\delta) )
\end{eqnarray*}
\end{lemma}

\begin{tcolorbox}[breakable]
\begin{proof}
We have that:
\begin{eqnarray*}
\{ (X,Y): |X+Y|>\alpha \} &=& \{ (X,Y): X+Y>\alpha \text{ or } X+Y<-\alpha \} \\
&=& \{ (X,Y): X+Y>\alpha \} \\
&& \cup \ \{ (X,Y): X+Y<-\alpha \} \\
&\subseteq & \{ (X,Y): X>\alpha \delta \text{ or }  Y >\alpha (1-\delta) \} \\
&& \cup \ \{ (X,Y): X<-\alpha \delta \text{ or }  Y <-\alpha (1-\delta) \} \\
&=& \{ (X,Y): X>\alpha \delta \text{ or }  Y >\alpha (1-\delta) \\
&& \ \ \ \ \ \text{ or } X<-\alpha \delta \text{ or }  Y <-\alpha (1-\delta) \} \\
&=& \{ (X,Y): |X|>\alpha \delta \text{ or } |Y| >\alpha (1-\delta) \}
\end{eqnarray*}
so it follows that:
\begin{eqnarray*}
\Prob( |X+Y|>\alpha ) & \leq & \Prob ( |X|>\alpha \delta \text{ or } |Y| >\alpha (1-\delta) ) \\
& \leq &  \Prob ( |X|>\alpha \delta ) + \Prob( |Y| >\alpha (1-\delta) )
\end{eqnarray*}
where the last inequality follows by the union bound.
\end{proof}
\end{tcolorbox}

\begin{thm}\label{thm_DworkDE_acc2}
If the sample maintained by Algorithm \ref{DworkDE} consists of $m = \mathcal{O}(\frac{1}{\eps^2 \alpha^2}\log{\frac{1}{\beta}})$ users from $\mathcal{U}$, then, for fixed input $\mathcal{S}$: 
$$\Prob(\ | \tilde{d}-d(\mathcal{S}) | \ \geq \ \alpha \ )\ \leq \ \beta$$
where the probability space is over the random choices of the algorithm.
\end{thm}

\begin{tcolorbox}[breakable]
\begin{proof}
Let $\hat{d}=\frac{1}{m}\sum_{i=1}^{m}{b_i}$. We apply Lemma \ref{lemma_absolute_sum} twice, so for some $\alpha>0$ and $\delta_1, \delta_2 \in (0,1)$:
\begin{eqnarray*}
\Prob(\ | \tilde{d}-d(\mathcal{S}) | \ \geq \ \alpha \ ) &=& \Prob(\ | \tilde{d}-d(\mathcal{S}_M)+d(\mathcal{S}_M)-d(\mathcal{S}) | \ \geq \ \alpha \ ) \\
& \leq &  \Prob(\ | \tilde{d}-d(\mathcal{S}_M) | \ \geq \ \delta_1 \alpha \ ) \\ 
&& + \Prob(\ |d(\mathcal{S}_M)-d(\mathcal{S}) | \ \geq \ (1-\delta_1)\alpha \ ) \\
&=& \Prob(\ | \frac{4}{\eps}\hat{d}-\frac{2}{\eps}+\text{Laplace}(0,\frac{1}{\eps m})-d(\mathcal{S}_M) | \ \geq \ \delta_1 \alpha \ ) \\ 
&& + \ \Prob(\ |d(\mathcal{S}_M)-d(\mathcal{S}) | \ \geq \ (1-\delta_1)\alpha \ ) \\
&=& \Prob(\ | \hat{d}-\frac{1}{2}+\frac{\eps}{4}\text{Laplace}(0,\frac{1}{\eps m})-\frac{\eps}{4} d(\mathcal{S}_M) | \\
&& \ \ \ \ \geq \ \frac{\eps}{4}\delta_1 \alpha \ ) \\ 
&& + \ \Prob(\ |d(\mathcal{S}_M)-d(\mathcal{S}) | \ \geq \ (1-\delta_1)\alpha \ ) \\
&=& \Prob(\ | \hat{d}-\E [\hat{d}] +\frac{\eps}{4}\text{Laplace}(0,\frac{1}{\eps m}) | \ \geq \ \frac{\eps}{4}\delta_1 \alpha \ ) \\ 
&& + \ \Prob(\ |d(\mathcal{S}_M)-d(\mathcal{S}) | \ \geq \ (1-\delta_1)\alpha \ ) \\
&\leq& \Prob(\ |\frac{\eps}{4}\text{Laplace}(0,\frac{1}{\eps m}) | \ \geq \ \frac{\eps}{4}\delta_1 \delta_2 \alpha \ ) \\
&& +\ \Prob(\ | \hat{d}-\E [\hat{d}] | \ \geq \ \frac{\eps}{4}\delta_1 (1-\delta_2) \alpha\ ) \\ 
&& + \ \Prob(\ |d(\mathcal{S}_M)-d(\mathcal{S}) | \ \geq \ (1-\delta_1)\alpha \ )
\end{eqnarray*}

We examine the quantities involved in the final expression.
For fixed input, $d(\mathcal{S})$ is deterministic, as we do not take into account the randomness of the input.
On the contrary, $d(\mathcal{S}_M)$ is random (due to sampling), $\hat{d}$ is random (as a sum of Bernoulli random variables), $\E[\hat{d}]$ is random (as a function of $d(\mathcal{S}_M)$), and $\tilde{d}$ is random (as the sum of $\hat{d}$ and a Laplace random variable).
We therefore have 3 sources of error to control, which correspond to the following probabilities:
\begin{itemize}
\item[-] $p_1 = \Prob(\ |d(\mathcal{S}_M)-d(\mathcal{S}) | \ \geq \ (1-\delta_1)\alpha \ )$
\item[-] $p_2 = \Prob(\ | \hat{d}-\E [\hat{d}] | \ \geq \ \frac{\eps}{4}\delta_1 (1-\delta_2) \alpha\ )$
\item[-] $p_3 = \Prob(\ |\frac{\eps}{4}\text{Laplace}(0,\frac{1}{\eps m}) | \ \geq \ \frac{\eps}{4}\delta_1 \delta_2 \alpha \ )$
\end{itemize}

We want $p_1+p_2+p_3 \leq \beta$. We bound each error probability separately, so for some $\delta_3, \delta_4 \in (0,1), \text{ such that } \delta_3 + \delta_4 < 1 $, we want:
$$p_1<\delta_3 \beta \ , \ p_2<\delta_4 \beta \ , \ p_3<(1 - \delta_3 - \delta_4) \beta$$

$\bullet$ \underline{First, we bound $p_1$.}\\
We define, $\forall i \in \{1,...,m\}$,
\[ 
X_i = \left\{
\begin{array}{ll}
      0 & M[i] \not\in \mathcal{S}_M\\
      1 & M[i] \in \mathcal{S}_M \\
\end{array} 
\right. 
\]
and since $\mathcal{S}_M$ was sampled uniformly at random from $\mathcal{S}$, $X_i \sim \text{Bernoulli}(d(\mathcal{S}))$. It is easy to see that, since $d(\mathcal{S}_M) = \frac{1}{m}\sum_{i=1}^m{X_i}$, $\E[d(\mathcal{S}_M)]=d(\mathcal{S})$.

Since $d(\mathcal{S}_M)$ is a weighted sum of i.i.d. Bernoulli random variables, we take an additive, two-sided Chernoff bound:
$$ p_1 = \Prob(\ |\frac{1}{m}\sum_{i=1}^m{X_i}-\frac{1}{m}\sum_{i=1}^m{\E[X_i]} | \ \geq \ (1-\delta_1)\alpha \ ) \ \leq \ 2 e^{-2m\alpha^2(1-\delta_1)^2}$$
and, therefore:
$$ p_1 \leq \delta_3 \beta \ \Leftrightarrow \ m \geq \underbrace{\frac{1}{2 \alpha^2 (1-\delta_1)^2} \ln(\frac{2}{\beta \delta_3})}_{m_1} \ = \ \mathcal{O}(\frac{1}{\alpha^2}\ln(\frac{1}{\beta}))$$\\

$\bullet$ \underline{Next, we bound $p_2$.}\\
As we already mentioned, both $\hat{d}$ and $\E[\hat{d}]=\frac{1}{2} + \frac{\eps}{4}d(\mathcal{S}_M)$ are random. $d(\mathcal{S}_M)$ takes values in the set $\mathcal{D} = \{0,\frac{1}{m},...,\frac{m-1}{m},1\}$, so using the law of total probability:
$$ p_2 = \sum_{\underline{d} \in \mathcal{D}}{ \Prob(\ | \hat{d}-\E[\hat{d}] | \ \geq \ \frac{\eps}{4}\delta_1 (1-\delta_2) \alpha\ \ | \ d(\mathcal{S}_M) = \underline{d} \ ) \ \Prob(\ d(\mathcal{S}_M) = \underline{d}\ ) }$$

We make the following two critical remarks:
\begin{itemize}
\item[-] For fixed $d(\mathcal{S}_M) = \underline{d}$, $\E[\hat{d}]$ is no longer random.

\item[-] Let $d_i, \ i\in \{1,...,m\}$ denote the distribution from which each $b_i$ is drawn from, so that $d_i=0 \Rightarrow b_i \sim \text{Bernoulli}(\frac{1}{2})$ and $d_i = 1 \Rightarrow b_i \sim \text{Bernoulli}(\frac{1}{2}+\frac{\eps}{4})$. Once we fix $d(\mathcal{S}_M) = \underline{d}$, the $d_i$'s are not independent, as $\sum_{i=1}^m{d_i}=m\underline{d}$.

However, the $b_i$'s are independent, as we impose no constraint on them and each is drawn independently from a fixed distribution. Therefore, $\hat{d}$ is a sum of independent Poisson trials.
\end{itemize}

We again make use of an additive, two-sided Chernoff bound:

$$\Prob(\ | \hat{d}-\E[\hat{d}] | \ \geq \ \frac{\eps}{4}\delta_1 (1-\delta_2) \alpha\ \ | \ d(\mathcal{S}_M) = \underline{d} \ ) \ \leq \ 2 e^{-2m \eps^2 \alpha^2 \delta_1^2 (1-\delta_2)^2 \frac{1}{16}}$$

We observe that the bound is independent of $\underline{d}$, so:

$$ p_2 \leq \sum_{\underline{d} \in \mathcal{D}}{ 2 e^{-2m \eps^2 \alpha^2 \delta_1^2 (1-\delta_2)^2 \frac{1}{16}} \ \Prob(\ d(\mathcal{S}_M) = \underline{d}\ ) } = 2 e^{- \frac{1}{8} m \eps^2 \alpha^2 \delta_1^2 (1-\delta_2)^2}$$

and, therefore:
$$ p_2 \leq \delta_4 \beta \ \Leftrightarrow \ m \geq \underbrace{\frac{8}{\eps^2 \alpha^2 \delta_1^2 (1-\delta_2)^2} \ln(\frac{2}{\beta \delta_4})}_{m_2} \ = \ \mathcal{O}(\frac{1}{\eps^2 \alpha^2}\ln(\frac{1}{\beta}))$$\\

$\bullet$ \underline{Finally, we compute $p_3$.}
\begin{eqnarray*}
p_3 &=& \Prob(\ |\frac{\eps}{4}\text{Laplace}(0,\frac{1}{\eps m}) | \ \geq \ \frac{\eps}{4}\delta_1 \delta_2 \alpha \ ) \\
&=& \Prob(\ |\text{Laplace}(0,\frac{1}{\eps m}) | \ \geq \ \delta_1 \delta_2 \alpha \ ) \\
&=& \Prob(\ \text{Exponential}(\eps m) \ \geq \ \delta_1 \delta_2 \alpha \ ) \\
&=& e^{-\eps \alpha m \delta_1 \delta_2}
\end{eqnarray*}

and, therefore:
$$ p_3 \leq (1-\delta_3-\delta_4) \beta \ \Leftrightarrow \ m \geq \underbrace{\frac{1}{\eps \alpha \delta_1 \delta_2} \ln(\frac{1}{\beta (1-\delta_3-\delta_4)})}_{m_3} \ = \ \mathcal{O}(\frac{1}{\eps \alpha}\ln(\frac{1}{\beta}))$$\\

By picking $m \geq \max\{m_1,m_2,m_3\}=\mathcal{O}(\frac{1}{\eps^2 \alpha^2}\ln(\frac{1}{\beta}))$, we ensure that:
$$\Prob(\ | \tilde{d}-d(\mathcal{S}) | \ \geq \ \alpha \ )\ \leq \ p_1+p_2+p_3\ \leq \ \delta_3 \beta+\delta_4 \beta+(1-\delta_3-\delta_4) \beta \ = \ \beta$$
which completes the proof.
\end{proof}
\end{tcolorbox}

Theorem \ref{thm_DworkDE_acc2} offers an additive error guarantee, which may not be so useful if the density of the input stream is small.
Dwork et al. \cite{dwork2010pan} show how to modify their algorithm to obtain a multiplicative error guarantee, which is more meaningful in such cases.
We do not examine this point in our work.

As we already stated, our parameterized proof allows us to optimally tune the parameters $\delta_1,\delta_2,\delta_3,\delta_4$ and, as a result, to compute the tightest version of the bound we derive.
In particular, for fixed $\eps, \alpha, m$, the tightest bound $\beta$ on $\Prob(\ | \tilde{d}-d(\mathcal{S}) | \ \geq \ \alpha \ )$ is computed by numerically solving the following optimization problem:

\begin{equation*}
\begin{aligned}
& \underset{\delta_1,\delta_2}{\text{minimize}} 
& &\beta(\delta_1,\delta_2) & = & \ 2 e^{-2m\alpha^2(1-\delta_1)^2} \\
& & &  & & \ + 2 e^{- \frac{1}{8} m \eps^2 \alpha^2 \delta_1^2 (1-\delta_2)^2} + e^{-\eps \alpha m \delta_1 \delta_2}\\
& \text{subject to}
& & 0\leq\delta_1\leq1 \\
& & & 0\leq\delta_2\leq1 
\end{aligned}
\end{equation*}

\subsection{Picking the Optimal Sample Size}\label{subSEC_sample_size}
Theorem \ref{thm_DworkDE_acc2} provides an asymptotic expression for the sample size $m$ to achieve the desired approximation accuracy.
A question that arises is how we should pick $m$ in practice.
This is again achieved by taking advantage of our parameterized proof and numerically solving a similar optimization problem,
which allows us to compute the optimal (minimum) sample size $m^*$ that achieves the desired approximation accuracy, according to the bounds we derived.
Specifically, for fixed $\eps, \alpha, \beta$, we define:

\begin{equation*}
\begin{aligned}
m(\delta_1,\delta_2,\delta_3,\delta_4) &=& \max \{ \ & m_1 , m_2 , m_3 \ \} \\
&=& \max \{ \ &\frac{1}{2 \alpha^2 (1-\delta_1)^2} \ln(\frac{2}{\beta \delta_3}) \ ,\\
& & & \frac{8}{\eps^2 \alpha^2 \delta_1^2 (1-\delta_2)^2} \ln(\frac{2}{\beta \delta_4}) \ ,\\
& & & \frac{1}{\eps \alpha \delta_1 \delta_2} \ln(\frac{1}{\beta (1-\delta_3-\delta_4)}) \  \}
\end{aligned}
\end{equation*}

Then, we pick $m^*$ as the solution to the following optimization problem:

\begin{equation*}
\begin{aligned}
& \underset{\delta_1,\delta_2,\delta_3,\delta_4}{\text{minimize}} 
& &m(\delta_1,\delta_2,\delta_3,\delta_4) \\
& \text{subject to}
& & 0\leq\delta_i\leq1 \ , \ i \in \{1,2,3,4\}\\
& & & \delta_3+\delta_4\leq1 
\end{aligned}
\end{equation*}

\begin{figure}%[H]%[ht!]
    \centering
    \includegraphics[width=\columnwidth]{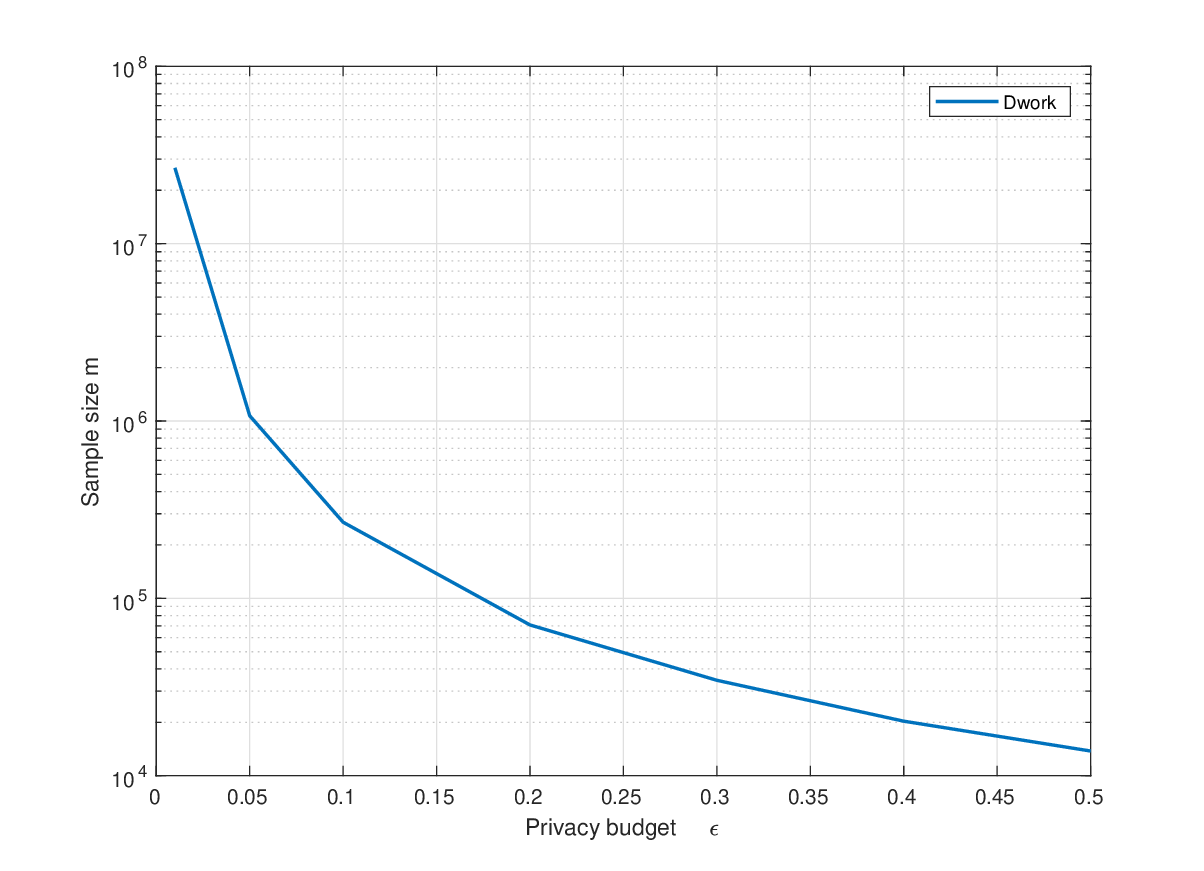}
    \caption{Proposed sample size}
	\label{FIG_sample_size_bound_dwork}
\end{figure}

Figure \ref{FIG_sample_size_bound_dwork} illustrates the proposed sample sizes for varying privacy budget.
We repeat that our approach allows us to compute the tightest version of the specific bound that we derive on the probability of error and hence on the sample size.
Specifically, the sample sizes we propose are smaller by more than an order of magnitude, compared to the sample sizes proposed by Roth in his lectures, who arbitrarily proposes to pick $m=\frac{128}{\eps^2 \alpha^2}\ln(\frac{1}{\beta})$.
Nevertheless, we remark that the bound itself is not tight;
this is an experimental observation and indicates that the same accuracy can be achieved with even smaller samples.

\section{Optimal Bernoulli Density Estimator}

In this section, we modify Algorithm \ref{DworkDE} and derive a novel algorithm that significantly outperforms the original algorithm (both theoretically and experimentally).
The key reason behind our algorithm's superiority is that -in contrast to Algorithm \ref{DworkDE}- it manages to use all the allocated privacy budget.

\subsection{On the Use of the Allocated Privacy Budget}
Recall that, in order to ensure that its state satisfies differential privacy, Algorithm \ref{DworkDE} utilizes two different distributions, one for users that do not appear in the stream, and one for users that do appear.
We now introduce a little extra notation; the bit that corresponds to a user from the former category is drawn from the distribution with pmf $f_{init} = \text{Bernoulli}(\frac{1}{2})$, while the bit that corresponds to a user from the latter category is drawn from $f_{upd} = \text{Bernoulli}(\frac{1}{2}+\frac{\eps}{4})$.
Although by $f_{init}$ and $f_{upd}$ we formally denote the probability mass functions of the two Bernoulli distributions, at some points we use the same notation to refer to the distributions themselves (abusing notation a little bit).
To satisfy differential privacy, we have to ensure that, $\forall b \in \{0,1\}$:
$$e^{-\eps} \leq R(b)=\frac{f_{upd}(b)}{f_{init}(b)} \leq e^{\eps}$$ 

\begin{figure}%[H]%[ht!]
    \centering
    \includegraphics[width=0.75\columnwidth]{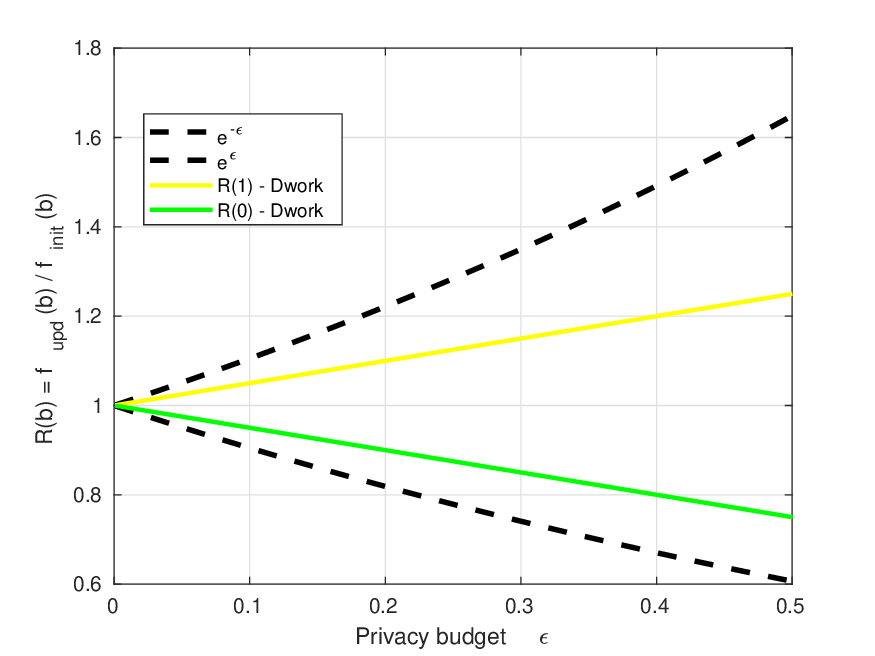}
    \caption{Dwork - State differential privacy}
	\label{FIG_dwork_privacy_budget}
\end{figure}

Figure \ref{FIG_dwork_privacy_budget} illustrates that Algorithm \ref{DworkDE} does ensure that its state satisfies differential privacy (as we have already proved in Theorem \ref{thm_DworkDE_priv}).

At the same time however, it is not hard to observe that Algorithm \ref{DworkDE} fails to use all the allocated privacy budget. We have already shown (in the proof of Theorem \ref{thm_DworkDE_priv}) that $R(0)=1-\frac{\eps}{2}$ and $R(1)=1+\frac{\eps}{2}$. Let $\eps'>0$ be the actual privacy budget that Algorithm \ref{DworkDE} consumes. Then:

\[ 
\left.
\begin{array}{ll}
     e^{-\eps'} \leq 1-\frac{\eps}{2} \leq e^{\eps'}\\
     e^{-\eps'} \leq 1+\frac{\eps}{2} \leq e^{\eps'} \\
\end{array} 
\right\}
\Rightarrow \eps \ ' \geq \max \{\ \ln(1+\frac{\eps}{2}) \ , \ -\ln(1-\frac{\eps}{2}) \ \}
\]

\begin{figure}%[H]%[ht!]
    \centering
    \includegraphics[width=0.75\columnwidth]{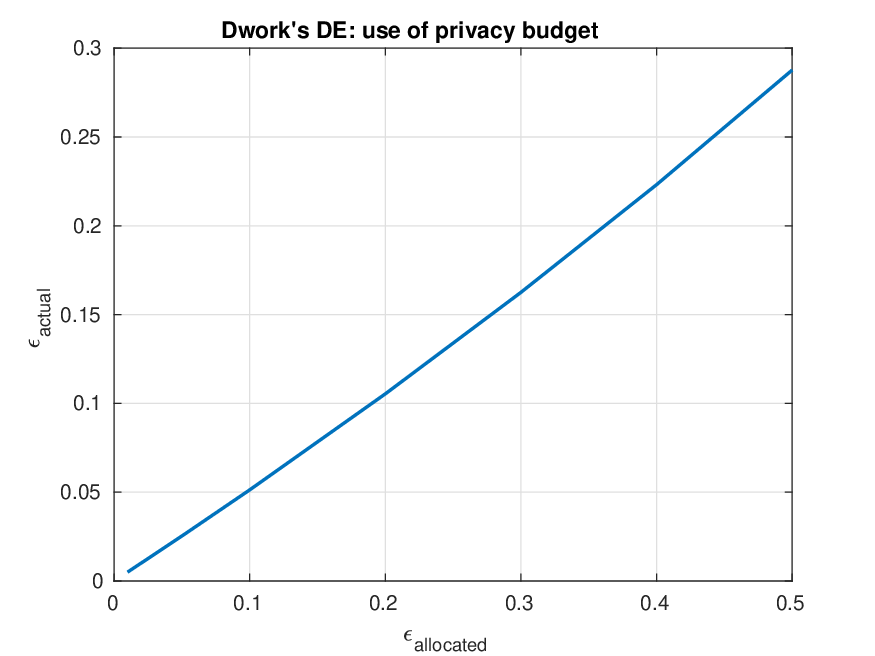}
    \caption{Dwork - Allocated vs actual privacy budget}
	\label{FIG_actual_privacy_budget}
\end{figure}

Figure \ref{FIG_actual_privacy_budget} illustrates the actual privacy budget used by Algorithm \ref{DworkDE} for each allocated privacy budget.

\subsection{Optimally Tuning the Bernoulli Distributions}
Based on the aforementioned observation, we optimally tune the Bernoulli distributions used (with pmf's $f_{init}$ and $f_{upd}$), by picking a pair of parameters that tightly satisfies $\eps$-differential privacy, for any given $\eps$.
In particular, we maximize the distributions' distance (or, more precisely, the difference of their parameters), which allows us to more accurately distinguish between users that did not appear and users that appeared in the stream, during the density estimation computation (without sacrificing the users' privacy!).

We propose that the parameters of $f_{init}$ and $f_{upd}$ are picked symmetric around some value $c$; although the choice of $c$ (among the permitted values) does not seem to affect the algorithm, we explicitly show later what values $c$ is permitted to take.
For example, we could set $c=\frac{1}{2}$, or $c=\frac{1}{2}+\frac{\eps}{8}$, as in Algorithm \ref{DworkDE}.
In doing so, we end up with the following distributions: 
$$f_{init} = \text{Bernoulli}(c-x)$$
$$f_{upd} = \text{Bernoulli}(c+x)$$
for some $x > 0$ that corresponds to half the difference of the distributions' parameters. 
Clearly, we want $0<c-x<c+x<1$.
For example, in Algorithm \ref{DworkDE}, $x=\frac{\eps}{8}$, and $0 < c-x=\frac{1}{2} < c+x=\frac{1}{2}+\frac{\eps}{4} < 1 \ , \ \forall \eps \in (0,\frac{1}{2}]$. 

The modification we propose is in the selection of $x$. Again, to satisfy differential privacy, we have to ensure that, $\forall b \in \{0,1\}$:
\begin{eqnarray*}
e^{-\eps} \leq \frac{f_{upd}(b)}{f_{init}(b)} \leq e^{\eps}
& \Leftrightarrow &
\left\{
\begin{array}{ll}
     e^{-\eps} \leq \frac{c+x}{c-x} \leq e^{\eps} \\
     e^{-\eps} \leq \frac{c-x}{c+x} \leq e^{\eps} \\
\end{array} 
\right. \\
& \Rightarrow &
x \ \leq \ \frac{e^{\eps}-1}{e^{\eps}+1} \ c
\ = \ \tanh(\frac{\eps}{2})c
\end{eqnarray*}
Since our goal was to maximize the difference of the distributions' parameters, we pick $x=\tanh(\frac{\eps}{2})c$.
The proposed distributions are:
\begin{eqnarray*}
f_{init} = \text{Bernoulli}(\ c \ (1-\tanh(\frac{\eps}{2})\ ) \\
f_{upd} = \text{Bernoulli}(\ c \ (1+\tanh(\frac{\eps}{2})\ )
\end{eqnarray*}
and, since both parameters have to be greater than zero and less than one, we can determine the values that $c$ can take for each fixed $\eps$. For $0 < \eps \leq \frac{1}{2}$, it is easy to see that $0 < \tanh(\frac{\eps}{2}) < \frac{1}{4}$ (since $\tanh(\frac{\eps}{2})$ is a monotonically increasing function of $\eps$), so we can pick any $c \in (0,\frac{4}{5}]$, regardless of the specific value of $\eps \in (0,\frac{1}{2}]$.

\begin{figure}%[H]%[ht!]
    \centering
    \includegraphics[width=0.75\columnwidth]{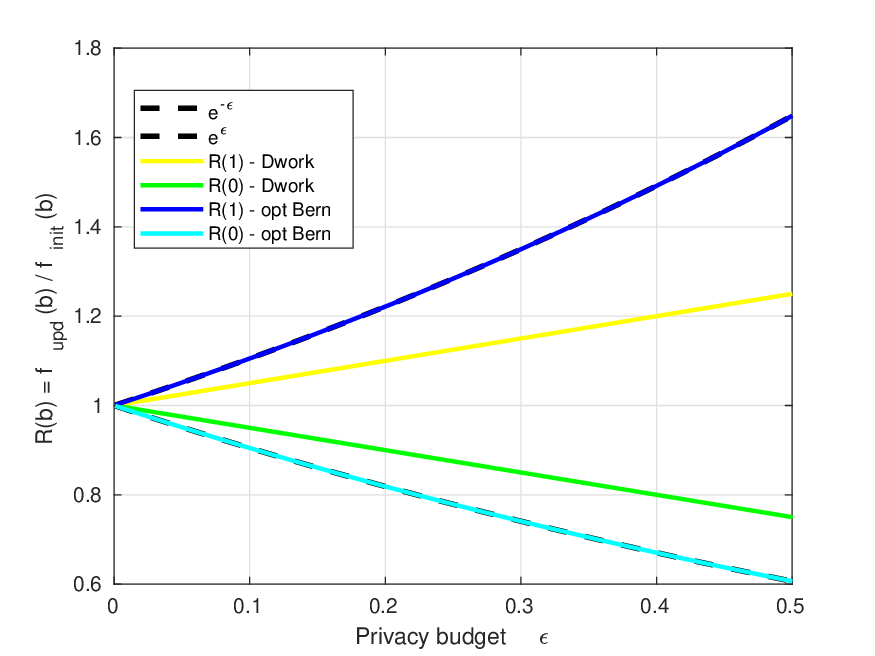}
    \caption{Optimal Bernoulli - State differential privacy}
	\label{FIG_opt_bern_privacy_budget}
\end{figure}

Figure \ref{FIG_opt_bern_privacy_budget} illustrates that the proposed Bernoulli distributions indeed use all the allocated privacy budget.

\subsection{Estimator \& Analysis}
We now present the modified algorithm, which we call the Optimal Bernoulli Density Estimator. For simplicity, we set $c=\frac{1}{2}$.

\begin{algorithm}%[H]

\caption{Optimal Bernoulli Density Estimator\label{OptBernDE}}

\DontPrintSemicolon

\KwIn{Data stream $\mathcal{S}$, Privacy budget $\eps$, Accuracy parameters $(\alpha,\beta)$}

\KwOut{Density $\tilde{d}(\mathcal{S})$}

Compute $m^*$ and set $m=m^*$\;

Sample a random subset $M \subseteq \mathcal{U}$ of $m$ users (without replacement) and define an arbitrary ordering over $M$\;

Create a bitarray $\mathbf{b}=[b_1\ ...\ b_{m}]$ and map $M[i] \rightarrow b_i,\ \forall \ i \in \{1,...,m\}$\;

Initialize $\mathbf{b}$ randomly: $b_i \sim \text{Bernoulli}(\frac{1}{2}(1-\tanh(\frac{\eps}{2}))), \ \forall \ i \in \{1,...,m\}$\;

\For{t = 1 \KwTo T}{

	\If{$s_t \in M$}{
		
		Find $i:\ M[i]=s_t$\;
		
		Re-sample: $b_i \sim \text{Bernoulli}(\frac{1}{2}(1+\tanh(\frac{\eps}{2})))$\;
			
	}
}

Return $\tilde{d}(\mathcal{S})\ = \ \frac{1}{\tanh(\frac{\eps}{2})}(\frac{1}{m}\sum_{i=1}^{m}{b_i}-\frac{1}{2} + \frac{1}{2} \tanh(\frac{\eps}{2}) ) \ + \ \text{Laplace}(0,\frac{1}{\eps m})$\;

\end{algorithm}

We proceed with the privacy and accuracy analysis of Algorithm \ref{OptBernDE}, following the lines of our analysis of Algorithm \ref{DworkDE}.

\begin{thm}\label{thm_OptBernDE_priv}
Assume $\eps \leq \frac{1}{2}$. Then Algorithm \ref{OptBernDE} satisfies $2\eps$-pan-privacy and utilizes all the allocated privacy budget.
\end{thm}

\begin{tcolorbox}[breakable]
\begin{proof}
The proof is identical with that of Theorem \ref{thm_DworkDE_priv}.
The only modification is on proving that the state (bitarray) satisfies $\eps$-differential privacy, and in particular, for a user $u \in M$ that appears in stream $\mathcal{S}$ and does not appear in stream $\mathcal{S}'$ (again, let $b_1$ be the entry that corresponds to $u$ in the bitarray), we have: 
\begin{eqnarray*}
\frac{\Prob( \ \mathbf{b(\mathcal{S})}=[1 \ \underline{b_2} \ ...\ \underline{b_{m}}]\ )}{\Prob( \ \mathbf{b(\mathcal{S'})}=[1 \ \underline{b_2} \ ...\ \underline{b_{m}}]\ )} &=& \frac{\Prob( \ b_1(\mathcal{S})=1\ )}{\Prob( \ b_1(\mathcal{S'})=1\ )} \\
&=& \frac{ \frac{1}{2}(1+\frac{e^{\eps}-1}{e^{\eps}+1}) }{ \frac{1}{2}(1-\frac{e^{\eps}-1}{e^{\eps}+1}) } \ = \ e^{\eps} \\
\frac{\Prob( \ \mathbf{b(\mathcal{S})}=[0 \ \underline{b_2} \ ...\ \underline{b_{m}}]\ )}{\Prob( \ \mathbf{b(\mathcal{S'})}=[0 \ \underline{b_2} \ ...\ \underline{b_{m}}]\ )} &=& \frac{\Prob( \ b_1(\mathcal{S})=0\ )}{\Prob( \ b_1(\mathcal{S'})=0\ )} \\
&=& \frac{ \frac{1}{2}(1-\frac{e^{\eps}-1}{e^{\eps}+1}) }{ \frac{1}{2}(1+\frac{e^{\eps}-1}{e^{\eps}+1}) } \ = \ e^{-\eps}
\end{eqnarray*}
so user $u$ is guaranteed $\eps$-differential privacy against an adversary that observes $u$'s entry in $\mathbf{b}$.
\end{proof}
\end{tcolorbox}

\begin{thm}\label{thm_OptBernDE_acc1}
For fixed sample $M$, Algorithm \ref{OptBernDE} provides an unbiased estimate $\tilde{d}$ of the density of $\mathcal{S}_M$ and has mean squared error: 
$$\E [ \ (\tilde{d}-d(\mathcal{S}_M))^2 \ ] \ \leq \ \frac{1}{4m \tanh^2(\frac{\eps}{2})} + \frac{2}{m^2 \eps^2}$$
\end{thm}

\begin{tcolorbox}[breakable]
\begin{proof}
The proof is similar with that of Theorem \ref{thm_DworkDE_acc1}.\\

We begin with the \emph{bias} computation. In order to examine the distribution of an arbitrary entry in $\textbf{b}$, we introduce the notation $p_{init}=\frac{1}{2}(1-\tanh(\frac{\eps}{2}))$ and $p_{upd}=\frac{1}{2}(1+\tanh(\frac{\eps}{2}))$. Then,
\[ 
b_i \sim \left\{
\begin{array}{ll}
      \text{Bernoulli}(p_{init}) & M[i] \not\in \mathcal{S}_M\\
      \text{Bernoulli}(p_{upd}) & M[i] \in \mathcal{S}_M \\
\end{array} 
\right. 
\]

Denoting $\hat{d}=\frac{1}{m}\sum_{i=1}^{m}{b_i}$, and applying a similar computation as in the proof of Theorem \ref{thm_DworkDE_acc1}, gives: 
\begin{eqnarray*}
\E[\hat{d}] &=& p_{init} + (p_{upd}-p_{init})d(\mathcal{S}_M) \\ 
&=& \frac{1}{2}(1-\tanh(\frac{\eps}{2})) + \tanh(\frac{\eps}{2}) d(\mathcal{S}_M)
\end{eqnarray*}

The final estimate (output) $\tilde{d}$ is:  
\begin{eqnarray*}
\tilde{d} &=& \frac{\hat{d}-p_{init}}{p_{upd}-p_{init}} + \text{Laplace}(0,\frac{1}{\eps m}) \\
&=& \frac{1}{\tanh(\frac{\eps}{2})}(\hat{d}-\frac{1}{2} + \frac{1}{2} \tanh(\frac{\eps}{2}) ) \ + \ \text{Laplace}(0,\frac{1}{\eps m})
\end{eqnarray*}
so it is an unbiased estimate.\\

We proceed with the \emph{mean squared error}. Again, since $\tilde{d}$ is an unbiased estimate of $d(\mathcal{S}_M)$, its mean squared error coincides with its variance.
Also, recall that we have $m$ Bernoulli random variables $b_i \ (i=1,...,m)$, so applying the law of total variance gives us:
\begin{equation*}
\begin{aligned}
& \text{var}(b_i) & = & \ p_{init}(1-p_{init})
\ + \ d(\mathcal{S}_M)[-p_{init}(1-p_{init})+p_{upd}(1-p_{upd})] \\
& & & + \ d(\mathcal{S}_M)(1-d(\mathcal{S}_M))(p_{upd}-p_{init})^2 \\
& & = & \ \frac{1}{4}[1-\tanh^2(\frac{\eps}{2})] \ + \ d(\mathcal{S}_M)(1-d(\mathcal{S}_M)) \tanh^2(\frac{\eps}{2}) \\
\Rightarrow \ & \text{var}(\hat{d}) & = & \ \frac{1}{m^2}\sum_{i=1}^{m}{ \text{var}(b_i)} \\
& & = & \ \frac{1}{4m}[1-\tanh^2(\frac{\eps}{2})] \ + \ \frac{d(\mathcal{S}_M)(1-d(\mathcal{S}_M))}{m} \tanh^2(\frac{\eps}{2}) \\
\Rightarrow \ & \text{var}(\tilde{d}) & = & \ (\frac{1}{p_{upd}-p_{init}})^2 \text{var}(\hat{d}) + \text{var}(\text{Laplace}(0,\frac{1}{\eps m})) \\
& & = & \ \frac{1}{4m}[\frac{1}{\tanh^2(\frac{\eps}{2})}-1] \ + \ \frac{d(\mathcal{S}_M)(1-d(\mathcal{S}_M))}{m} + \frac{2}{m^2 \eps^2}\\
& & \leq & \ \frac{1}{4m \tanh^2(\frac{\eps}{2})} + \frac{2}{m^2 \eps^2}
\end{aligned}
\end{equation*}
which completes the proof. To derive the last inequality, we used the fact that, since $0\leq d(\mathcal{S}_M) \leq 1$, it follows that $d(\mathcal{S}_M)(1-d(\mathcal{S}_M)) \leq \frac{1}{4}$.
\end{proof}
\end{tcolorbox}

\begin{thm}\label{thm_OptBernDE_acc2}
If the sample maintained by Algorithm \ref{OptBernDE} consists of $m = \mathcal{O}(\frac{1}{\eps^2 \alpha^2}\log{\frac{1}{\beta}})$ users from $\mathcal{U}$, then, for fixed input $\mathcal{S}$: 
$$\Prob(\ | \tilde{d}-d(\mathcal{S}) | \ \geq \ \alpha \ )\ \leq \ \beta$$
where the probability space is over the random choices of the algorithm.
\end{thm}

\begin{tcolorbox}[breakable]
\begin{proof}
The proof is similar with that of Theorem \ref{thm_DworkDE_acc2}, so we only focus on the differences.\\

Let $\hat{d}=\frac{1}{m}\sum_{i=1}^{m}{b_i}$. We again apply Lemma \ref{lemma_absolute_sum} twice, so for some $\alpha>0$ and $\delta_1, \delta_2 \in (0,1)$:
\begin{eqnarray*}
\Prob(\ | \tilde{d}-d(\mathcal{S}) | \ \geq \ \alpha \ ) &=& \Prob(\ | \tilde{d}-d(\mathcal{S}_M)+d(\mathcal{S}_M)-d(\mathcal{S}) | \ \geq \ \alpha \ ) \\
& \leq &  \Prob(\ | \tilde{d}-d(\mathcal{S}_M) | \ \geq \ \delta_1 \alpha \ ) \\ 
&& + \Prob(\ |d(\mathcal{S}_M)-d(\mathcal{S}) | \ \geq \ (1-\delta_1)\alpha \ ) \\
&=& \Prob(\ | \frac{1}{\tanh(\frac{\eps}{2})}\hat{d}-\frac{1}{2}\frac{1}{\tanh(\frac{\eps}{2})} + \frac{1}{2} \ + \ \text{Laplace}(0,\frac{1}{\eps m})\\
&& \ \ \ - d(\mathcal{S}_M) | \ \geq \ \delta_1 \alpha \ ) \\ 
&& + \ \Prob(\ |d(\mathcal{S}_M)-d(\mathcal{S}) | \ \geq \ (1-\delta_1)\alpha \ ) \\
&=& \Prob(\ | \hat{d}-\E [\hat{d}] +\tanh(\frac{\eps}{2})\text{Laplace}(0,\frac{1}{\eps m}) | \\
&& \ \ \ \geq \ \tanh(\frac{\eps}{2})\delta_1 \alpha \ ) \\ 
&& + \ \Prob(\ |d(\mathcal{S}_M)-d(\mathcal{S}) | \ \geq \ (1-\delta_1)\alpha \ ) \\
&\leq& \Prob(\ |\tanh(\frac{\eps}{2}) \text{Laplace}(0,\frac{1}{\eps m}) | \ \geq \ \tanh(\frac{\eps}{2})\delta_1 \delta_2 \alpha \ ) \\
&& +\ \Prob(\ | \hat{d}-\E [\hat{d}] | \ \geq \ \tanh(\frac{\eps}{2}) \delta_1 (1-\delta_2) \alpha\ ) \\ 
&& + \ \Prob(\ |d(\mathcal{S}_M)-d(\mathcal{S}) | \ \geq \ (1-\delta_1)\alpha \ )\\
&=& p_3 + p_2 + p_1 \ \text{ (respectively) }
\end{eqnarray*}

As in the proof of Theorem \ref{thm_DworkDE_acc2}, we have 3 sources of error to control, and we want $p_1+p_2+p_3 \leq \beta$. We bound each error probability separately, so for some $\delta_3, \delta_4 \in (0,1), \text{ such that } \delta_3 + \delta_4 < 1 $, we want:
$$p_1<\delta_3 \beta \ , \ p_2<\delta_4 \beta \ , \ p_3<(1 - \delta_3 - \delta_4) \beta$$

Bounding $p_1$ and $p_3$ is identical, as these error probabilities are unchanged. We copy the bounds we derived in Theorem \ref{thm_DworkDE_acc2}:
\begin{eqnarray*}
& p_1 & \leq 2 e^{-2m\alpha^2(1-\delta_1)^2} \leq \delta_3 \beta \\
&& \Leftrightarrow \ m \geq \underbrace{\frac{1}{2 \alpha^2 (1-\delta_1)^2} \ln(\frac{2}{\beta \delta_3})}_{m_1} \ = \ \mathcal{O}(\frac{1}{\alpha^2}\ln(\frac{1}{\beta})) \\
& p_3 & \leq e^{-\eps \alpha m \delta_1 \delta_2} \leq (1-\delta_3-\delta_4) \beta \\
&& \Leftrightarrow \ m \geq \underbrace{\frac{1}{\eps \alpha \delta_1 \delta_2} \ln(\frac{1}{\beta (1-\delta_3-\delta_4)})}_{m_3} \ = \ \mathcal{O}(\frac{1}{\eps \alpha}\ln(\frac{1}{\beta}))
\end{eqnarray*}

We focus on $p_2$.
The difference is due to the fact that we now have $\E[\hat{d}]=\frac{1}{2}(1-\tanh(\frac{\eps}{2})) + \tanh(\frac{\eps}{2})d(\mathcal{S}_M)$. Although $\E[\hat{d}]$ has changed, it still is a random mean (as $d(\mathcal{S}_M)$ is random), so we again apply the total probability theorem:
$$ p_2 = \sum_{\underline{d} \in \mathcal{D}}{ \Prob(\ | \hat{d}-\E [\hat{d}] | \ \geq \ \tanh(\frac{\eps}{2}) \delta_1 (1-\delta_2) \alpha \ | \ d(\mathcal{S}_M) = \underline{d} \ ) \ \Prob(\ d(\mathcal{S}_M) = \underline{d}\ ) }$$

Based on observations identical to those we made when proving theorem \ref{thm_DworkDE_acc2}, we again make use of an additive, two-sided Chernoff bound:

$$\Prob(\ | \hat{d}-\E [\hat{d}] | \ \geq \ \tanh(\frac{\eps}{2}) \delta_1 (1-\delta_2) \alpha \ | \ d(\mathcal{S}_M) = \underline{d} \ ) \ \leq \ 2 e^{-2m \tanh^2(\frac{\eps}{2}) \alpha^2 \delta_1^2 (1-\delta_2)^2}$$

and since the bound is independent of $\underline{d}$:

$$ p_2 \leq \sum_{\underline{d} \in \mathcal{D}}{ 2 e^{-2m \tanh^2(\frac{\eps}{2}) \alpha^2 \delta_1^2 (1-\delta_2)^2} \ \Prob(\ d(\mathcal{S}_M) = \underline{d}\ ) } = 2 e^{-2m \tanh^2(\frac{\eps}{2}) \alpha^2 \delta_1^2 (1-\delta_2)^2}$$

and, therefore, noting that $\tanh(\frac{x}{2})=\frac{x}{2}-\frac{x^3}{24}+\mathcal{O}(x^5)=\mathcal{O}(x)$:
$$ p_2 \leq \delta_4 \beta \ \Leftrightarrow \ m \geq \underbrace{\frac{1}{2 \tanh^2(\frac{\eps}{2}) \alpha^2 \delta_1^2 (1-\delta_2)^2} \ln(\frac{2}{\beta \delta_4})}_{m_2} \ = \ \mathcal{O}(\frac{1}{\eps^2 \alpha^2}\ln(\frac{1}{\beta}))$$\\

By picking $m \geq \max\{m_1,m_2,m_3\}=\mathcal{O}(\frac{1}{\eps^2 \alpha^2}\ln(\frac{1}{\beta}))$, we ensure that:
$$\Prob(\ | \tilde{d}-d(\mathcal{S}) | \ \geq \ \alpha \ )\ \leq \ p_1+p_2+p_3\ \leq \ \delta_3 \beta+\delta_4 \beta+(1-\delta_3-\delta_4) \beta \ = \ \beta$$
which completes the proof.
\end{proof}
\end{tcolorbox}

\begin{figure}%[H]%[ht!]
    \centering
    \includegraphics[width=\columnwidth]{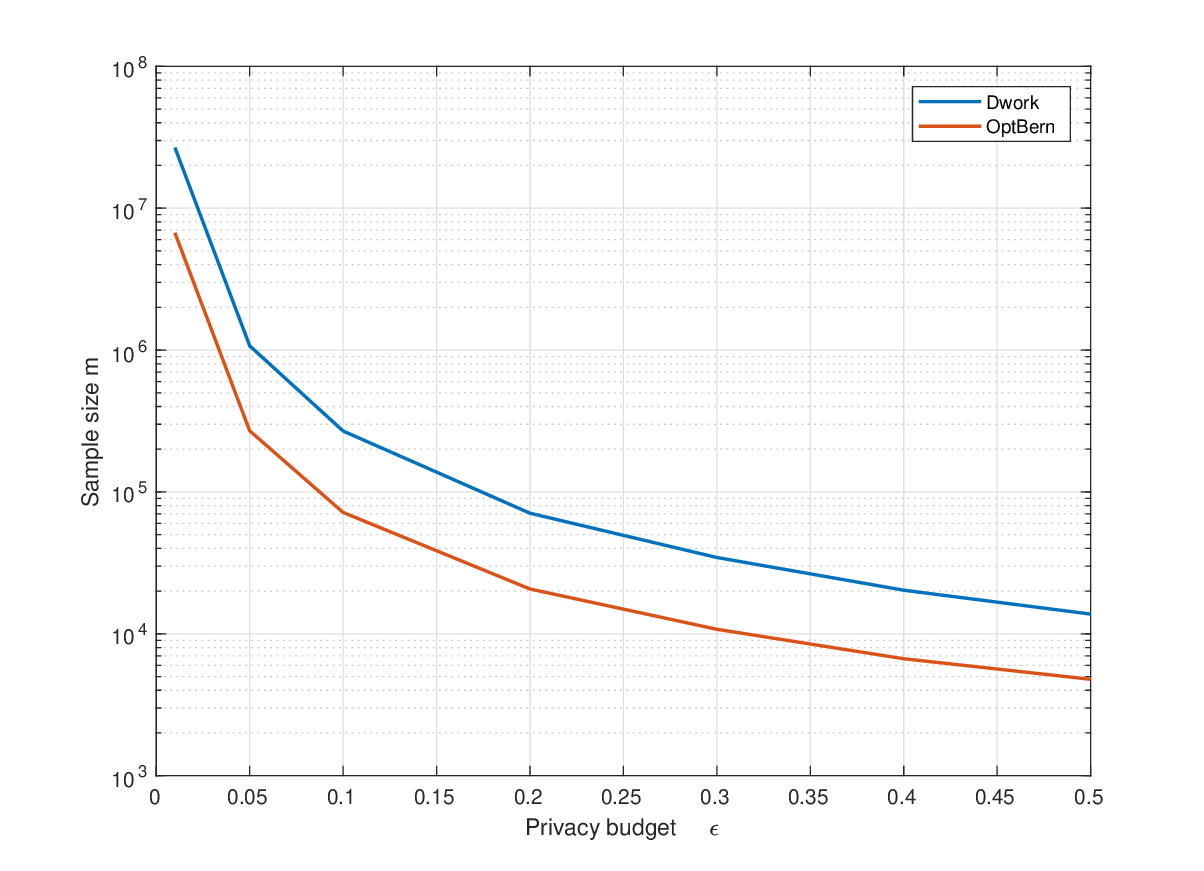}
    \caption{Proposed sample size}
	\label{FIG_sample_size_bound}
\end{figure}

Asymptotically, the lower bound on the sample size to achieve the desired approximation accuracy is not improved.
We again compute the proposed sample size by numerically solving the optimization problem described in section \ref{subSEC_sample_size}, with the slightly modified cost function:
\begin{equation*}
\begin{aligned}
m(\delta_1,\delta_2,\delta_3,\delta_4) &=& \max \{ \ & m_1 , m_2 , m_3 \ \} \\
&=& \max \{ \ &\frac{1}{2 \alpha^2 (1-\delta_1)^2} \ln(\frac{2}{\beta \delta_3}) \ ,\\
& & & \frac{1}{2 \tanh^2(\frac{\eps}{2}) \alpha^2 \delta_1^2 (1-\delta_2)^2} \ln(\frac{2}{\beta \delta_4}) \ ,\\
& & & \frac{1}{\eps \alpha \delta_1 \delta_2} \ln(\frac{1}{\beta (1-\delta_3-\delta_4)}) \  \}
\end{aligned}
\end{equation*}
The resulting proposed sample size is illustrated in Figure \ref{FIG_sample_size_bound}.

\section{Laplace Density Estimator}
In this section, we examine an alternative modification to Algorithm \ref{DworkDE}.
Although the algorithm we derive fails to match the performance of Algorithm \ref{OptBernDE}, it also manages to outperform the original algorithm and may provide useful theoretical insights.

\subsection{Using Continuous Distributions}
We again focus on modifying the state of Algorithm \ref{DworkDE}.
The main idea is to replace the bitarray $\mathbf{b}$ (which stores bits drawn from either of the two Bernoulli distributions $f_{init}$ and $f_{upd}$), by an array of real numbers $\textbf{x}$, drawn from two continuous distribution.
Our motivation is that the algorithm's output is itself a real number, so by storing a flexible, real value per user (instead of a hard, binary value), we expect to boost accuracy.
Although this approach increases the algorithm's space requirements ($\times 32$ or $\times 64$ bits per user if floating point arithmetic is used), we argue that, to achieve the same levels of accuracy, we will need to maintain a sample of less users, and we will therefore offer perfect privacy for more users.

We propose using the following two Laplace distributions:
$$f_{init} = \text{Laplace}(\mu_{init},b)$$
$$f_{upd} = \text{Laplace}(\mu_{upd},b)$$
where $\mu_{init} < \mu_{upd}$ and the scale of the two distributions is the same. Therefore, if a (sampled) user does not appear in the stream, his entry in \textbf{x} will be drawn from $\text{Laplace}(\mu_{init},b)$, whereas if a user appears, his entry will be drawn from $\text{Laplace}(\mu_{upd},b)$.

Before presenting the new algorithm (Algorithm \ref{LaplaceDE}), we examine the privacy guarantees of the modified state. As we show, the definition of differential privacy determines the values the parameters of the two distributions may take.

\begin{thm}\label{thm_LaplaceDE_priv}
Assume $\eps \leq \frac{1}{2}$. If $\frac{\mu_{upd} - \mu_{init}}{b} = \eps$, then Algorithm \ref{LaplaceDE} satisfies $2\eps$-pan-privacy.
\end{thm}

\begin{tcolorbox}[breakable]
\begin{proof}
The proof is identical with that of Theorem \ref{thm_DworkDE_priv}.
The only modification is on proving that the modified state \textbf{x} (array of real values) satisfies $\eps$-differential privacy, and in particular, for a user $u \in M$ that appears in stream $\mathcal{S}$ and does not appear in stream $\mathcal{S}'$ (again, let $x_1$ be the entry that corresponds to $u$ in the array), we want, $\forall x \in (-\infty,\infty)$: 
\begin{eqnarray*}
\frac{\Prob( \ \mathbf{x(\mathcal{S})}=[x \ \underline{x_2} \ ...\ \underline{x_{m}}]\ )}{\Prob( \ \mathbf{x(\mathcal{S'})}=[x \ \underline{x_2} \ ...\ \underline{x_{m}}]\ )} &=& \frac{\Prob( \ x_1(\mathcal{S})=x\ )}{\Prob( \ x_1(\mathcal{S'})=x\ )} \\
&=& \frac{f_{upd}(x)}{f_{init}(x)} \\
&=& \frac{\frac{1}{2b}e^{-\frac{|x-\mu_{upd}|}{b}}}{\frac{1}{2b}e^{-\frac{|x-\mu_{init}|}{b}}} \\
&=& e^{\frac{|x-\mu_{init}|-|x-\mu_{upd}|}{b}} \\
& \leq & e^{\frac{|x-\mu_{init}-x+\mu_{upd}|}{b}} \\
&=& e^{\frac{\mu_{upd}-\mu_{init}}{b}} \\
&=& e^{\eps}
\end{eqnarray*}
where the inequality holds due to the triangle inequality. Proving that $\frac{f_{upd}(x)}{f_{init}(x)} \geq e^{-\eps}$ is identical. Therefore, user $u$ is guaranteed $\eps$-differential privacy against an adversary that observes $u$'s entry in $\mathbf{x}$.
\end{proof}
\end{tcolorbox}

\begin{figure}%[H]%[ht!]
    \centering
    \includegraphics[width=\columnwidth]{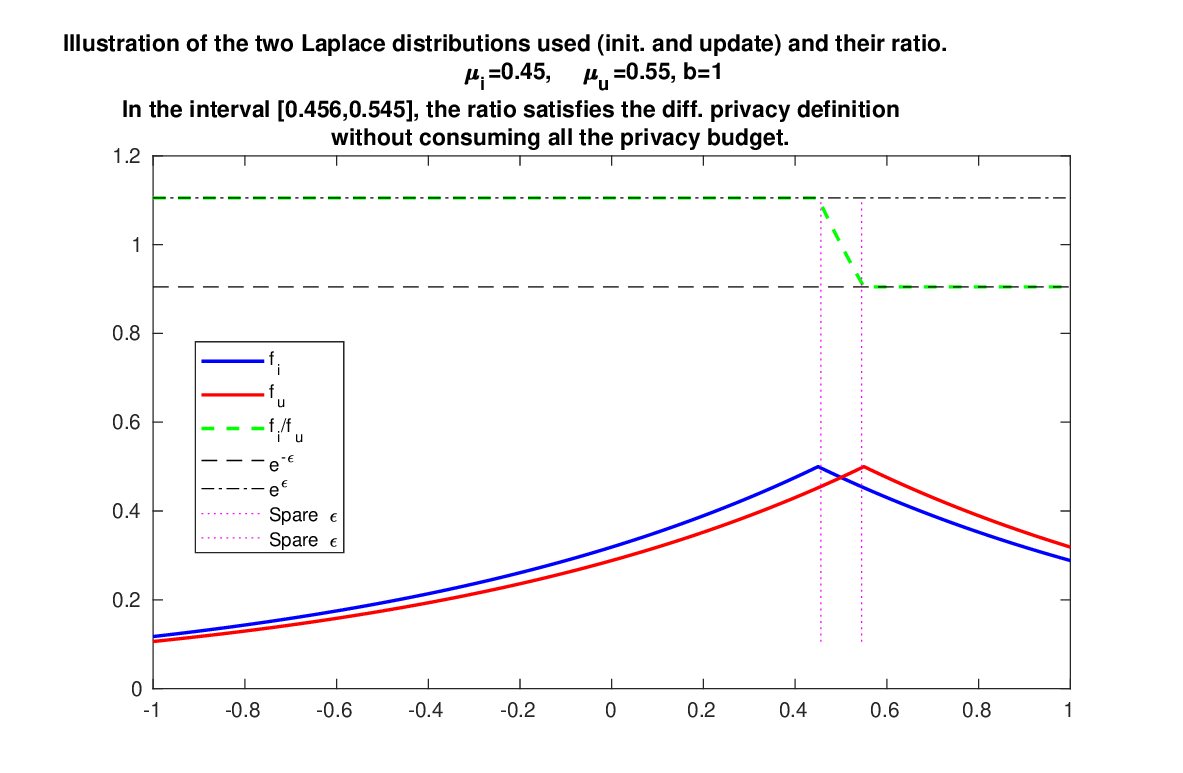}
    \caption{Illustration of the proposed Laplace distributions}
	\label{laplace_distributions_illustration}
\end{figure}

Figure \ref{laplace_distributions_illustration} provides an example illustration of the two distributions, for $\eps = 0.1$. Observe that the ratio of the two distributions tightly satisfies the differential privacy definition, except for some $x \in (\mu_{init},\mu_{upd})$. However, since $\exists x$ such that the ratio is almost equal to either $e^{\eps}$ or $e^{-\eps}$, we conclude that Algorithm \ref{LaplaceDE} uses all the allocated privacy budget.

To achieve the best possible accuracy, we would want to maximize the difference $\mu_{upd}-\mu_{init}$, as well as to pick the minimum possible $b$.
In doing so, it would have been easier to distinguish between users that did not appear and users that appeared in the stream, during the density estimation computation.
However, what Theorem \ref{thm_LaplaceDE_priv} tells us is that we only have one degree of freedom,
so increasing the difference of the means, leads to an increase in the scale of the distributions (that is, the distributions become more uniform, and, hence, harder to distinguish).

\subsection{Estimator \& Analysis}
We now present the new estimator, namely the Laplace Density Estimator.

\begin{algorithm}%[H]

\caption{Laplace Density Estimator\label{LaplaceDE}}

\DontPrintSemicolon

\KwIn{Data stream $\mathcal{S}$, Privacy budget $\eps$, Accuracy parameters $(\alpha,\beta)$}

\KwOut{Density $\tilde{d}(\mathcal{S})$}

Compute $m^*$ and set $m=m^*$\;

Sample a random subset $M \subseteq \mathcal{U}$ of $m$ users (without replacement) and define an arbitrary ordering over $M$\;

Create an array $\mathbf{x}=[x_1\ ...\ x_{m}]$ and map $M[i] \rightarrow x_i,\ \forall \ i \in \{1,...,m\}$\;

Arbitrarily pick $\mu_{init},\mu_{upd}$ and set $b=\frac{\mu_{upd}-\mu_{init}}{\eps}$\;

Initialize $\mathbf{x}$ randomly: $x_i \sim \text{Laplace}(\mu_{init},b), \ \forall \ i \in \{1,...,m\}$\;

\For{t = 1 \KwTo T}{

	\If{$s_t \in M$}{
		
		Find $i:\ M[i]=s_t$\;
		
		Re-sample: $x_i \sim \text{Laplace}(\mu_{upd},b)$\;
			
	}
}

Return $\tilde{d}(\mathcal{S})\ = \ \frac{1}{\mu_{upd}-\mu_{init}}(\frac{1}{m}\sum_{i=1}^{m}{x_i}-\mu_{init} )\ + \ \text{Laplace}(0,\frac{1}{\eps m})$\;

\end{algorithm}

We have already argued on the algorithm's privacy guarantees, so we proceed with the accuracy analysis.

\begin{thm}\label{thm_LaplaceDE_acc1}
For fixed sample $M$, Algorithm \ref{LaplaceDE} provides an unbiased estimate $\tilde{d}$ of the density of $\mathcal{S}_M$ and has mean squared error: 
$$\E [ \ (\tilde{d}-d(\mathcal{S}_M))^2 \ ] \ = \ \frac{2(m+1)}{m^2 \eps^2}$$
\end{thm}

\begin{tcolorbox}[breakable]
\begin{proof}
The proof is similar with that of Theorem \ref{thm_DworkDE_acc1}.\\

We begin with the \emph{bias} computation. We examine the distribution of an arbitrary entry in $\textbf{x}$:
\[ 
x_i \sim \left\{
\begin{array}{ll}
      \text{Laplace}(\mu_{init},b) & M[i] \not\in \mathcal{S}_M\\
      \text{Laplace}(\mu_{upd},b) & M[i] \in \mathcal{S}_M \\
\end{array} 
\right. 
\]

Denoting $\hat{d}=\frac{1}{m}\sum_{i=1}^{m}{x_i}$, and applying a similar computation as in the proof of Theorem \ref{thm_DworkDE_acc1}, gives: 
\begin{eqnarray*}
\E[\hat{d}] &=& \mu_{init} + (\mu_{upd}-\mu_{init})d(\mathcal{S}_M)
\end{eqnarray*}

The final estimate (output) $\tilde{d}$ is:  
\begin{eqnarray*}
\tilde{d} &=& \frac{\hat{d}-\mu_{init}}{\mu_{upd}-\mu_{init}} + \text{Laplace}(0,\frac{1}{\eps m})
\end{eqnarray*}
so it is an unbiased estimate.\\

We proceed with the \emph{mean squared error}. Again, since $\tilde{d}$ is an unbiased estimate of $d(\mathcal{S}_M)$, its mean squared error coincides with its variance.
We now have $m$ Laplace random variables $x_i \ (i=1,...,m)$ with the same scale parameter $b$, so:
\begin{equation*}
\begin{aligned}
& \text{var}(x_i) & = & \ 2b^2 \\
\Rightarrow \ & \text{var}(\hat{d}) & = & \ \frac{1}{m^2}\sum_{i=1}^{m}{ \text{var}(x_i)} \ = \ \frac{2b^2}{m} \\
\Rightarrow \ & \text{var}(\tilde{d}) & = & \ (\frac{1}{\mu_{upd}-\mu_{init}})^2 \text{var}(\hat{d}) + \text{var}(\text{Laplace}(0,\frac{1}{\eps m})) \\
& & = & \ \frac{2b^2}{m(\mu_{upd}-\mu_{init})^2} + \frac{2}{m^2 \eps^2}\\
& & = & \ \frac{2b^2}{m(b \eps)^2} + \frac{2}{m^2 \eps^2}\\
& & = & \ \frac{2(m+1)}{m^2 \eps^2}
\end{aligned}
\end{equation*}
which completes the proof.
\end{proof}
\end{tcolorbox}

We conclude that neither the bias, nor the variance (and mean squared error) of our estimator depends on the choice of $\mu_{init}$, $\mu_{upd}$ and $b$.

Before proceeding with the next theorem, we give the following definition and provide some useful lemmas.

\begin{defn}\label{defn_subexp}
A random variable $X$ with mean $\E [X] = \mu$ is sub-exponential if there are non-negative parameters $(v_s,b_s)$ such that:
$$ \E [ e^{\lambda (X-\mu) } ] \leq e^{\frac{v_s^2 \lambda^2}{2}} \ , \ \forall |\lambda| < \frac{1}{b_s}$$
\end{defn}

The control on the moment generating function, when combined with the Chernoff technique allows us to prove the following concentration inequality, called the sub-exponential tail bound. We omit the proof, as it is based on a standard technique.

\begin{lemma}\label{lemma_subexp_tail}
Suppose that random variable $X$ is sub-exponential with parameters $(v_s,b_s)$. Then,
\[ 
\Prob (|X-\E[X]| \geq \alpha) \leq \left\{
\begin{array}{ll}
      e^{-\frac{\alpha^2}{2v_s^2}} & 0 \leq \alpha \leq \frac{v_s^2}{b_s} \\
      e^{-\frac{\alpha}{2b_s}} & \alpha > \frac{v_s^2}{b_s} \\
\end{array} 
\right. 
\]
\end{lemma}

We observe that, when $\alpha$ is small enough, the exponent is quadratic in $\alpha$, whereas for larger $\alpha$, the exponent is linear (so the bound is not as tight).

\begin{lemma}\label{lemma_subexp1}
Let $ X \sim \text{Laplace}(\mu,b)$. Then $X$ is sub-exponential with parameters $(v_s,b_s)=(2b,\sqrt{2}b)$.
\end{lemma}

\begin{tcolorbox}[breakable]
\begin{proof}
The moment generating function of the random variable $X-\mu$ is:
$$\E [e^{\lambda ( X - \mu ) }] = \frac{1}{1-b^2\lambda^2}\ , \ \forall |\lambda|<\frac{1}{b}$$
Taking into account that: $\frac{1}{1-x} \leq 1+2x \leq e^{2x} , \ \forall \ 0<x<\frac{1}{2}$
and setting $x=b^2 \lambda^2$, gives:
$$\E [e^{\lambda ( X - \mu )}] \leq e^{2b^2 \lambda^2}\ , \ \forall |\lambda|<\frac{1}{\sqrt{2}b}$$
We therefore set $v_s=2b$, which gives:
$$\E [e^{\lambda ( X - \mu )}] \leq e^{\frac{v_s^2 \lambda^2}{2}}\ , \ \forall |\lambda|<\frac{1}{\sqrt{2}b}$$
and proves the desired result, according to Definition \ref{defn_subexp}.
\end{proof}
\end{tcolorbox}

\begin{lemma}\label{lemma_subexp2}
Let $ X = \sum_{i=1}^m{X_i}$, where $ X_i \sim \text{Laplace}(\mu_i,b)$. Then $X$ is sub-exponential with parameters $(v_s,b_s)=(2b\sqrt{m},\sqrt{2}b)$.
\end{lemma}

\begin{tcolorbox}[breakable]
\begin{proof}
The moment generating function of the random variable $X-\E[X]$ is:
$$\E [e^{\lambda ( X - \E [X] ) }] = \prod_{i=1}^m{e^{\lambda ( X_i - \mu_i ) }} \leq  \prod_{i=1}^m{e^{2b^2 \lambda^2}} = e^{2b^2 \lambda^2 m}\ , \ \forall |\lambda|<\frac{1}{\sqrt{2}b}$$
where the inequality holds due to Lemma \ref{lemma_subexp1}. We therefore set $v_s=2b\sqrt{m}$, which gives:
$$\E [e^{\lambda ( X - \E [X] )}] \leq e^{\frac{v_s^2 \lambda^2}{2}}\ , \ \forall |\lambda|<\frac{1}{\sqrt{2}b}$$
and proves the desired result, according to Definition \ref{defn_subexp}.
\end{proof}
\end{tcolorbox}

\begin{lemma}\label{lemma_laplace_tail}
Let $ X = \sum_{i=1}^m{X_i}$, where $ X_i \sim \text{Laplace}(\mu_i,b)$. Then,
\[ 
\Prob (|X-\E[X]| \geq \alpha) \leq \left\{
\begin{array}{ll}
      e^{-\frac{\alpha^2}{8b^2m}} & 0 \leq \alpha \leq 2\sqrt{2}bm \\
      e^{-\frac{\alpha}{2\sqrt{2}b}} & \alpha > 2\sqrt{2}bm \\
\end{array} 
\right. 
\]
\end{lemma}

\begin{tcolorbox}[breakable]
\begin{proof}
The result follows by simply combining Lemma \ref{lemma_subexp_tail} and Lemma \ref{lemma_subexp2}.
\end{proof}
\end{tcolorbox}

\begin{thm}\label{thm_LaplaceDE_acc2}
If the sample maintained by Algorithm \ref{LaplaceDE} consists of $m = \mathcal{O}(\frac{1}{\eps^2 \alpha^2}\log{\frac{1}{\beta}})$ users from $\mathcal{U}$, then, for fixed input $\mathcal{S}$: 
$$\Prob(\ | \tilde{d}-d(\mathcal{S}) | \ \geq \ \alpha \ )\ \leq \ \beta$$
where the probability space is over the random choices of the algorithm.
\end{thm}

\begin{tcolorbox}[breakable]
\begin{proof}
The proof is similar with that of Theorem \ref{thm_DworkDE_acc2}, so we again only focus on the differences.\\

Let $\hat{d}=\frac{1}{m}\sum_{i=1}^{m}{x_i}$ and recall that $\mu_{upd}-\mu_{init}=\eps b$ (by the differential privacy requirement). We again apply Lemma \ref{lemma_absolute_sum} twice, so for some $\alpha>0$ and $\delta_1, \delta_2 \in (0,1)$:
\begin{eqnarray*}
\Prob(\ | \tilde{d}-d(\mathcal{S}) | \ \geq \ \alpha \ ) &=& \Prob(\ | \tilde{d}-d(\mathcal{S}_M)+d(\mathcal{S}_M)-d(\mathcal{S}) | \ \geq \ \alpha \ ) \\
& \leq &  \Prob(\ | \tilde{d}-d(\mathcal{S}_M) | \ \geq \ \delta_1 \alpha \ ) \\ 
&& + \Prob(\ |d(\mathcal{S}_M)-d(\mathcal{S}) | \ \geq \ (1-\delta_1)\alpha \ ) \\
&=& \Prob(\ | \frac{\hat{d}}{\mu_{upd}-\mu_{init}}-\frac{\mu_{init}}{\mu_{upd}-\mu_{init}} + \text{Laplace}(0,\frac{1}{\eps m})\\
&& \ \ \ - d(\mathcal{S}_M) | \ \geq \ \delta_1 \alpha \ ) \\ 
&& + \ \Prob(\ |d(\mathcal{S}_M)-d(\mathcal{S}) | \ \geq \ (1-\delta_1)\alpha \ ) \\
&=& \Prob(\ | \hat{d}-\E [\hat{d}] +\eps b \text{Laplace}(0,\frac{1}{\eps m}) | \ \geq \ \eps b \delta_1 \alpha \ ) \\ 
&& + \ \Prob(\ |d(\mathcal{S}_M)-d(\mathcal{S}) | \ \geq \ (1-\delta_1)\alpha \ ) \\
&\leq & \Prob(\ |\eps b \text{Laplace}(0,\frac{1}{\eps m}) | \ \geq \ \eps b \delta_1 \delta_2 \alpha \ ) \\
&& +\ \Prob(\ | \hat{d}-\E [\hat{d}] | \ \geq \ \eps b \delta_1 (1-\delta_2) \alpha\ ) \\ 
&& + \ \Prob(\ |d(\mathcal{S}_M)-d(\mathcal{S}) | \ \geq \ (1-\delta_1)\alpha \ )\\
&=& p_3 + p_2 + p_1 \ \text{ (respectively) }
\end{eqnarray*}

As in the proof of Theorem \ref{thm_DworkDE_acc2}, we have 3 sources of error to control, and we want $p_1+p_2+p_3 \leq \beta$. We bound each error probability separately, so for some $\delta_3, \delta_4 \in (0,1), \text{ such that } \delta_3 + \delta_4 < 1 $, we want:
$$p_1<\delta_3 \beta \ , \ p_2<\delta_4 \beta \ , \ p_3<(1 - \delta_3 - \delta_4) \beta$$

Bounding $p_1$ and $p_3$ is identical, as these error probabilities are unchanged. We copy the bounds we derived in Theorem \ref{thm_DworkDE_acc2}:
\begin{eqnarray*}
& p_1 & \leq 2 e^{-2m\alpha^2(1-\delta_1)^2} \leq \delta_3 \beta \\
&& \Leftrightarrow \ m \geq \underbrace{\frac{1}{2 \alpha^2 (1-\delta_1)^2} \ln(\frac{2}{\beta \delta_3})}_{m_1} \ = \ \mathcal{O}(\frac{1}{\alpha^2}\ln(\frac{1}{\beta})) \\
& p_3 & \leq e^{-\eps \alpha m \delta_1 \delta_2} \leq (1-\delta_3-\delta_4) \beta \\
&& \Leftrightarrow \ m \geq \underbrace{\frac{1}{\eps \alpha \delta_1 \delta_2} \ln(\frac{1}{\beta (1-\delta_3-\delta_4)})}_{m_3} \ = \ \mathcal{O}(\frac{1}{\eps \alpha}\ln(\frac{1}{\beta}))
\end{eqnarray*}

We focus on $p_2$.
We again apply the total probability theorem:
$$ p_2 = \sum_{\underline{d} \in \mathcal{D}}{ \Prob(\ | \hat{d}-\E [\hat{d}] | \ \geq \ \eps b \delta_1 (1-\delta_2) \alpha \ | \ d(\mathcal{S}_M) = \underline{d} \ ) \ \Prob(\ d(\mathcal{S}_M) = \underline{d}\ ) }$$

For fixed $d(\mathcal{S}_M)$, $\E [\hat{d}]$ is no longer random. Now $\hat{d}$ is a weighted sum of Laplace random variables $x_i$, so by Lemma \ref{lemma_laplace_tail}:

\begin{equation*}
\begin{aligned}
& \Prob(\ | \hat{d}-\E [\hat{d}] | \ \geq \ \eps b \delta_1 (1-\delta_2) \alpha \ | \ d(\mathcal{S}_M) = \underline{d} \ ) \\
& = \Prob(\ | \sum_{i=1}^{m}{x_i}-\E [\sum_{i=1}^{m}{x_i}] | \ \geq \ m \eps b \delta_1 (1-\delta_2) \alpha \ | \ d(\mathcal{S}_M) = \underline{d} \ ) \\
& \leq 2 e^{- \frac{ [m \eps b \alpha \delta_1 (1-\delta_2)]^2 }{8b^2m} } \ = \ 2 e^{- \frac{1}{8} m \eps^2 \alpha^2 \delta_1^2 (1-\delta_2)^2  }
\end{aligned}
\end{equation*}

where we used the fact that, if $\alpha<4$, we are in the area where the exponent in the tail bound is quadratic,
since for $ \eps \in (0,\frac{1}{2}]$ and $\delta_1,\delta_2 \in (0,1)$:
$$m \eps b \delta_1 (1-\delta_2) \alpha < m \eps b \alpha < 2bm$$
In practice, since the density takes values in the $[0,1]$ interval, we only care about additive deviations $\alpha \in [0,1]$.

Again, the bound is independent of $\underline{d}$, so:

$$ p_2 \leq \sum_{\underline{d} \in \mathcal{D}}{ 2 e^{- \frac{1}{8} m \eps^2 \alpha^2 \delta_1^2 (1-\delta_2)^2  } \ \Prob(\ d(\mathcal{S}_M) = \underline{d}\ ) } = 2 e^{- \frac{1}{8} m \eps^2 \alpha^2 \delta_1^2 (1-\delta_2)^2  }$$

and, therefore:
$$ p_2 \leq \delta_4 \beta \ \Leftrightarrow \ m \geq \underbrace{\frac{8}{\eps^2 \alpha^2 \delta_1^2 (1-\delta_2)^2} \ln(\frac{2}{\beta \delta_4})}_{m_2} \ = \ \mathcal{O}(\frac{1}{\eps^2 \alpha^2}\ln(\frac{1}{\beta})) $$\\

By picking $m \geq \max\{m_1,m_2,m_3\}=\mathcal{O}(\frac{1}{\eps^2 \alpha^2}\ln(\frac{1}{\beta}))$, we ensure that:
$$\Prob(\ | \tilde{d}-d(\mathcal{S}) | \ \geq \ \alpha \ )\ \leq \ p_1+p_2+p_3\ \leq \ \delta_3 \beta+\delta_4 \beta+(1-\delta_3-\delta_4) \beta \ = \ \beta$$
which completes the proof.
\end{proof}
\end{tcolorbox}

Observe that the bound we derive is identical with that of Algorithm \ref{DworkDE} (proof of Theorem \ref{thm_DworkDE_acc2}), both asymptotically and in terms of constants. Therefore, the cost function we use to compute the proposed sample size for Algorithm \ref{LaplaceDE} is the same with that presented in section \ref{subSEC_sample_size}, so we expect the solution to the optimization problem to be the same as well.

\subsection{Quantized Laplace Density Estimator}
In order to reduce the space requirements of Algorithm \ref{LaplaceDE}, we also examine the impact of quantization on the performance of the algorithm.
In particular, instead of storing floating point numbers in the array \textbf{x}, we quantize each user's entry (which is a real value sampled either from $f_{init}$ or $f_{upd}$) into $L$ fixed quantization levels, and store the quantized value.
Thus, the only modification in Algorithm \ref{LaplaceDE} is that, after drawing a value $x$ from either $f_{init} = \text{Laplace}(\mu_{init},b)$, or $f_{upd} = \text{Laplace}(\mu_{upd},b)$, we quantize it (using our pre-designed quantizer), and store $q(x)$.

In general, a quantizer consists of $L$ quantization regions, which are separated by $L-1$ boundaries $B_1,...,B_{L-1}$, and $L$ representation points $a_1,...,a_{L}$.
Assume that we wish to quantize real numbers drawn from a random variable $x_i$ with pdf $f_{x_i}(x), \ x \in \mathbb{R}$, and that $q(x_i)$ is the resulting random variable after the quantization. 
Then, the minimum mean squared error quantizer is constructed by selecting the boundaries and representation points that minimize:
$$ \E [ (x_i - q(x_i) )^2 ] = \sum_{j=1}^{L}{ \int_{B_{j-1}}^{B_j}{f_{x_i}(x) (x-a_j)^2 dx} } $$
where $B_0=-\infty$ and $B_L=\infty$.
If both the boundaries and the representation points are unknown, an iterative algorithm can be used to find them (e.g. Lloyd-Max). 
However, for fixed boundaries, it is a well known result that the representation points that minimize the mean squared error are selected as:
$$ a_j = \E [x_i|B_{j-1}<x_i\leq B_j] = \int_{B_{j-1}}^{B_j}{x f_{x_i|B_{j-1}<x_i\leq B_j}(x) dx}\ , \ j=1,...,L$$

In what follows, we focus on the case of $L=2$, which in fact leads us back to storing a single bit per user - this case will be proven to be particularly interesting, and it inspired us in designing the Optimal Bernoulli Density Estimator (Algorithm \ref{OptBernDE}).
We briefly present the design of our quantizer.
\begin{itemize}
\item[-] The distribution of each $x_i$ depends on the density of the stream: for a sparse stream, $x_i$ is more likely to have been drawn from $f_{init}$, whereas for a dense stream, $x_i$ is more likely to have been drawn from $f_{upd}$.
Since we have no prior knowledge on the density of the stream, we assume that $x_i$ is equally likely to have been drawn from either distribution, which gives us the following mixture distribution:
$$f_{x_i}(x) = \frac{1}{2} f_{init}(x) + \frac{1}{2} f_{upd}(x) \ , \ x \in \mathbb{R}$$

\item[-] For $L=2$, we set the boundary $B=\frac{\mu_{init}+\mu_{upd}}{2}$ due to symmetry.
Therefore, we only have to pick the proper representation points to minimize the mean squared error.

\item[-] By simple calculus, the representation points are:
\begin{equation*}
\begin{aligned}
a_1 & \ = \ & \int_{-\infty}^{B}{x \frac{f_{x_i}(x)}{2}dx} & \ = \ & \mu_{init} - \frac{\mu_{upd}-\mu_{init}}{\eps} e^{-\frac{\eps}{2}} \\
a_2 & \ = \ & \int_{B}^{\infty}{x \frac{f_{x_i}(x)}{2}dx} & \ = \ & \mu_{upd} + \frac{\mu_{upd}-\mu_{init}}{\eps} e^{-\frac{\eps}{2}}
\end{aligned}
\end{equation*}
For example, we may want to end up with $a_1=0$ and $a_2=1$. To do so, it is easy to see that we should pick: $\mu_{init} = \frac{e^{-\frac{\eps}{2}}}{2e^{-\frac{\eps}{2}}+\eps}$ and $\mu_{upd} = \frac{e^{-\frac{\eps}{2}}+\eps}{2e^{-\frac{\eps}{2}}+\eps}$.
\end{itemize}

We now have our binary quantizer at hand.
Assume that a value $x \in \mathbb{R}$ is drawn for the user whose entry in \textbf{x} is $x_i$;
if $x<B=\frac{\mu_{init}+\mu_{upd}}{2}$, we set $x_i=a_1$, otherwise, we set $x_i=a_2$.
We examine the distribution of $q(x_i)$, which is a binary random variable:
\begin{eqnarray*}
\Prob (q(x_i)=1 | M[i] \not\in \mathcal{S}) &=& \Prob (\text{Laplace}(\mu_{init},b) \geq \frac{\mu_{init}+\mu_{upd}}{2}) \\
&=& \frac{1}{2} e^{\frac{\frac{\mu_{init}+\mu_{upd}}{2}+\mu_{init}}{b}}
= \frac{1}{2} e^{-\frac{\eps}{2}} \\
\Prob (q(x_i)=1 | M[i] \in \mathcal{S}) &=& \Prob (\text{Laplace}(\mu_{upd},b) \geq \frac{\mu_{init}+\mu_{upd}}{2}) \\
&=& 1 - \frac{1}{2} e^{\frac{\frac{\mu_{init}+\mu_{upd}}{2}-\mu_{upd}}{b}}
= 1 - \frac{1}{2} e^{-\frac{\eps}{2}}
\end{eqnarray*}
where we used the fact that $\mu_{upd}-\mu_{init}=\eps b$.
Therefore, by picking $\mu_{upd}$ and $\mu_{init}$ such that $a_1=0$ and $a_2=1$, the quantized Laplace Density Estimator can be viewed as a Bernoulli Density Estimator, that utilizes the following two Bernoulli distributions:
$$f_{init} = \text{Bernoulli}(\frac{1}{2} e^{-\frac{\eps}{2}})$$
$$f_{upd} = \text{Bernoulli}(1 - \frac{1}{2} e^{-\frac{\eps}{2}})$$

\begin{figure}%[H]%[ht!]
    \centering
    \includegraphics[width=0.75\columnwidth]{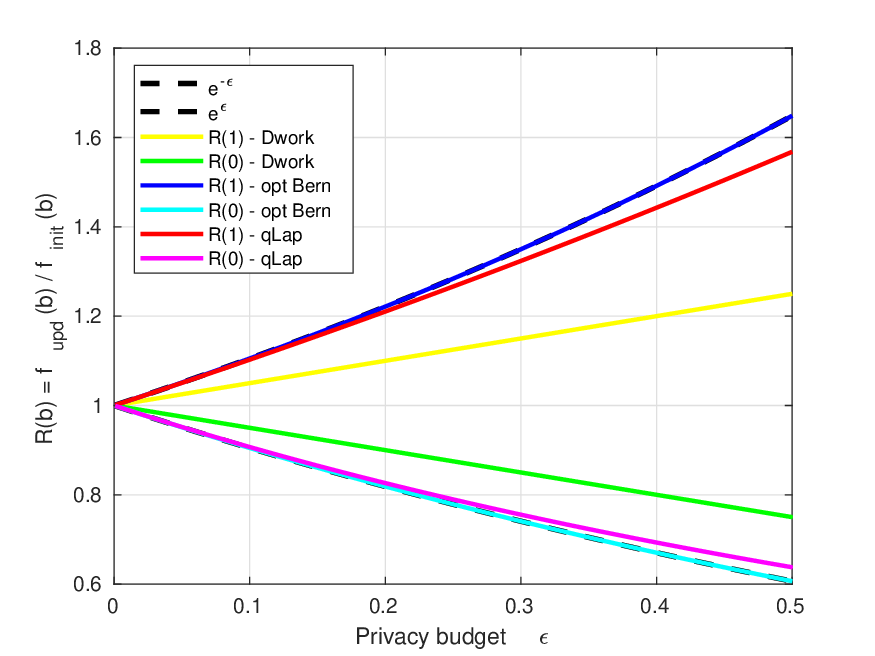}
    \caption{Quantized Laplace - State differential privacy}
	\label{FIG_qLap_privacy_budget}
\end{figure}

In Figure \ref{FIG_qLap_privacy_budget} we examine the use of the allocated privacy budget of the derived Bernoulli distributions, and compare it with that of Algorithm \ref{DworkDE} and \ref{OptBernDE}.

\subsection{Comments on the Use of Other Continuous Distributions}
The connections of the Laplace distribution with the definition of differential privacy are well understood, so our choice of this particular continuous distribution is quite natural.\\

\underline{\textbf{Truncated Distributions.}} One may argue that it would be beneficial to truncate the utilized distributions, in order to avoid large (either positive or negative) values, which may lead to a poor estimate.

For example, it would be a reasonable choice to truncate the distributions and allow them to take values in the $[0,1]$ interval, so that each user's entry is interpreted as a probability of the user having appeared in the stream.
Taking into account that the truncated probability density that corresponds to a pdf $f(x), \ x \in (-\infty,\infty)$, is computed as $f_{trunc}(x) = \frac{f(x)}{\int_{-\infty}^1{f(x)}-\int_{-\infty}^0{f(x)}}, \ \forall x \in [0,1]$, we would pick the distributions' means symmetric around $\frac{1}{2}$, so that the normalization factors cancel out when examining the ratio $\frac{f_{upd}}{f_{init}}$ (in order to satisfy differential privacy).
If user $u$ (whose entry is $x_1$) has not appeared, then $x_1$ is more likely to be closer to 0, whereas if $u$ has appeared, then $x_1$ is more likely to be closer to 1.

However, the truncation appears to be problematic.
The truncated distributions' means come closer to each other (because of the truncation), while the (common) scale parameter does not change.
Consequently, for fixed privacy budget, we end up with two harder-to-distinguish distributions.\\

\underline{\textbf{Gaussian Distributions.}}
The incompatibility of the Gaussian distribution with $\eps$-differential privacy is well understood, and unfortunately appears in our setting as well.
In particular, the problem arises when attempting to keep the ratio $\frac{f_{upd}}{f_{init}}$ bounded between $e^{-\eps}$ and $e^{\eps}$,
and is due to the square in the exponent of the Gaussian distribution.
Although we managed to keep the ratio bounded by truncating the Gaussian distributions and properly selecting their means and (common) variance, the performance of the derived estimator is not promising.

\section{Experimental Evaluation}
In this section, we experimentally compare the algorithms we presented.
We conduct all our experiments in MATLAB.
In each experiment, we generate a stream of length $T=10^5$.
The universe is the set $\mathcal{U}=\{1,...,10^5\}$ and the stream is either uniform or zipfian (with parameter 1).
In the former case we expect to observe a stream density of $0.63$, while in the latter of $0.25$.

In our first set of experiments, we examine the mean squared error of the algorithms, as a function of the allocated privacy budget.
We remark that in those experiments, we do not take into account the input parameters $\alpha,\beta$ in order to pick the proper sample size;
instead we fix the sample size (as a fraction of the universe size) and examine how each the algorithm performs, for varying $\eps$.
For each $\eps$, we independently repeat the experiment $300$ times.
Notice that in each sub-figure we also plot the theoretical mean squared error (exact or upper bound) of each algorithm.

\begin{figure}%[H]%[ht!]
    \centering
    \includegraphics[width=\columnwidth]{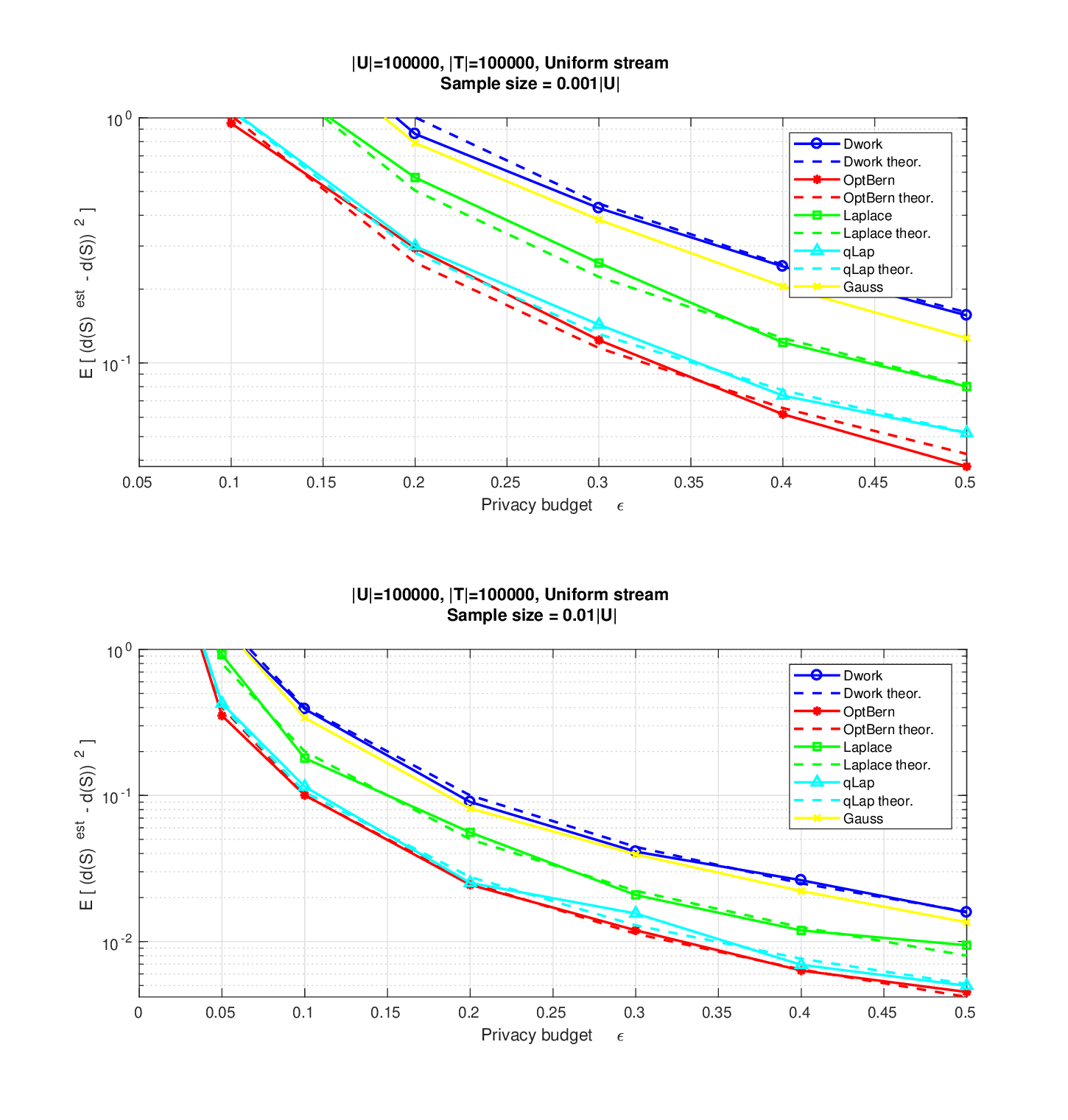}
    \caption{MSE for varying privacy budget, and for various sample sizes (Uniform Stream)}
	\label{FIG_MSE_uniform}
\end{figure}

\begin{figure}%[H]%[ht!]
    \centering
    \includegraphics[width=\columnwidth]{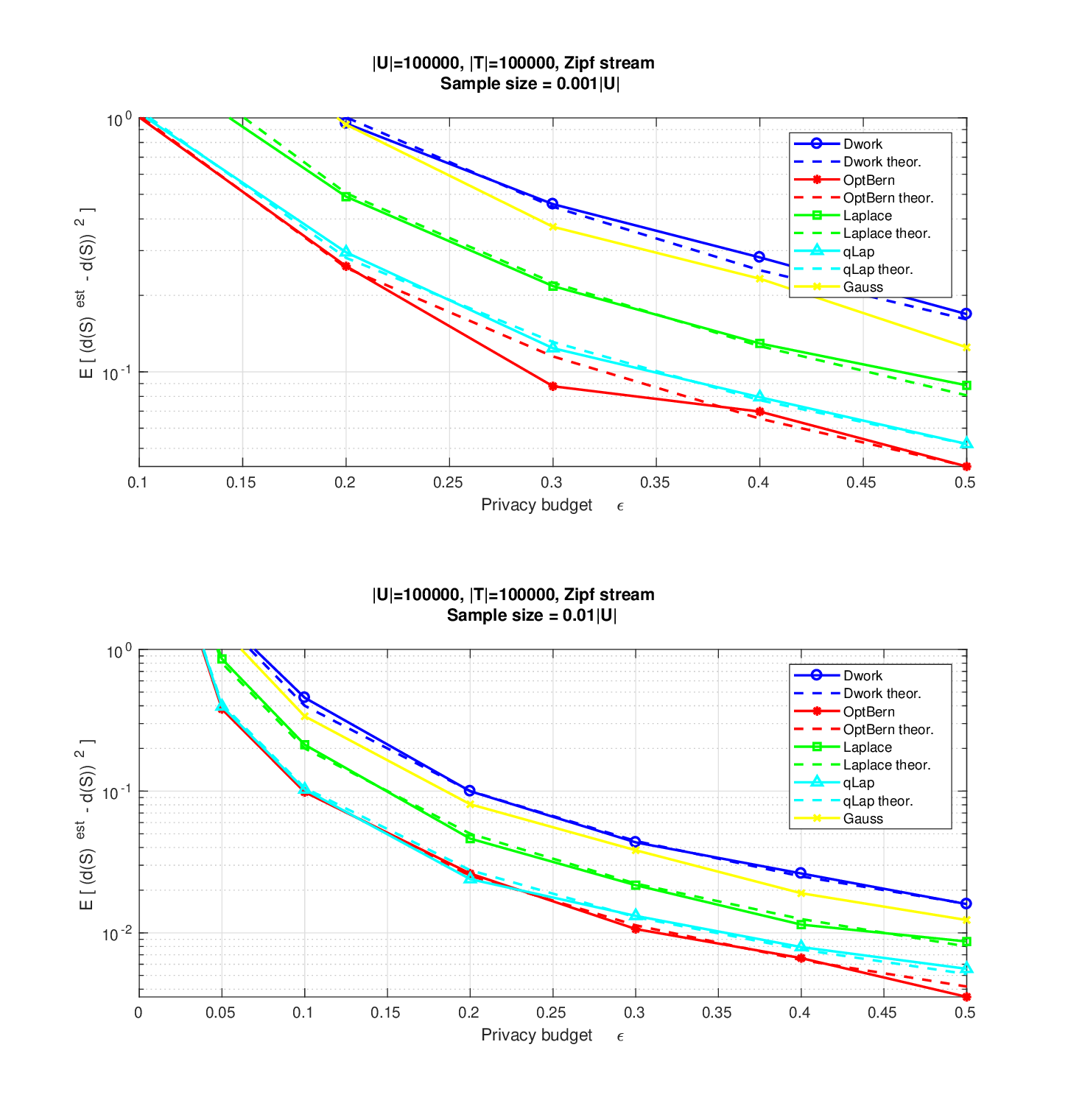}
    \caption{MSE for varying privacy budget, and for various sample sizes (Zipf Stream)}
	\label{FIG_MSE_zipf}
\end{figure}

We present the experimental results in Figures \ref{FIG_MSE_uniform} and \ref{FIG_MSE_zipf}.
The experimental and theoretical mean squared errors coincide for all algorithms.
The first thing we observe is the superiority of all the algorithms we propose over Dwork's algorithm.
The second is that, among our algorithms, the binary ones (Optimal Bernoulli Density Estimator denoted as optBern, Quantized Laplace Density Estimator denoted qLap) clearly beat the continuous ones (Laplace Density Estimator, Gauss Density Estimator).
Finally, we mention the robustness of all algorithms to the input stream distribution; the differences between uniform and zipfian are insignificant.

We next examine our second evaluation metric, that is, the probability $\Prob(| \tilde{d}-d(\mathcal{S}) | \geq \alpha)$ (which we call probability of error for simplicity).
Our experiments are again for fixed sample size (as a fraction of the universe size), for fixed $\alpha$, and for varying privacy budget $\eps$.
The illustrated probability of error is the empirical probability, computed over $1000$ repetitions per $\eps$. 

\begin{figure}%[H]%[ht!]
    \centering
    \includegraphics[scale=0.5]{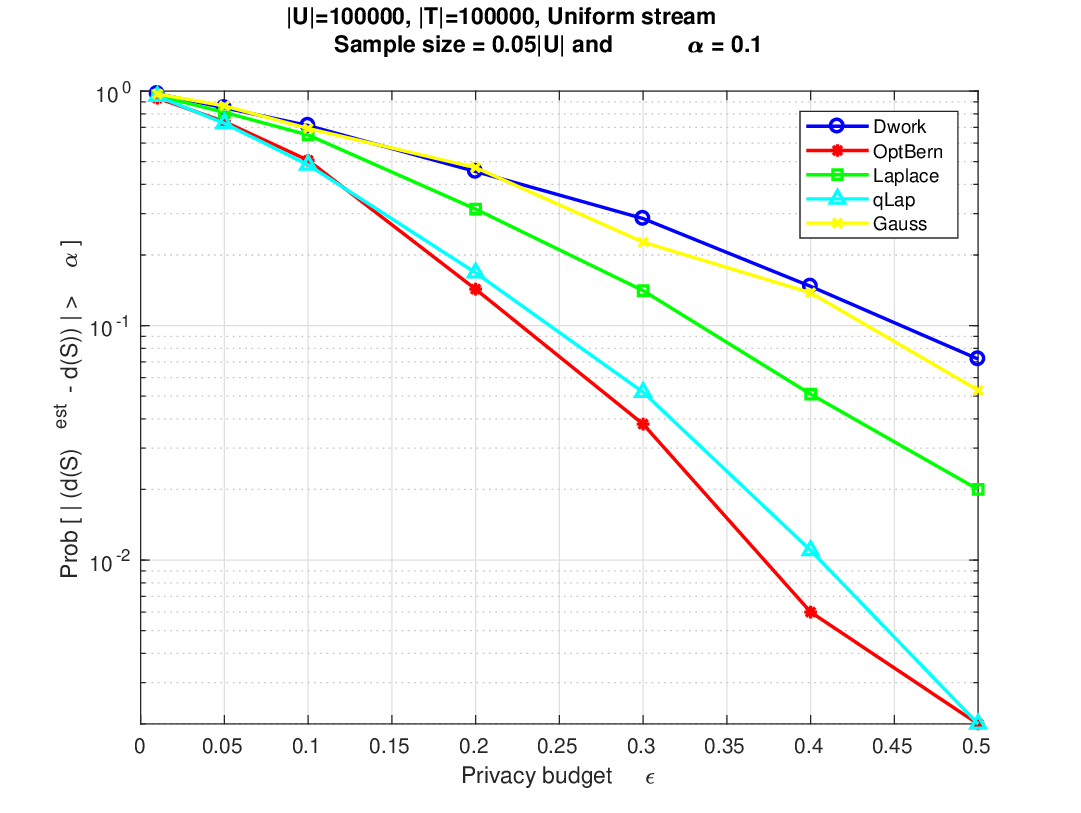}
    \caption{Probability of error for varying privacy budget (Uniform Stream)}
	\label{FIG_prob_error_uniform}
\end{figure}

\begin{figure}%[H]%[ht!]
    \centering
    \includegraphics[scale=0.5]{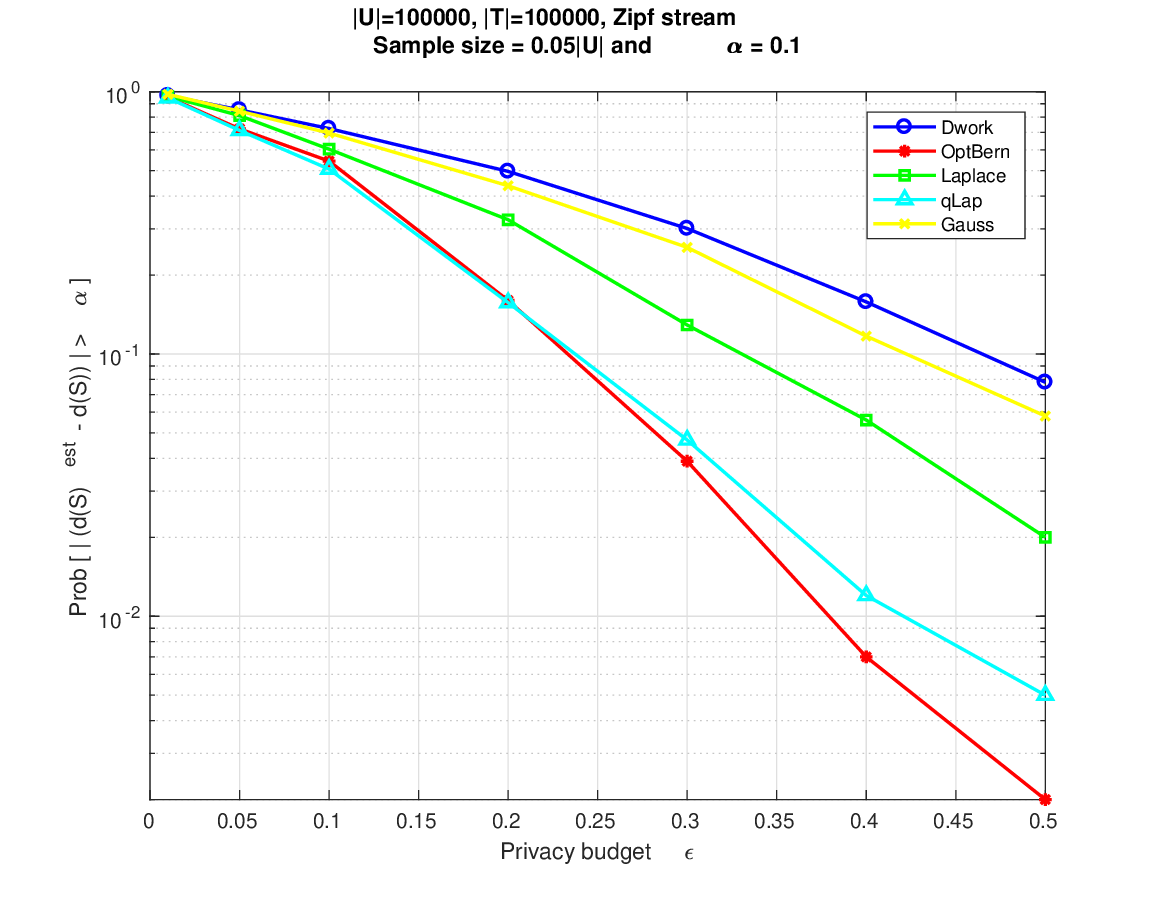}
    \caption{Probability of error for varying privacy budget (Zipf Stream)}
	\label{FIG_prob_error_zipf}
\end{figure}

We present the experimental results in Figures \ref{FIG_prob_error_uniform} and \ref{FIG_prob_error_zipf}.
Our second metric confirms the ranking of the algorithms in terms of performance, which we observed in the first set of experiments.
The key thing to notice is that the bounds on the probability of error which we computed theoretically are not tight;
we computed the tightest version of the bound, and the resulting probability was significantly larger than all the empirical probabilities we plot in Figures \ref{FIG_prob_error_uniform}, \ref{FIG_prob_error_zipf}.

\section{Discussion}
To summarize, in this chapter we examined differential privacy in the streaming model, and addressed the particular problem of pan-private stream density estimation.
We analyzed the sampling-based pan-private density estimator proposed by Dwork et al. \cite{dwork2010pan},
and we improved it by identifying that it does not use all the allocated privacy budget.
We managed to tackle this problem by proposing modifications to Dwork's estimator
that are based on optimally tuning the Bernoulli distributions it uses, as well as on using continuous distributions (Laplace, Gaussian).
Based on both our theoretical and experimental results, the Optimal Bernoulli Density Estimator is the algorithm of choice.

\chapter{Conclusions \& Future Work}
In this thesis, we developed differentially private algorithms to analyze distributed and streaming data.

\section{Distributed Model}
In the distributed model, we addressed the problem of distributed learning of Bayesian Networks with differential privacy.
After formally describing our model and identifying the challenges that arise when moving from a centralized to a distributed environment,
we examined three solutions, namely Sharing the Noisy Sufficient Statistics, Noisy Majority Voting and Sharing the Noisy Model.

Through our theoretical analysis and our detailed experimental evaluation,
we conclude that the first approach,
Sharing the Noisy Sufficient Statistics,
asymptotically outperforms the other approaches when the privacy budget exceeds a certain threshold.
However, its performance is significantly worse below that threshold,
and it is more sensitive to the data dimension, the number of data holders and the Bayesian Network degree.
The second approach, Noisy Majority Voting is robust to all the parameters we examined (privacy budget, data dimensions, number of data holders, Bayesian Network degree),
and significantly outperforms the other algorithms when the data are high-dimensional.

There are many directions in which the work we presented in the distributed model (Chapter 3) can be continued.
Firstly, instead of having the data holders share their full, high-dimensional frequency distributions,
we may ask them to send a more compact representation, such as a Sketch.
Secondly, as we argued, the (global) scoring function (mutual information) used in the Bayesian Network learning problem is non-linear and hence it is not composable from its local values;
however, sketch-based methods for approximating the global entropy of distributed streams have been developed (Gabel et al. \cite{gabel2017anarchists}),
and we could apply their ideas in our setting as well.
Finally, although we followed a score-based approach in learning the global Bayesian Network,
we remark that an approach based on Bayesian model averaging could also be examined.

%%%%%%%%%%%%%%%%%%%

\section{Streaming Model}
In the streaming model, we addressed the problem of pan-private stream density estimation.
We analyzed for the first time the sampling-based pan-private density estimator proposed by Dwork et al. \cite{dwork2010pan},
and we improved it by identifying that it does not use all the allocated privacy budget.
We managed to tackle this problem by proposing novel modifications to Dwork's estimator
that are based on optimally tuning the Bernoulli distributions it uses, as well as on using continuous distributions (Laplace, Gaussian).
Based on both our theoretical and experimental results, the Optimal Bernoulli Density Estimator is the algorithm of choice.

There are also several directions in which the work we presented in the streaming model (Chapter 3) can be extended.
Firstly, we identify that the sampling step performed by our algorithm's is naive;
we may apply more sophisticated sampling techniques
that are optimized for our particular problem,
such as Distinct Sampling \cite{gibbons2001distinct}.
Secondly, we could examine other approaches in density estimation,
based on the well-known FM-Sketch and its variants \cite{gibbons2016distinct};
although our early work on these approaches was not fruitful,
there is definitely much more to try.
Finally, it would be interesting to combine the model we described in Chapter 2 with that of Chapter 3,
and develop differentially private algorithms for our problems in the model of distributed streams.
For example, Zhang et al. \cite{zhang2017learning} address the problem of learning Bayesian Network from distributed streams, although without privacy considerations.

\printbibliography[heading=bibintoc]

@inproceedings{dwork2006calibrating,
  title={Calibrating noise to sensitivity in private data analysis},
  author={Dwork, Cynthia and McSherry, Frank and Nissim, Kobbi and Smith, Adam},
  booktitle={Theory of cryptography conference},
  pages={265--284},
  year={2006},
  organization={Springer}
}

@article{dwork2014algorithmic,
  title={The algorithmic foundations of differential privacy},
  author={Dwork, Cynthia and Roth, Aaron and others},
  journal={Foundations and Trends{\textregistered} in Theoretical Computer Science},
  volume={9},
  number={3--4},
  pages={211--407},
  year={2014},
  publisher={Now Publishers, Inc.}
}

@article{sweeney2002k,
  title={k-anonymity: A model for protecting privacy},
  author={Sweeney, Latanya},
  journal={International Journal of Uncertainty, Fuzziness and Knowledge-Based Systems},
  volume={10},
  number={05},
  pages={557--570},
  year={2002},
  publisher={World Scientific}
}

@article{zhang2017privbayes,
  title={Privbayes: Private data release via bayesian networks},
  author={Zhang, Jun and Cormode, Graham and Procopiuc, Cecilia M and Srivastava, Divesh and Xiao, Xiaokui},
  journal={ACM Transactions on Database Systems (TODS)},
  volume={42},
  number={4},
  pages={25},
  year={2017},
  publisher={ACM}
}

@inproceedings{zhang2014privbayes,
  title={PrivBayes: private data release via bayesian networks},
  author={Zhang, Jun and Cormode, Graham and Procopiuc, Cecilia M and Srivastava, Divesh and Xiao, Xiaokui},
  booktitle={Proceedings of the 2014 ACM SIGMOD International Conference on Management of Data},
  pages={1423--1434},
  year={2014},
  organization={ACM}
}

@inproceedings{dwork2010pan,
  title={Pan-Private Streaming Algorithms.},
  author={Dwork, Cynthia and Naor, Moni and Pitassi, Toniann and Rothblum, Guy N and Yekhanin, Sergey},
  booktitle={ Proceedings of The First Symposium on Innovations in Computer Science},
  pages={66--80},
  year={2010}
}

@inproceedings{kifer2011no,
  title={No free lunch in data privacy},
  author={Kifer, Daniel and Machanavajjhala, Ashwin},
  booktitle={Proceedings of the 2011 ACM SIGMOD International Conference on Management of data},
  pages={193--204},
  year={2011},
  organization={ACM}
}

@inproceedings{hsu2014differential,
  title={Differential privacy: An economic method for choosing epsilon},
  author={Hsu, Justin and Gaboardi, Marco and Haeberlen, Andreas and Khanna, Sanjeev and Narayan, Arjun and Pierce, Benjamin C and Roth, Aaron},
  booktitle={Computer Security Foundations Symposium (CSF), 2014 IEEE 27th},
  pages={398--410},
  year={2014},
  organization={IEEE}
}

@book{mitzenmacher2005probability,
  title={Probability and computing: Randomized algorithms and probabilistic analysis},
  author={Mitzenmacher, Michael and Upfal, Eli},
  year={2005},
  publisher={Cambridge university press}
}

@article{warner1965randomized,
  title={Randomized response: A survey technique for eliminating evasive answer bias},
  author={Warner, Stanley L},
  journal={Journal of the American Statistical Association},
  volume={60},
  number={309},
  pages={63--69},
  year={1965},
  publisher={Taylor \& Francis}
}

@inproceedings{mcsherry2007mechanism,
  title={Mechanism design via differential privacy},
  author={McSherry, Frank and Talwar, Kunal},
  booktitle={Foundations of Computer Science, 2007. FOCS'07. 48th Annual IEEE Symposium on},
  pages={94--103},
  year={2007},
  organization={IEEE}
}

@article{sarwate2013signal,
  title={Signal processing and machine learning with differential privacy: Algorithms and challenges for continuous data},
  author={Sarwate, Anand D and Chaudhuri, Kamalika},
  journal={IEEE signal processing magazine},
  volume={30},
  number={5},
  pages={86--94},
  year={2013},
  publisher={IEEE}
}

@article{kasiviswanathan2011can,
  title={What can we learn privately?},
  author={Kasiviswanathan, Shiva Prasad and Lee, Homin K and Nissim, Kobbi and Raskhodnikova, Sofya and Smith, Adam},
  journal={SIAM Journal on Computing},
  volume={40},
  number={3},
  pages={793--826},
  year={2011},
  publisher={SIAM}
}

@inproceedings{pathak2010multiparty,
  title={Multiparty differential privacy via aggregation of locally trained classifiers},
  author={Pathak, Manas and Rane, Shantanu and Raj, Bhiksha},
  booktitle={Advances in Neural Information Processing Systems},
  pages={1876--1884},
  year={2010}
}

@inproceedings{alhadidi2012secure,
  title={Secure distributed framework for achieving $\varepsilon$-differential privacy},
  author={Alhadidi, Dima and Mohammed, Noman and Fung, Benjamin CM and Debbabi, Mourad},
  booktitle={International Symposium on Privacy Enhancing Technologies Symposium},
  pages={120--139},
  year={2012},
  organization={Springer}
}

@inproceedings{goryczka2013secure,
  title={Secure multiparty aggregation with differential privacy: A comparative study},
  author={Goryczka, Slawomir and Xiong, Li and Sunderam, Vaidy},
  booktitle={Proceedings of the Joint EDBT/ICDT 2013 Workshops},
  pages={155--163},
  year={2013},
  organization={ACM}
}

@thesis{zheng2015differential,
  title={The differential privacy of Bayesian inference},
  author={Zheng, Shijie},
  type={Bachelor's Thesis},
  school={Harvard University},
  year={2015}
}

@inproceedings{rajkumar2012differentially,
  title={A differentially private stochastic gradient descent algorithm for multiparty classification},
  author={Rajkumar, Arun and Agarwal, Shivani},
  booktitle={Artificial Intelligence and Statistics},
  pages={933--941},
  year={2012}
}

@inproceedings{shokri2015privacy,
  title={Privacy-preserving deep learning},
  author={Shokri, Reza and Shmatikov, Vitaly},
  booktitle={Proceedings of the 22nd ACM SIGSAC conference on computer and communications security},
  pages={1310--1321},
  year={2015},
  organization={ACM}
}

@article{ji2014differentially,
  title={Differentially private distributed logistic regression using private and public data},
  author={Ji, Zhanglong and Jiang, Xiaoqian and Wang, Shuang and Xiong, Li and Ohno-Machado, Lucila},
  journal={BMC medical genomics},
  volume={7},
  number={1},
  pages={S14},
  year={2014},
  publisher={BioMed Central}
}

@inproceedings{xie2016data,
  title={Data-weighted ensemble learning for privacy-preserving distributed learning},
  author={Xie, Liyang and Plis, Sergey and Sarwate, Anand D},
  booktitle={2016 IEEE International Conference on Acoustics, Speech and Signal Processing (ICASSP)},
  pages={2309--2313},
  year={2016},
  organization={IEEE}
}

@article{chaudhuri2011differentially,
  title={Differentially private empirical risk minimization},
  author={Chaudhuri, Kamalika and Monteleoni, Claire and Sarwate, Anand D},
  journal={Journal of Machine Learning Research},
  volume={12},
  number={Mar},
  pages={1069--1109},
  year={2011}
}

@inproceedings{hamm2016learning,
  title={Learning privately from multiparty data},
  author={Hamm, Jihun and Cao, Yingjun and Belkin, Mikhail},
  booktitle={International Conference on Machine Learning},
  pages={555--563},
  year={2016}
}

@inproceedings{abdel2014darm,
  title={DARM: a privacy-preserving approach for distributed association rules mining on horizontally-partitioned data},
  author={Abdel Wahab, Omar and Hachami, Moulay Omar and Zaffari, Arslan and Vivas, Mery and Dagher, Gaby G},
  booktitle={Proceedings of the 18th International Database Engineering \& Applications Symposium},
  pages={1--8},
  year={2014},
  organization={ACM}
}

@inproceedings{barak2007privacy,
  title={Privacy, accuracy, and consistency too: a holistic solution to contingency table release},
  author={Barak, Boaz and Chaudhuri, Kamalika and Dwork, Cynthia and Kale, Satyen and McSherry, Frank and Talwar, Kunal},
  booktitle={Proceedings of the twenty-sixth ACM SIGMOD-SIGACT-SIGART symposium on Principles of database systems},
  pages={273--282},
  year={2007},
  organization={ACM}
}

@article{hay2010boosting,
  title={Boosting the accuracy of differentially private histograms through consistency},
  author={Hay, Michael and Rastogi, Vibhor and Miklau, Gerome and Suciu, Dan},
  journal={Proceedings of the VLDB Endowment},
  volume={3},
  number={1-2},
  pages={1021--1032},
  year={2010},
  publisher={VLDB Endowment}
}

@article{xu2013differentially,
  title={Differentially private histogram publication},
  author={Xu, Jia and Zhang, Zhenjie and Xiao, Xiaokui and Yang, Yin and Yu, Ge and Winslett, Marianne},
  journal={The VLDB Journal—The International Journal on Very Large Data Bases},
  volume={22},
  number={6},
  pages={797--822},
  year={2013},
  publisher={Springer-Verlag New York, Inc.}
}

@inproceedings{ding2011differentially,
  title={Differentially private data cubes: optimizing noise sources and consistency},
  author={Ding, Bolin and Winslett, Marianne and Han, Jiawei and Li, Zhenhui},
  booktitle={Proceedings of the 2011 ACM SIGMOD International Conference on Management of data},
  pages={217--228},
  year={2011},
  organization={ACM}
}

@inproceedings{cormode2012differentially,
  title={Differentially private spatial decompositions},
  author={Cormode, Graham and Procopiuc, Cecilia and Srivastava, Divesh and Shen, Entong and Yu, Ting},
  booktitle={2012 IEEE 28th international conference on Data engineering (ICDE)},
  pages={20--31},
  year={2012},
  organization={IEEE}
}

@inproceedings{xiao2010differential,
  title={Differential privacy via wavelet transforms},
  author={Xiao, Xiaokui and Wang, Guozhang and Gehrke, Johannes},
  year={2010},
  organization={2010 IEEE 26th International Conference on Data Engineering (ICDE)}
}

@inproceedings{williams2010probabilistic,
  title={Probabilistic inference and differential privacy},
  author={Williams, Oliver and McSherry, Frank},
  booktitle={Advances in Neural Information Processing Systems},
  pages={2451--2459},
  year={2010}
}

@inproceedings{dimitrakakis2014robust,
  title={Robust and private Bayesian inference},
  author={Dimitrakakis, Christos and Nelson, Blaine and Mitrokotsa, Aikaterini and Rubinstein, Benjamin IP},
  booktitle={International Conference on Algorithmic Learning Theory},
  pages={291--305},
  year={2014},
  organization={Springer}
}

@inproceedings{zhang2016differential,
  title={On the Differential Privacy of Bayesian Inference.},
  author={Zhang, Zuhe and Rubinstein, Benjamin IP and Dimitrakakis, Christos and others},
  booktitle={Proceedings AAAI'16 Proceedings of the Thirtieth AAAI Conference on Artificial Intelligence},
  pages={2365--2371},
  year={2016}
}

@article{foulds2016theory,
  title={On the theory and practice of privacy-preserving Bayesian data analysis},
  author={Foulds, James and Geumlek, Joseph and Welling, Max and Chaudhuri, Kamalika},
  journal={arXiv preprint arXiv:1603.07294},
  year={2016}
}

@article{bernstein2017differentially,
  title={Differentially Private Learning of Undirected Graphical Models Using Collective Graphical Models},
  author={Bernstein, Garrett and McKenna, Ryan and Sun, Tao and Sheldon, Daniel and Hay, Michael and Miklau, Gerome},
  journal={arXiv preprint arXiv:1706.04646},
  year={2017}
}

@inproceedings{chen2015differentially,
  title={Differentially private high-dimensional data publication via sampling-based inference},
  author={Chen, Rui and Xiao, Qian and Zhang, Yu and Xu, Jianliang},
  booktitle={Proceedings of the 21th ACM SIGKDD International Conference on Knowledge Discovery and Data Mining},
  pages={129--138},
  year={2015},
  organization={ACM}
}

@inproceedings{ping2017datasynthesizer,
  title={DataSynthesizer: Privacy-preserving synthetic datasets},
  author={Ping, Haoyue and Stoyanovich, Julia and Howe, Bill},
  booktitle={Proceedings of the 29th International Conference on Scientific and Statistical Database Management},
  pages={42},
  year={2017},
  organization={ACM}
}

@inproceedings{su2016differentially,
  title={Differentially private multi-party high-dimensional data publishing},
  author={Su, Sen and Tang, Peng and Cheng, Xiang and Chen, Rui and Wu, Zequn},
  booktitle={2016 IEEE 32nd International Conference on Data Engineering (ICDE)},
  pages={205--216},
  year={2016},
  organization={IEEE}
}

@article{pearl1986fusion,
  title={Fusion, propagation, and structuring in belief networks},
  author={Pearl, Judea},
  journal={Artificial intelligence},
  volume={29},
  number={3},
  pages={241--288},
  year={1986},
  publisher={Elsevier}
}

@book{koller2009probabilistic,
  title={Probabilistic graphical models: principles and techniques},
  author={Koller, Daphne and Friedman, Nir},
  year={2009},
  publisher={MIT press}
}

@article{chow1968approximating,
  title={Approximating discrete probability distributions with dependence trees},
  author={Chow, C and Liu, Cong},
  journal={IEEE transactions on Information Theory},
  volume={14},
  number={3},
  pages={462--467},
  year={1968},
  publisher={IEEE}
}

@book{cormen2009introduction,
  title={Introduction to algorithms},
  author={Cormen, Thomas H and Leiserson, Charles E and Rivest, Ronald L and Stein, Clifford},
  year={2009},
  publisher={MIT press}
}

@article{howell2008asymptotic,
  title={On asymptotic notation with multiple variables},
  author={Howell, Rodney R},
  journal={Dept. of Computing and Information Sciences, Tech. Rep},
  year={2008}
}

@techreport{harris1975statistical,
  title={The statistical estimation of entropy in the non-parametric case},
  author={Harris, Bernard},
  year={1975},
  institution={WISCONSIN UNIV-MADISON MATHEMATICS RESEARCH CENTER}
}

@article{basharin1959statistical,
  title={On a statistical estimate for the entropy of a sequence of independent random variables},
  author={Basharin, Georgij P},
  journal={Theory of Probability \& Its Applications},
  volume={4},
  number={3},
  pages={333--336},
  year={1959},
  publisher={SIAM}
}

@article{strack2014impact,
  title={Impact of HbA1c measurement on hospital readmission rates: analysis of 70,000 clinical database patient records},
  author={Strack, Beata and DeShazo, Jonathan P and Gennings, Chris and Olmo, Juan L and Ventura, Sebastian and Cios, Krzysztof J and Clore, John N},
  journal={BioMed research international},
  volume={2014},
  year={2014},
  publisher={Hindawi}
}

@inproceedings{dwork2010differential,
  title={Differential privacy in new settings},
  author={Dwork, Cynthia},
  booktitle={Proceedings of the twenty-first annual ACM-SIAM symposium on Discrete Algorithms},
  pages={174--183},
  year={2010},
  organization={SIAM}
}

@inproceedings{dwork2010continual,
  title={Differential privacy under continual observation},
  author={Dwork, Cynthia and Naor, Moni and Pitassi, Toniann and Rothblum, Guy N},
  booktitle={Proceedings of the forty-second ACM symposium on Theory of computing},
  pages={715--724},
  year={2010},
  organization={ACM}
}

@article{kellaris2014differentially,
  title={Differentially private event sequences over infinite streams},
  author={Kellaris, Georgios and Papadopoulos, Stavros and Xiao, Xiaokui and Papadias, Dimitris},
  journal={Proceedings of the VLDB Endowment},
  volume={7},
  number={12},
  pages={1155--1166},
  year={2014},
  publisher={VLDB Endowment}
}

@article{flajolet1985probabilistic,
  title={Probabilistic counting algorithms for data base applications},
  author={Flajolet, Philippe and Martin, G Nigel},
  journal={Journal of computer and system sciences},
  volume={31},
  number={2},
  pages={182--209},
  year={1985},
  publisher={Elsevier}
}

@article{cormode2005improved,
  title={An improved data stream summary: the count-min sketch and its applications},
  author={Cormode, Graham and Muthukrishnan, Shan},
  journal={Journal of Algorithms},
  volume={55},
  number={1},
  pages={58--75},
  year={2005},
  publisher={Elsevier}
}

@book{garofalakis2016data,
  title={Data Stream Management: Processing High-Speed Data Streams},
  author={Garofalakis, Minos and Gehrke, Johannes and Rastogi, Rajeev},
  year={2016},
  publisher={Springer}
}

@article{cormode2011synopses,
  title={Synopses for massive data: Samples, histograms, wavelets, sketches},
  author={Cormode, Graham and Garofalakis, Minos and Haas, Peter J and Jermaine, Chris and others},
  journal={Foundations and Trends{\textregistered} in Databases},
  volume={4},
  number={1--3},
  pages={1--294},
  year={2011},
  publisher={Now Publishers, Inc.}
}

@inproceedings{mir2011pan,
  title={Pan-private algorithms via statistics on sketches},
  author={Mir, Darakhshan and Muthukrishnan, S and Nikolov, Aleksandar and Wright, Rebecca N},
  booktitle={Proceedings of the thirtieth ACM SIGMOD-SIGACT-SIGART symposium on Principles of database systems},
  pages={37--48},
  year={2011},
  organization={ACM}
}

@inproceedings{indyk2000stable,
  title={Stable distributions, pseudorandom generators, embeddings and data stream computation},
  author={Indyk, Piotr},
  booktitle={Proceedings 41st Annual Symposium on Foundations of Computer Science},
  pages={189},
  year={2000},
  organization={IEEE}
}

@inproceedings{cormode2002comparing,
  title={Comparing data streams using hamming norms (how to zero in)},
  author={Cormode, Graham and Datar, Mayur and Indyk, Piotr and Muthukrishnan, S},
  booktitle={Proceedings of the 28th international conference on Very Large Data Bases},
  pages={335--345},
  year={2002},
  organization={VLDB Endowment}
}

@inproceedings{chan2010private,
  title={Private and continual release of statistics},
  author={Chan, TH Hubert and Shi, Elaine and Song, Dawn},
  booktitle={International Colloquium on Automata, Languages, and Programming},
  pages={405--417},
  year={2010},
  organization={Springer}
}

@inproceedings{bolot2013private,
  title={Private decayed predicate sums on streams},
  author={Bolot, Jean and Fawaz, Nadia and Muthukrishnan, S and Nikolov, Aleksandar and Taft, Nina},
  booktitle={Proceedings of the 16th International Conference on Database Theory},
  pages={284--295},
  year={2013},
  organization={ACM}
}

@inproceedings{chan2012differentially,
  title={Differentially private continual monitoring of heavy hitters from distributed streams},
  author={Chan, T-H Hubert and Li, Mingfei and Shi, Elaine and Xu, Wenchang},
  booktitle={International Symposium on Privacy Enhancing Technologies Symposium},
  pages={140--159},
  year={2012},
  organization={Springer}
}

@inproceedings{fan2012real,
  title={Real-time aggregate monitoring with differential privacy},
  author={Fan, Liyue and Xiong, Li},
  booktitle={Proceedings of the 21st ACM international conference on Information and knowledge management},
  pages={2169--2173},
  year={2012},
  organization={ACM}
}

@inproceedings{friedman2014privacy,
  title={Privacy-Preserving Distributed Stream Monitoring.},
  author={Friedman, Arik and Sharfman, Izchak and Keren, Daniel and Schuster, Assaf},
  booktitle={Proceedings of The Network and Distributed System Security Symposium },
  year={2014}
}

@article{bonomi2016differentially,
  title={On Differentially Private Longest Increasing Subsequence Computation in Data Stream.},
  author={Bonomi, Luca and Xiong, Li},
  journal={Transactions on Data Privacy},
  volume={9},
  number={1},
  pages={73--100},
  year={2016}
}

@article{upadhyay2014differentially,
  title={Differentially private linear algebra in the streaming model},
  author={Upadhyay, Jalaj},
  journal={arXiv preprint arXiv:1409.5414},
  year={2014}
}

@article{sharfman2007geometric,
  title={A geometric approach to monitoring threshold functions over distributed data streams},
  author={Sharfman, Izchak and Schuster, Assaf and Keren, Daniel},
  journal={ACM Transactions on Database Systems (TODS)},
  volume={32},
  number={4},
  pages={23},
  year={2007},
  publisher={ACM}
}

@inproceedings{gabel2017anarchists,
  title={Anarchists, unite: Practical entropy approximation for distributed streams},
  author={Gabel, Moshe and Keren, Daniel and Schuster, Assaf},
  booktitle={Proceedings of the 23rd ACM SIGKDD International Conference on Knowledge Discovery and Data Mining},
  pages={837--846},
  year={2017},
  organization={ACM}
}

@inproceedings{gibbons2001distinct,
  title={Distinct sampling for highly-accurate answers to distinct values queries and event reports},
  author={Gibbons, Phillip B},
  booktitle={VLDB},
  volume={1},
  pages={541--550},
  year={2001}
}

@incollection{gibbons2016distinct,
  title={Distinct-values estimation over data streams},
  author={Gibbons, Phillip B},
  booktitle={Data Stream Management},
  pages={121--147},
  year={2016},
  publisher={Springer}
}

@article{zhang2017learning,
  title={Learning Graphical Models from a Distributed Stream},
  author={Zhang, Yu and Tirthapura, Srikanta and Cormode, Graham},
  journal={arXiv preprint arXiv:1710.02103},
  year={2017}
}

\end{document}